\documentclass[11pt]{article}

\usepackage[preprint]{acl}

\usepackage{times}
\usepackage{latexsym}

\usepackage[T1]{fontenc}

\usepackage[utf8]{inputenc}

\usepackage{microtype}

\usepackage{inconsolata}

\usepackage{graphicx}

\usepackage{amsmath,amsfonts,bm}

\def\eqref#1{equation~\ref{#1}}

\def\1{\bm{1}}

\DeclareMathAlphabet{\mathsfit}{\encodingdefault}{\sfdefault}{m}{sl}
\SetMathAlphabet{\mathsfit}{bold}{\encodingdefault}{\sfdefault}{bx}{n}

\def\gE{{\mathcal{E}}}

\def\gG{{\mathcal{G}}}

\def\gV{{\mathcal{V}}}

\def\sA{{\mathbb{A}}}

\def\sF{{\mathbb{F}}}

\def\sK{{\mathbb{K}}}

\DeclareMathOperator*{\argmin}{arg\,min}

\usepackage{hyperref}
\usepackage{url}

\usepackage{booktabs}

\usepackage{xspace}
\usepackage{multirow}
\usepackage{multicol}
\usepackage{booktabs}
\usepackage{amssymb}
\usepackage{makecell}
\usepackage{subcaption}
\usepackage{placeins}
\usepackage{dsfont}
\usepackage{enumitem}
\usepackage{caption}
\usepackage{titlesec}

\usepackage{cleveref}
\usepackage{fontawesome5}
\usepackage{twemojis}

\usepackage{xcolor}
\usepackage{pifont}
\usepackage{soul}
\usepackage{tikz}
\usepackage[most]{tcolorbox}

\usetikzlibrary{arrows.meta, positioning, calc, shapes.callouts}

\definecolor{softgreen}{RGB}{46,160,67}
\definecolor{softred}{RGB}{218,54,51}
\newcommand{\greencheck}{\textcolor{softgreen}{\Large\ding{51}}}
\newcommand{\redcross}{\textcolor{softred}{\Large\ding{55}}}

\definecolor{templatebluetitle}{HTML}{2E62AB}
\colorlet{templateblueframe}{templatebluetitle}
\definecolor{templateblueback}{HTML}{F2F6FC}
\definecolor{promptgreentitle}{HTML}{267349}
\colorlet{promptgreenframe}{promptgreentitle}
\definecolor{promptgreenback}{HTML}{F1F8F3}
\tcbuselibrary{skins}

\tcbuselibrary{listings}

\newlength{\boxtextindent}
\makeatletter
\begingroup
\catcode`\^^M=\active%
\gdef\tpl@activeeol{\catcode`\^^M=\active\def^^M{\par\futurelet\tpl@next\tpl@blankcheck}}%
\gdef\tpl@blankcheck{\ifx\tpl@next^^M\vskip\baselineskip\fi}%
\endgroup
\def\tpl@nolig#1{\catcode`#1=\active\begingroup\lccode`\~=`#1\relax
  \lowercase{\endgroup\def~{\leavevmode\kern\z@\char`#1\relax}}}
\newcommand{\tpl@inputtext}{%
  \begingroup
  \let\do\@makeother\dospecials
  \catcode`\ =10
  \tpl@nolig\`\tpl@nolig\'\tpl@nolig\-\tpl@nolig\<\tpl@nolig\>\tpl@nolig\,%
  \tpl@activeeol
  \parindent=\z@
  \spaceskip=\fontdimen2\font plus \fontdimen2\font minus .3\fontdimen2\font
  \xspaceskip=\spaceskip
  \@@input\kvtcb@listingfile\relax
  \endgroup}
\tcbset{justified text only/.style={listing@process@inside={\tpl@inputtext}{\tpl@inputtext}}}
\makeatother

\tcbset{
  promptbox/.style={
    justified text only,
    width=\textwidth,
    enhanced,
    colback=promptgreenback,
    colframe=promptgreenframe,
    colbacktitle=promptgreentitle,
    coltitle=white,
    coltext=black,
    fonttitle=\bfseries,
    boxrule=0.6pt,
    arc=2pt,
    boxsep=1mm,
    left=\boxtextindent-1mm,
    right=\boxtextindent-1mm,
    top=2mm,
    bottom=2mm,
    before skip=2mm,
    lefttitle=2mm,
    righttitle=2mm,
    toptitle=0.5mm,
    bottomtitle=0.5mm,
    fontupper=\ttfamily\small
  }
}

\tcbset{
  templatebox/.style={
    justified text only,
    width=\linewidth,
    enhanced,
    colback=templateblueback,
    colframe=templateblueframe,
    colbacktitle=templatebluetitle,
    coltitle=white,
    coltext=black,
    fonttitle=\bfseries,
    boxrule=0.6pt,
    arc=2pt,
    boxsep=1mm,
    left=\boxtextindent-1mm,
    right=\boxtextindent-1mm,
    top=2mm,
    bottom=2mm,
    before skip=2mm,
    lefttitle=2mm,
    righttitle=2mm,
    toptitle=0.5mm,
    bottomtitle=0.5mm,
    fontupper=\ttfamily\small
  }
}

\definecolor{myorange}{RGB}{220, 90, 30}
\definecolor{myblue}{RGB}{50, 115, 210}
\definecolor{mygreen}{RGB}{55, 148, 55}
\definecolor{mypink}{RGB}{215, 45, 145}
\definecolor{myyellow}{RGB}{235, 210, 20}
\definecolor{forwardarrow}{RGB}{203, 42, 48}   
\definecolor{backwardarrow}{RGB}{146, 43, 168} 
\colorlet{hopone}{myorange}
\colorlet{hoptwo}{myblue}
\colorlet{hopthree}{mygreen}
\colorlet{hopfour}{mypink}
\colorlet{hopfive}{myyellow}
\colorlet{pathhopone}{myorange}
\colorlet{pathhoptwo}{myblue}
\colorlet{pathhopthree}{mygreen}
\colorlet{pathhopfour}{mypink}
\colorlet{pathhopfive}{myyellow}

\titlespacing\paragraph{0pt}{1.1ex}{1.1ex plus 0.3ex}

\setlist[itemize]{leftmargin=*, labelsep=0.5em, itemsep=1pt, parsep=1pt}

\newcommand{\Benchmark}{GraphNIAH\xspace}
\newcommand{\model}[1]{\textsc{#1}}
\newcommand{\PaperTitle}{\Benchmark: A Benchmark for Long-Context Reasoning \\over Real-World Graphs}
\definecolor{projectpageicon}{HTML}{1A73E8}
\newcommand{\projectpage}[1]{%
  \href{#1}{\textcolor{projectpageicon}{\faGlobe}\ \textcolor{blue}{\texttt{Project Page}}}%
}
\newcommand{\datasetpage}[1]{%
  \href{#1}{\raisebox{-.2ex}{\twemoji[height=1.35\fontcharht\font`X]{hugs}}\ \textcolor{blue}{\texttt{Dataset}}}%
}
\newcommand{\githubrepo}[1]{%
  \href{#1}{\faGithub\ \textcolor{blue}{\texttt{Code}}}%
}

\title{\PaperTitle{}}

\author{Minh-Nam Tran and Abulhair Saparov\\
  Department of Computer Science\\
  Purdue University\\
  West Lafayette, IN 47907, USA \\
  \texttt{\{tran555,asaparov\}@purdue.edu}
  \\[0.5em]
  {
    \small
    \projectpage{https://trminhnam.github.io/projects/graphniah/}
    \qquad
    \datasetpage{https://huggingface.co/datasets/minhnam/GraphNIAH}
    \qquad
    \githubrepo{https://github.com/trminhnam/GraphNIAH}
  }
}

\begin{document}

\maketitle

\begin{abstract}
  The needle-in-a-haystack (NIAH) benchmark was introduced as a test for measuring a model's ability to find small relevant facts within a very large body of input text. However, existing NIAH benchmarks each suffer from one or more of the following limitations: (1) reliance on synthetic data, so their evaluations may not generalize to real-world settings; (2) reuse of previously published datasets, leading to the risk of benchmark contamination; (3) dependence on manual annotation, which requires human effort to create and update data and is prone to future benchmark contamination; and (4) unrealistic simplification of long-context multi-hop reasoning where each next step is trivial given previous steps, while real-world reasoning often requires planning over multiple possible proof paths. To address these limitations, we introduce \Benchmark, a dynamic benchmark built from real citation relationships among scientific papers published after the evaluated models' knowledge cutoff date, which utilizes real-world information while ensuring no benchmark contamination. The fully automated benchmark construction pipeline samples representative subgraphs of the citation graph formed by these papers and converts each subgraph into a long-context QA example with a verifiable answer. We use \Benchmark to evaluate eight LLMs from the \model{GPT-5.6}, \model{Gemini-3}, and \model{Qwen3.5} families. We find that accuracy falls as questions require longer reasoning paths: at question depths of 16 and 32, the \model{GPT-5.6} and \model{Qwen3.5} models answer YES/NO questions no better than random guessing, mostly by answering NO; four of the eight models answer no counting example correctly, and the others model get worse as the subgraph size increases. These findings suggest that current LLMs often stop searching the citation relationships in their context before they reach the answer, even though every fact they need is in the context.
\end{abstract}

\begin{table*}[tbp]
  \centering
  \resizebox{\textwidth}{!}{%
    \begin{tabular}{lccccc}
      \toprule
      \textbf{Benchmark} &
      \makecell{\textbf{Multi-hop}\\\textbf{Reasoning}} &
      \makecell{\textbf{Real}\\\textbf{Data}} &
      \makecell{\textbf{Updatable}} &
      \makecell{\textbf{Construction without}\\\textbf{Human Annotation}} &
      \makecell{\textbf{Non-linear}\\\textbf{Reasoning}} \\
      \midrule
      Naive NIAH~\citep{needle}
      & \redcross & \redcross & \greencheck & \greencheck & \redcross \\
      RULER~\citep{hsieh2024ruler}
      & \greencheck & \redcross & \redcross & \greencheck & \redcross \\
      BABILong~\citep{kuratov2024babilong}
      & \greencheck & \redcross & \greencheck & \greencheck & \redcross \\
      LV-Eval~\citep{lveval_yuan2025lveval}
      & \greencheck & \greencheck & \redcross & \redcross & \redcross \\
      NeedleBench~\citep{li2025needlebench}
      & \greencheck & \redcross & \greencheck & \greencheck & \greencheck \\
      Counting-Stars~\citep{song-etal-2025-counting}
      & \greencheck & \redcross & \greencheck & \greencheck & \redcross \\
      Sequential-NIAH~\citep{yu-etal-2025-sequential}
      & \greencheck & \redcross & \redcross & \greencheck & \redcross \\
      NeedleChain~\citep{moon2025needlechain}
      & \greencheck & \redcross & \greencheck & \greencheck & \redcross \\
      NoLiMa~\citep{modarressi2025nolima}
      & \greencheck & \redcross & \greencheck & \greencheck & \redcross \\
      U-NIAH~\citep{uniah_gao2026u}
      & \greencheck & \redcross & \greencheck & \redcross & \redcross \\
      \textbf{\Benchmark} (Ours)
      & \greencheck & \greencheck & \greencheck & \greencheck & \greencheck \\
      \bottomrule
    \end{tabular}%
  }
  \caption{Comparison of NIAH benchmarks across five desiderata: (1) support for multi-hop reasoning, (2) use of real data, (3) the ability to update benchmark data in the future, (4) construction without manual annotation, and (5) support for non-linear reasoning. Here, non-linear reasoning requires searching among possible proof paths to find a valid proof or establish that none exists, which is in contrast with linear reasoning, where each next proof step is unambiguous given the previous steps.}
  \label{tab:benchmark_comparison}
\end{table*}

\section{Introduction}
\label{sec:introduction}

\begin{figure}[htbp]
  \centering
  \includegraphics[width=\columnwidth]{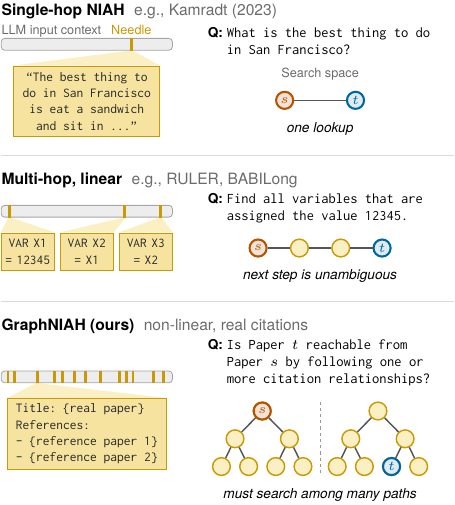}
  \caption{Three kinds of NIAH tests. In each row, the left column shows the LLM input context with its needles, and the yellow boxes show the text of needles; the right column shows the question and its search space from $s$ to $t$. Single-hop NIAH~\citep{needle} needs one lookup, and in RULER~\citep{hsieh2024ruler} and BABILong~\citep{kuratov2024babilong} the next step is unambiguous. In \Benchmark, needles describe papers and their references; $s$ and $t$ lie in different connected components (dashed line), so the model must search all paths from $s$ to answer \texttt{NO}.
  }
  \label{fig:introduction_summary}
\end{figure}

With the rapid evolution of large language models (LLMs) in recent years, their native context length has also increased significantly from 2,048 tokens of \model{GPT-3}~\citep{gpt3_NEURIPS2020_1457c0d6} to 128k tokens of \model{Llama-3.2}~\citep{grattafiori2024llama}, and now, to 1M tokens of \model{DeepSeek-V4}~\citep{deepseekai2026deepseekv4}. With context extension techniques, such as RoPE-based scaling~\citep{liu2024scaling, ding2024longrope} and long-context training~\citep{hu-etal-2025-longrecipe, gao-etal-2025-train}, context length can be extended to upwards of 4M tokens. With this ability to receive extremely long input text sequences, LLMs have been adopted for many complex long-context tasks, such as deep research~\citep{du2026deepresearch}, long-context summarization~\citep{laban-etal-2024-summary, wright-etal-2025-unstructured}, and long-horizon reasoning and planning~\citep{chen-etal-2025-lr2bench, erdogan2025planandact}. Furthermore, as downstream tasks increase in complexity and input prompts increase in length, it is becoming increasingly important for LLMs to be able to process key information in very long contexts, and then combine it with other information to solve such tasks~\citep{koh2022empirical, pal-etal-2025-tagging, adams2025longhealth}. The needle-in-a-haystack (NIAH) test~\citep{needle} evaluates whether LLMs can retrieve a small relevant fact, called a \emph{needle}, from a long context containing unrelated information. However, single-hop NIAH tests do not accurately reflect real-world scenarios where relevant facts are often derived from multiple parts of the context or conversation. Hence, several multi-hop NIAH benchmarks have been proposed, including RULER~\citep{hsieh2024ruler}, Sequential-NIAH~\citep{yu-etal-2025-sequential}, and U-NIAH~\citep{uniah_gao2026u}, which require models to connect multiple facts.

Despite these advances, existing NIAH benchmarks face at least one of four limitations (see Table~\ref{tab:benchmark_comparison}). First, many of them are built on synthetic data~\citep{vodrahalli2024michelangelo, song-etal-2025-counting, hsieh2024ruler, yu-etal-2025-sequential, modarressi2025nolima} or LLM-generated content~\citep{uniah_gao2026u}, which may poorly reflect real-world use cases. Second, some benchmarks, including ZeroSCROLLS~\citep{shaham-etal-2023-zeroscrolls}, LongBench~\citep{bai-etal-2024-longbench}, InfiBench~\citep{li2024infibench}, LongGenBench~\citep{liu-etal-2024-longgenbench}, and ICR\textsuperscript{2}~\citep{qiu-etal-2025-eliciting}, are built on previously published NLP datasets.
Relying on these published datasets leads to benchmark contamination when the data appears in the model's training corpus~\citep{sainz-etal-2023-nlp, xu2024benchmarking, choi2025how, cheng2025surveydatacontaminationlarge}.
Third, some NIAH benchmarks rely on human effort to create new datasets~\citep{lveval_yuan2025lveval, uniah_gao2026u}. Expanding or updating these benchmarks requires additional
human effort, making them difficult to scale; once the data is published, it is invariably incorporated into later LLMs' training data, resulting in future benchmark contamination. Finally, most current NIAH benchmarks require LLMs to connect retrieved facts along a single fixed logical path, where the next retrieval step is unambiguous given the previous step, to derive the final answer~\citep{hsieh2024ruler, yu-etal-2025-sequential, song-etal-2025-counting, moon2025needlechain}. However, downstream tasks and applications often involve interconnected information in the context, where the next retrieval step is not always unambiguous and the model must perform search over multiple potential proof paths~\citep{wang2026positiongraphshelplarge, edge2025localglobalgraphrag, wang-etal-2026-exploring-graph}. We use \textit{non-linear reasoning} to describe this kind of reasoning, which requires proof search to find the correct proof in a large space of possible proofs (or to find that no such proof exists).
Most NIAH benchmarks (see Table~\ref{tab:benchmark_comparison}) do not fully evaluate this kind of proof search; Figure~\ref{fig:introduction_summary} illustrates the difference.

\begin{figure*}[t]
  \centering
  \includegraphics[width=\textwidth]{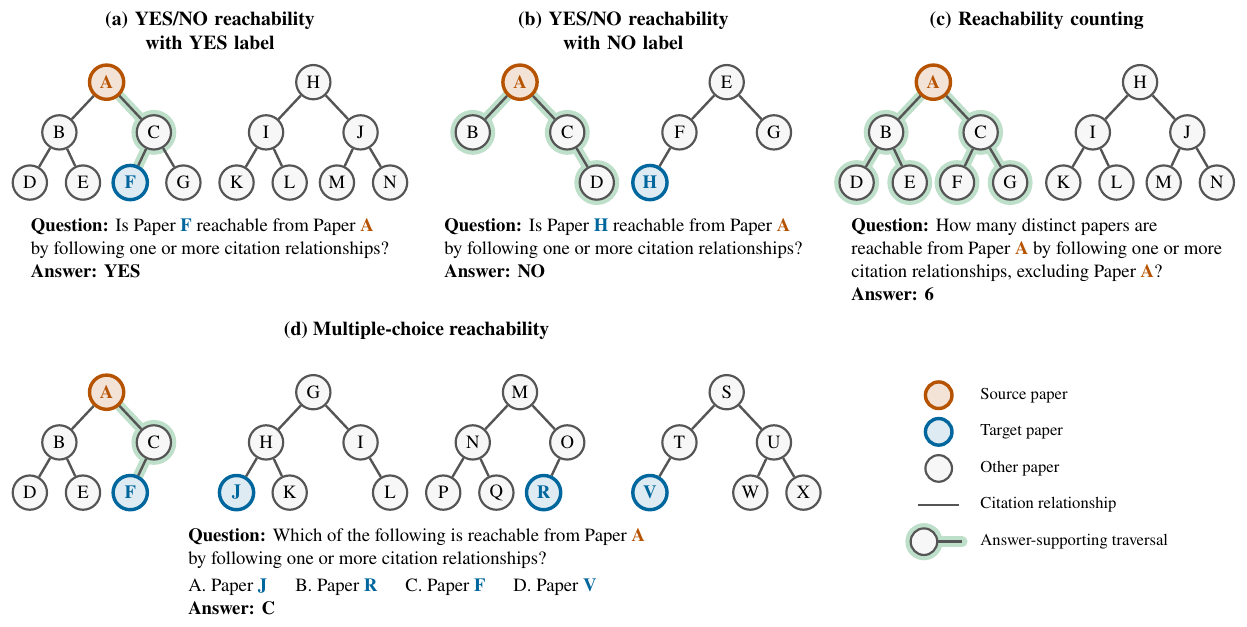}
  \vspace{-1.7em}
  \caption{
    Abstract visualization of three reachability tasks in \Benchmark.
    Each task is visualized with the question and ground truth answer from an example, together with the connected components constructed for that task. Each paper node in the diagram represents a paper, while each edge between two paper nodes represents the citation relationship between the two papers (i.e., \texttt{cites} or \texttt{cited\_by} relationships). Orange and blue paper nodes indicate source and target papers, respectively. For the YES/NO reachability task shown in panel (b), the source and target paper nodes are unreachable from each other because they belong to different connected components, resulting in the \texttt{NO} label for this example. Green highlights around some paper nodes and citation edges in the diagrams indicate the correct paths the model must take to travel from the source paper to the target paper. For the reachability counting task (c), the answer counts all other papers starting from the source paper, so the entire source component is highlighted in green.
  }
  \label{fig:reachability-task-examples}
  \vspace{-1.0em}
\end{figure*}

To address these gaps, we propose \Benchmark, a dynamic NIAH benchmark built from real scientific papers and their citation relationships. For each example, the benchmark construction pipeline samples a representative subgraph of the real citation graph from Semantic Scholar~\citep{kinney2023semantic}, in which each node is a paper published after the evaluated LLM's training cutoff date and nodes are connected by \texttt{cites} and \texttt{cited\_by} edges. The pipeline then converts each subgraph into a long-context QA example, where the papers in the subgraph and their bibliographies comprise the context, and the ground truth answer is computed directly from the subgraph. This process requires no manual annotation, so new examples can be generated automatically as new papers are published.
\Benchmark focuses on \textit{reachability}: whether a target paper is \textit{reachable} from a source paper, i.e., connected to it by a sequence of citation relationships stated in the input text. Each sampled subgraph contains several connected components, so that some papers are reachable from the source paper while others are not. \Benchmark supports three tasks: (1) YES/NO reachability, (2) multiple-choice reachability, and (3) reachability counting (see Figure~\ref{fig:reachability-task-examples}).
To measure and analyze LLMs' ability to perform non-linear reasoning over facts from very long contexts, \Benchmark is designed with many different variables as ``control knobs'', including question depth (i.e., the number of nodes on the path from the source paper to the target paper in the component), the subgraph size, the target context length, and various types of distracting content, i.e., information not supporting the correct reasoning or the answer.

We evaluate eight LLMs on \Benchmark{}: \model{GPT-5.6-Sol} and \model{GPT-5.6-Luna}~\citep{gpt-5.6}, \model{Gemini-3.1-Pro}~\citep{google2026gemini31pro}, \model{Gemini-3.8-Flash}~\citep{gemini-3.8-flash}, \model{Gemini-3.6-Flash}~\citep{gemini-3.6-flash}, and three \model{Qwen3.5} models~\citep{qwen3.5}: \model{Qwen3.5-397B-A17B}, \model{Qwen3.5-122B-A10B}, and \model{Qwen3.5-35B-A3B}. Our analysis shows three main findings. First, accuracy decreases as question depth grows, and at large depths, \model{GPT-5.6} and \model{Qwen3.5} models answer YES/NO questions no better than random guessing, mostly by answering NO (\S\ref{subsec:complexity_scaling}). Second, reachability counting is the hardest task: half of the models fail to correctly answer any counting question, and the other models struggle as the component size grows (\S\ref{subsec:complexity_scaling} and \S\ref{subsec:performance_on_different_reachability_tasks}). Third, neither the type of distracting content nor a higher level of reasoning effort reliably changes accuracy (\S\ref{subsec:effects_of_distractor_type} and \S\ref{subsec:effects_of_reasoning_effort}). We make the code and the benchmark data publicly available at \url{https://github.com/trminhnam/GraphNIAH}.

\section{Related Work}
\label{sec:related_work}
\paragraph{Long-Context Benchmarks.} The context length of LLMs has grown in recent years, leading to the emergence of long-context benchmarks. Broad benchmarks, such as ZeroSCROLLS~\citep{shaham-etal-2023-zeroscrolls}, LongBench~\citep{bai-etal-2024-longbench}, InfiBench~\citep{li2024infibench}, L-Eval~\citep{an-etal-2024-l}, and LongICLBench~\citep{li2025longcontext}, evaluate LLMs' ability to handle long-context inputs across different downstream tasks, including question answering (QA), summarization, and code understanding. These benchmarks are highly important as the input sequences of LLMs in real-world scenarios grow longer. However, many existing long-context benchmarks are built on previously published datasets~\citep{shaham-etal-2023-zeroscrolls, an-etal-2024-l}, making them vulnerable to benchmark contamination and difficult to update over time. In contrast, our work focuses on constructing a scalable and automatically updatable long-context evaluation using real-world data.

\paragraph{Needle-in-a-haystack Benchmarks.} \citet{needle} introduced the NIAH test, which hides one or a few needles in very large block of unrelated text and asks single-hop questions about the embedded information. Later benchmarks extend this setting: LV-Eval~\citep{lveval_yuan2025lveval} and LooGLE~\citep{li-etal-2024-loogle} cover several long-context tasks, RULER~\citep{hsieh2024ruler} adds multi-hop tasks with synthetic needles, U-NIAH~\citep{uniah_gao2026u} compares retrieval-augmented generation (RAG) systems with LLMs, and Sequential-NIAH~\citep{yu-etal-2025-sequential} asks the LLM to extract multiple needles and list them in their correct temporal or logical order. As Table~\ref{tab:benchmark_comparison} shows, most compared NIAH benchmarks rely on synthetic data, some are difficult to update, and all except NeedleBench~\citep{li2025needlebench} evaluate only linear reasoning, in which every next step is unambiguous given the previous steps~\citep{hsieh2024ruler, moon2025needlechain, yu-etal-2025-sequential, song-etal-2025-counting, modarressi2025nolima}. \Benchmark keeps the NIAH setting but uses real citation relationships among papers published after the evaluated LLMs' knowledge cutoff date and requires non-linear reasoning.

\paragraph{Dynamic Benchmarks.} Unlike static benchmarks, whose examples are fixed after creation, dynamic benchmarks can be updated over time~\citep{shirali2023a, choi-etal-2024-unigen, guan2025is, chen-etal-2025-benchmarking-large}. They follow three main strategies. The first strategy builds new examples from data published after the knowledge cutoff date of the evaluated LLMs, which lowers the risk of benchmark contamination~\citep{chen-etal-2025-benchmarking-large}: for example, recent arXiv papers in AcademicEval~\citep{zhang2025academiceval}, post-cutoff knowledge in AntiLeak-Bench~\citep{wu-etal-2025-antileakbench}, recent programming contests in LiveCodeBench~\citep{jain2025livecodebench}, and recent math competitions in LiveBench~\citep{white2025livebench}. The second strategy generates examples with rules, for example, from directed acyclic graphs in DyVal~\citep{zhu2024dyval}, from the rules of the Mathador game in Mathador-LM~\citep{kurtic-etal-2024-mathador}, or from randomly generated databases in S3Eval~\citep{lei-etal-2024-s3eval}. The third strategy uses LLMs to paraphrase examples of static benchmarks~\citep{chen2025dycodeeval, ying2024automating, zhu-etal-2024-inference} or to generate new ones~\citep{qian-etal-2024-varbench, wang-etal-2025-benchmark}. Some benchmarks combine these strategies, such as LatestEval~\citep{li2024latesteval}, DARG~\citep{zhang2024darg}, and C\textsuperscript{2}LEVA~\citep{li-etal-2025-c2leva}. \Benchmark combines the first two: it samples subgraphs from a citation graph of papers published after the knowledge cutoff date and converts each subgraph into an example with rules, so new examples can be generated automatically as new papers appear.

\section{\Benchmark Benchmark}
\label{sec:benchmark}
Each example in \Benchmark is generated using a subgraph which we sample from the citation graph in the Semantic Scholar Academic Graph Datasets~\citep{kinney2023semantic}. To sample these subgraphs, we use a preferential-attachment-based sampling procedure inspired by the scale-free nature of real-world citation graphs~\citep{wang2010scale, zhong2024scale}.

\subsection{Task Definition}
\label{subsec:task_definition}

In \Benchmark, each example consists of a question, a context, and a ground truth answer. The context contains descriptions of scientific papers and their references, which indicate the relationships between the papers. Given a source paper, our task asks one of the three types of questions: (1) whether a target paper is reachable via a path of citation relationships, (2) which paper out of a list of candidates is reachable, or (3) how many other papers are reachable. In this subsection, we define the citation subgraph of each example and the reachability tasks.

\subsubsection{Citation Subgraph}
\label{subsec:context_graph}

The Semantic Scholar citation graph is far too large to fit into the context window of an LLM, so for each example $q$, we sample a subgraph $\gG_q = (\gV_q, \gE_q)$ of the Semantic Scholar graph. Each node in $\gV_q$ is a paper, and each directed edge $(u, v) \in \gE_q$ means that paper $u$ cites paper $v$, or equivalently, that $v$ is cited by $u$. %

\subsubsection{Reachability Tasks}
\label{subsec:reachability_tasks}

In all three reachability tasks, the fundamental subtask is to determine whether a paper is \emph{reachable} from another paper via a sequence of edges, where each edge is either \texttt{cites} or \texttt{cited\_by}; Or equivalently, whether there exists an undirected path between the corresponding source and target nodes.
To construct examples in which some papers are reachable from a given source paper and others are not, we sample subgraphs containing multiple connected components. For a paper $u$, let $\sK_u$ denote the connected component of $\gG_q$ that contains $u$; then a paper $v$ is reachable from $u$ if and only if $v \in \sK_u$. Importantly, reachability is defined with respect to the subgraph $\gG_q$, rather than the Semantic Scholar citation graph. Thus, two papers in different connected components of $\gG_q$ are considered unreachable from each other, even if they are connected in the Semantic Scholar citation graph. The model must therefore answer each question using only the citation relationships stated in the context. Based on this definition of reachability, we formulate three reachability tasks: YES/NO reachability, multiple-choice reachability, and reachability counting.

We use $q_{\mathrm{YN}}$, $q_{\mathrm{MC}}$, and $q_{\mathrm{count}}$ to denote an example of the YES/NO, multiple-choice, and counting tasks, respectively, and $y_{\mathrm{YN}}$, $y_{\mathrm{MC}}$, and $y_{\mathrm{count}}$ for their ground truth answers. Additionally, $q_{\mathrm{YES}}$ and $q_{\mathrm{NO}}$ denote examples of the YES/NO reachability task with YES and NO labels, respectively.

\paragraph{YES/NO reachability.} Given a source paper $s$ and a target paper $t$, the model has to answer whether $t$ is reachable from $s$, i.e., whether $t \in \sK_s$. The ground truth answer is $y_{\mathrm{YN}} = \texttt{YES}$ if $t \in \sK_s$ and $y_{\mathrm{YN}} = \texttt{NO}$ otherwise.

\paragraph{Multiple-choice reachability.} Given a source paper $s$ and $N$ candidate papers $t_1, t_2, \ldots, t_N$, the model has to select the candidate that is reachable from $s$. In our implementation, there is exactly one reachable candidate $t_i$ from $s$ (i.e., $t_i \in \sK_s$ for some $1 \leq i \leq N$ and $t_j \notin \sK_s$ for all $j\ne i$), and the ground truth answer is $t_i$.

\paragraph{Reachability counting.} Starting from a source paper $s$, the model has to determine the number of distinct papers that are reachable from $s$, i.e., the component size of $\sK_s$, but excluding $s$ itself. The ground truth answer is a non-negative integer, formulated as $y_{\mathrm{count}} = \lvert \sK_s \rvert - 1$.

\subsection{Example Construction Pipeline}
\label{subsec:task_specific_example_construction_pipeline}

The generation of each example in \Benchmark is dependent on the following variables:
\begin{itemize}[topsep=5pt]
  \item \textbf{Task:} the task of the example (i.e., YES/NO reachability, multiple-choice reachability, or reachability counting).
  \item \textbf{Question depth $k$:} the number of nodes on the path from the source paper to the target paper, including both endpoints (\S\ref{subsec:complexity_metrics}).
  \item \textbf{Component size:} the number of nodes in each connected component (\S\ref{subsec:complexity_metrics}).
  \item \textbf{Distractor type:} \texttt{code}, \texttt{essays}, \texttt{books}, \texttt{wikipedia}, or \texttt{disjoint-component} (Step~4).
  \item \textbf{Target context length:} the maximum number of tokens in the context (Step~4).
\end{itemize}

In the subgraph of each example, the number of connected components is determined by its reachability task. For $q_{\mathrm{NO}}$, we need one component that contains $s$ and another that contains $t$. For $q_{\mathrm{YES}}$, we also generate two components, and $s$ and $t$ are in the same component, so that YES-label and NO-label examples differ only in the location of $t$. For $q_{\mathrm{MC}}$, we generate $N$ components: one contains $s$ and the reachable candidate $t_i$, and each other component contains one of the other candidates. For $q_{\mathrm{count}}$, we generate two components, one of which contains $s$ while the other prevents the model from simply counting all papers in the context.  It is important that these components are sampled i.i.d. (independent and identically distributed), so that there are no shortcuts that would allow the model to answer the question without searching through the citation relationships. We require that no two components share a paper.

Below, we describe the generative process for the connected components (Steps~1 and~2), and then how we convert the subgraph into the input prompt (Steps~3 and~4). 
Details are in Appendix~\ref{appdx:dataset_details}.

\paragraph{Step 1: Sampling the backbone.}

Our goal is to generate each example such that at least $d = k-1$ steps of search are \emph{required} to find the solution. To this end, we want the subgraph to contain at least two disjoint paths with $d$ edges starting from the source paper $s$: one \emph{true path}, which leads to the target paper $t$, and one \emph{distractor path}. Similarly, since the model may search backward from $t$, we want at least two disjoint paths with $d$ edges starting from $t$: the true path and another distractor path. Thus, to generate such a connected component, we first sample an undirected path of $3k-2$ distinct papers, all published on or after April 1, 2026. We partition this path into three paths of $k$ nodes each, where consecutive paths share one endpoint: a distractor path that ends at $s_c$, the ``true path'' from $s_c$ to $t_c$, and a distractor path that starts at $t_c$, where $c$ indicates the component. The sampling of every connected component $c$ in the subgraph of an example starts from this backbone structure. Note that $s_c$ and $t_c$ are not necessarily identified with the source and target paper specified in the question ($s$ and $t$);  %
for example, in $q_{\mathrm{NO}}$, $s$ and $t$ belong to two different components. 
Details are in Appendix~\ref{appdx:initial_component_structure}.

\paragraph{Step 2: Sampling the subgraph.}
Next, we repeatedly add papers from the Semantic Scholar citation graph to the subgraph. At each sampling step, we select one eligible node, i.e., a node that has neighboring nodes in the citation graph which have not yet been added to the subgraph, and whose component has not yet reached the target component size.
To select this node, we aim to mimic the scale-free degree distribution of citation graphs~\citep{bollobas2003directed}, we use a preferential-attachment rule akin to the one-parameter Chinese restaurant process~\citep{aldous2006exchangeability}: a node is selected with probability proportional to the number of children it has received in Step~2, or proportional to the concentration parameter $\alpha > 0$ if it has received none. 
For the selected node $v$, we sample the direction of the new edge $e$ uniformly from $\{\texttt{cites}, \texttt{cited\_by}\}$. Next, we uniformly sample a candidate child $c$ of $v$ that is connected via an edge with label $e$ in the Semantic Scholar citation graph and has not yet been added to the subgraph (if no such neighbor exists, we consider the other edge direction). 
We reject a candidate child if it would create an undirected cycle in the subgraph, or if the candidate child already belongs to another component in the subgraph. Each step adds one new leaf, so every component is a tree and its true path never changes. Sampling stops when each component reaches the termination criteria. Details are in Appendix~\ref{appdx:component_growth}.

\paragraph{Step 3: Rendering the subgraph in text.}
We render each paper in the subgraph as blocks of natural language text, each containing the title, authors, and publication information of that paper, followed by an APA-style reference list restricted to its citations within the component (see Appendix~\ref{appdx:needle_templates} for the template). A paper that cites no other paper in its component has no block of its own and appears only in the reference lists of the papers that cite it.

\paragraph{Step 4: Distractors and prompt assembly.}
As in other NIAH benchmarks, the context also contains distracting content. Each example uses one of five distractor types: \texttt{code}, \texttt{essays}, \texttt{books}, and \texttt{wikipedia}. For each type, unrelated text is drawn randomly from, respectively: source files of the Linux kernel\footnote{\url{https://github.com/torvalds/linux}}, Paul Graham's essays\footnote{\url{https://www.paulgraham.com}}, PG-19 books~\citep{Rae2020Compressive}, and Wikipedia passages~\citep{Lin_etal_SIGIR2021_Pyserini}. The \texttt{disjoint-component} type adds connected components sampled using the same procedure described in Steps~1--3 with the same question depth and component size. We add distractors until no additional distractors can fit within the target context length, shuffle them randomly with the blocks from Step~3, and insert everything into a task-specific prompt template (Appendix~\ref{appdx:prompt_templates}).

\subsection{Complexity Metrics}
\label{subsec:complexity_metrics}

We characterize each example with two complexity metrics: %
\begin{itemize}[topsep=5pt]
  \item \textbf{Question depth $k$:} The number of nodes on the true path from the source paper to the target paper, including both endpoints. For $q_{\mathrm{YES}}$ and $q_{\mathrm{MC}}$, this is the number of nodes on the unique path from $s$ to $t$ or $t_i$. For $q_{\mathrm{NO}}$ and $q_{\mathrm{count}}$, this is the number of nodes on the path from $s_c$ to $t_c$ in the backbone of each component $c$. In both cases, $k$ is the number of BFS steps required to solve each example.
  \item \textbf{Connected component size:} The number of nodes in a connected component, which is the same for all components of an example. Especially, for $q_{\mathrm{NO}}$ and $q_{\mathrm{count}}$, a correct answer requires the model to explore all nodes in $\sK_s$, so the component size directly measures the difficulty of these tasks.
\end{itemize}

\section{Experiments and Analysis}
\label{sec:experiments_and_analysis}
\begin{figure*}[htbp]
  \centering
  \includegraphics[width=\textwidth]{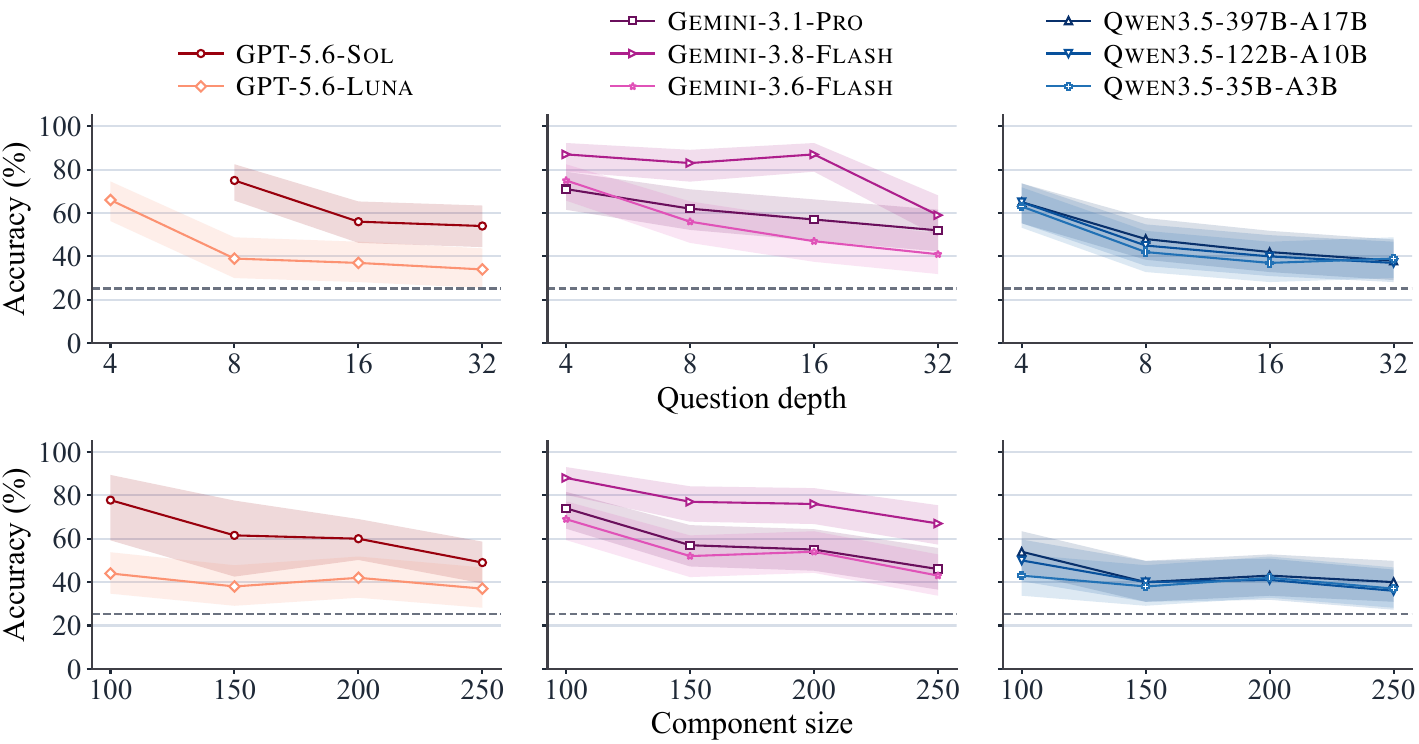}
  \caption{Accuracy by question depth (top row) and component size (bottom row). The dashed gray line denotes random guessing accuracy. Due to cost limitations, \model{GPT-5.6-Sol} was not run for question depth $4$.}
  \label{fig:accuracy_by_question_depth_and_component_size}
\end{figure*}

In this section, we aim to answer the following research questions:

\begin{itemize}[leftmargin=2.5em, labelsep=0.5em, itemsep=1pt, parsep=1pt]
  \item[\textbf{RQ1}] \textbf{Complexity scaling} (\S\ref{subsec:complexity_scaling}). Can frontier LLMs find papers in large in-context citation graphs via a sequence of citation relations (i.e., the reachability task) as the complexity of the questions scales with question depth and component size?
  \item[\textbf{RQ2}] \textbf{Task-specific performance} (\S\ref{subsec:performance_on_different_reachability_tasks}). How does the LLM's performance vary in different reachability tasks?
  \item[\textbf{RQ3}] \textbf{Context length scaling} (\S\ref{subsec:effects_of_context_length}). 
  How does the performance of LLMs on this task vary as context length increases?
  \item[\textbf{RQ4}] \textbf{Distractor content} (\S\ref{subsec:effects_of_distractor_type}). Are LLMs distracted more by some types of distracting content than others when answering long-context questions about citation relations?
  \item[\textbf{RQ5}] \textbf{Reasoning effort} (\S\ref{subsec:effects_of_reasoning_effort}). Does accuracy on this task depend on the reasoning effort of the LLMs?
\end{itemize}

\subsection{Experimental Setup}
\label{subsec:experimental_setup}

\paragraph{Models.} We evaluate the following frontier models on the three reachability tasks in \Benchmark: two models from the \model{GPT-5.6} model family~\citep{gpt-5.6} (\model{GPT-5.6-Sol} and \model{GPT-5.6-Luna}), \model{Gemini-3.1-Pro}~\citep{google2026gemini31pro}, \model{Gemini-3.8-Flash}~\citep{gemini-3.8-flash}, \model{Gemini-3.6-Flash}~\citep{gemini-3.6-flash}, and three \model{Qwen3.5} models~\citep{qwen3.5} (\model{Qwen3.5-397B-A17B}, \model{Qwen3.5-122B-A10B}, and \model{Qwen3.5-35B-A3B}). 
While we test all aforementioned models in the complexity scaling experiment, we choose to test a select subset of the models on other experiments to mitigate the high cost of inference. %
Appendix~\ref{appdx:models_and_inference_settings} provides the model specifications and inference parameters.

\paragraph{Data and variables.} Each experiment varies exactly one of the following variables: question depth $k \in \{ 4, 8, 16, 32 \}$, component size $ \in \{ 100, 150, 200, 250 \}$, reachability task (i.e., YES/NO reachability, multiple-choice reachability, or reachability counting), distractor type (\texttt{code}, \texttt{essays}, \texttt{books}, \texttt{wikipedia}, or \texttt{disjoint-component}), target context length (200,000, 500,000, or 922,000\footnote{GPT-5.6 models accept at most 922k input tokens.} tokens), or reasoning effort (\texttt{low}, \texttt{medium}, or \texttt{high}). For each value of the tested variable, we sample all other variables independently and uniformly at random: $k$ from $\{4, 8, 16, 32\}$, the component size from the integers in $[100, 250]$, the task from the three reachability tasks, and the distractor type from the set of all distractor types. All examples use a target context length of 200,000 tokens (except for the context-length experiment) and the concentration parameter $\alpha = 0.1$ (see Appendix~\ref{appdx:selection_of_sampling_parameter} for details on the selection of the concentration parameter). The number of examples for each experiment is reported in~\Cref{appdx:dataset_statistics}.

\paragraph{Evaluation metrics.} We extract the model answer from the last \verb|\answer{}| block of each response and report accuracy (the fraction of examples whose extracted answer matches the ground truth answer). For each configuration, we always include a random guessing baseline as a gray dashed line (see Appendix~\ref{appdx:random_guessing_baseline} for details). Error bars and shaded bands show 95\% Wilson confidence intervals~\citep{wilson1927probable}, and Appendix~\ref{appdx:extra_experimental_results} details all accuracy quantities with their intervals.

\subsection{Complexity Scaling Experiments}
\label{subsec:complexity_scaling}

We vary the question depth over $k \in \{4, 8, 16, 32\}$ in one experiment and the component size over $\{100, 150, 200, 250\}$ in another.

\paragraph{Overall results.}

Accuracy decreases as question depth increases for every model, but the point at which this decrease occurs differs across models (Figure~\ref{fig:accuracy_by_question_depth_and_component_size}). \model{GPT-5.6-Luna} and the \model{Qwen3.5} models drop from 63-66\% at $k=4$ to 39-48\% at $k=8$ and then seem to level off at 34-42\%. \model{Gemini-3.1-Pro} and \model{Gemini-3.6-Flash} decline steadily, from 71\% and 75\% at $k=4$, respectively, to 52\% and 41\% at $k=32$. \model{GPT-5.6-Sol} falls from 75\% at $k=8$ to 54\% at $k=32$. \model{Gemini-3.8-Flash} stays at 83-87\% up to $k=16$ and drops to 59\% only at $k=32$. Component size mainly affects the stronger models: as it grows from 100 to 250, \model{GPT-5.6-Sol} and the three \model{Gemini} models fall from 69-88\% to 43-67\%, whereas \model{GPT-5.6-Luna} and the \model{Qwen3.5} models stay between 36\% and 54\%, which are uniformly below the performance of \model{GPT-5.6-Sol} and the three \model{Gemini} models at every component size.

\paragraph{Breakdown by task.}

On YES/NO and multiple-choice reachability, accuracy declines with question depth for every model (Tables~\ref{tab:v1_accuracy_by_question_depth_and_yesno} and~\ref{tab:v1_accuracy_by_question_depth_and_multiple}). For $k \geq 16$, most models fall back to one answer in YES/NO reachability (Table~\ref{tab:v1_yesno_accuracy_by_question_depth_and_label}): \model{GPT-5.6-Luna}, \model{GPT-5.6-Sol}, and the \model{Qwen3.5} models answer \texttt{NO} to almost every example, which is no better than random guessing; \model{Gemini-3.1-Pro} and \model{Gemini-3.6-Flash} also lean towards \texttt{NO}, and only \model{Gemini-3.8-Flash} tends to answer \texttt{YES}. Component size has a weaker, model-dependent effect on these two tasks (Tables~\ref{tab:v1_accuracy_by_node_count_and_yesno} and~\ref{tab:v1_accuracy_by_node_count_and_multiple}): as component size increases from 100 to 250, YES/NO accuracy drops for all models except \model{GPT-5.6-Sol} and \model{Gemini-3.8-Flash} (e.g., \model{Gemini-3.1-Pro}: 86.84\% to 64.29\%), and multiple-choice accuracy shows no consistent trend.

Reachability counting shows the opposite pattern (Tables~\ref{tab:v1_accuracy_by_question_depth_and_counting} and~\ref{tab:v1_accuracy_by_node_count_and_counting}). For the four models with non-zero accuracy (\model{GPT-5.6-Sol}, \model{Gemini-3.1-Pro}, \model{Gemini-3.8-Flash}, and \model{Gemini-3.6-Flash}), question depth has no consistent effect, but accuracy falls sharply as the component size grows: from 100 to 250, \model{Gemini-3.8-Flash} drops from 79\% to 25\%, and the other three drop to 5\% or less.

Overall, we conclude that the two YES/NO and multiple-choice reachability tasks are more negatively affected by question depth than component size, while component size has a greater negative effect on the counting task than question depth.

\subsection{Performance by Reachability Tasks}
\label{subsec:performance_on_different_reachability_tasks}

\begin{figure}[htbp]
  \centering
  \includegraphics[width=\linewidth]{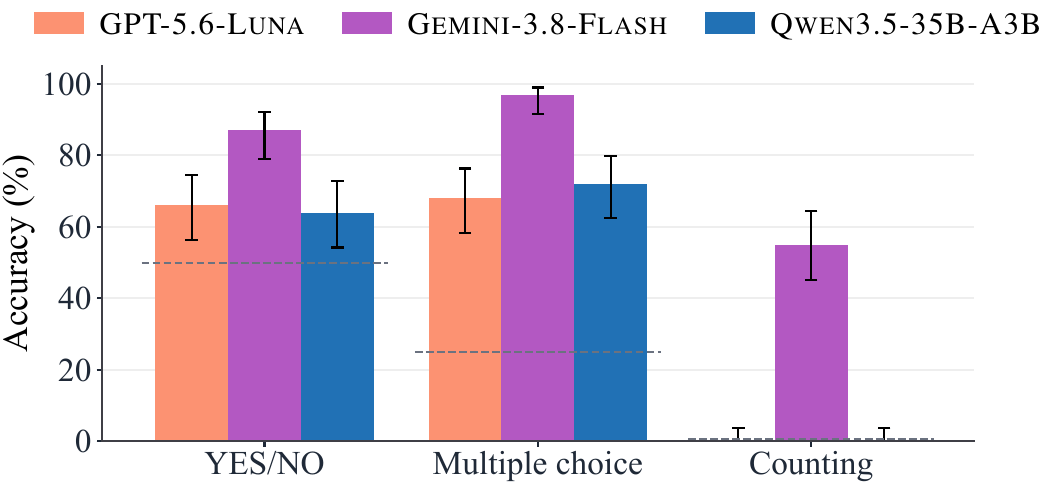}
  \caption{Accuracy by reachability task; each dashed gray line denotes random guessing accuracy.}
  \label{fig:accuracy_by_task}
\end{figure}

Figure~\ref{fig:accuracy_by_task} compares the three reachability tasks. Reachability counting is the hardest task for all three models: \model{GPT-5.6-Luna} and \model{Qwen3.5-35B-A3B} do not correctly answer any counting example, and \model{Gemini-3.8-Flash} reaches only 55\%, its lowest accuracy across the three tasks (Table~\ref{tab:v1_accuracy_by_question_format}). The complexity scaling experiments show the same phenomenon: \model{GPT-5.6-Luna} and all three \model{Qwen3.5} models do not correctly answer any counting example at any question depth or component size (Tables~\ref{tab:v1_accuracy_by_question_depth_and_counting} and~\ref{tab:v1_accuracy_by_node_count_and_counting}).%

The other two tasks differ with respect to each model's performance as compared to random guessing: on YES/NO reachability, \model{GPT-5.6-Luna}, \model{Qwen3.5-35B-A3B}, and \model{Gemini-3.8-Flash} exceed random guessing accuracy by 16, 14, and 37 percentage points, respectively, but on multiple-choice reachability by 43, 47, and 72 points. On YES/NO reachability, the models are also biased toward one answer: \model{GPT-5.6-Luna} and \model{Qwen3.5-35B-A3B} answer \texttt{NO} on 84 and 86 of the 100 examples. This is even more apparent for high question depth examples ($k \geq 8$; \Cref{tab:v1_yesno_no_answers_by_question_depth}), resulting in 100\% accuracy on NO-label examples but only 32\% and 28\% on YES-label examples, respectively (\Cref{tab:v1_yesno_accuracy_by_label}).

We hypothesize that multiple-choice reachability is easier because there is always a ``correct'' option, so the model never has to conclude that no target is reachable. To test this, we append a fifth option, ``None of these targets is reachable from the source.'', to the same multiple-choice examples and keep all other aspects of each example unchanged, so the new option is never correct. With this option, \model{GPT-5.6-Luna} and \model{Qwen3.5-35B-A3B} drop from 68\% and 72\% to 14\% and 15\%, respectively, no better than random guessing (20\%), because they choose the new option in 85 and 84 of the 100 examples, whereas \model{Gemini-3.8-Flash} stays at 97\% (Table~\ref{tab:v1_multiple_choice_none_of_the_above}). The two weaker models also produce roughly half as many output and reasoning tokens, while \model{Gemini-3.8-Flash} is unaffected (Table~\ref{tab:v1_multiple_choice_none_of_the_above_tokens}). This suggests that, once a negative answer is available, these models stop searching earlier and fall back to it, as in YES/NO reachability.

Overall, reachability counting is the hardest task, and for \model{GPT-5.6-Luna} and \model{Qwen3.5-35B-A3B}, the advantage of multiple-choice over YES/NO reachability disappears once a negative answer is available.

\subsection{Effects of Context Length}
\label{subsec:effects_of_context_length}

\begin{figure}[htbp]
  \centering
  \includegraphics[width=\linewidth]{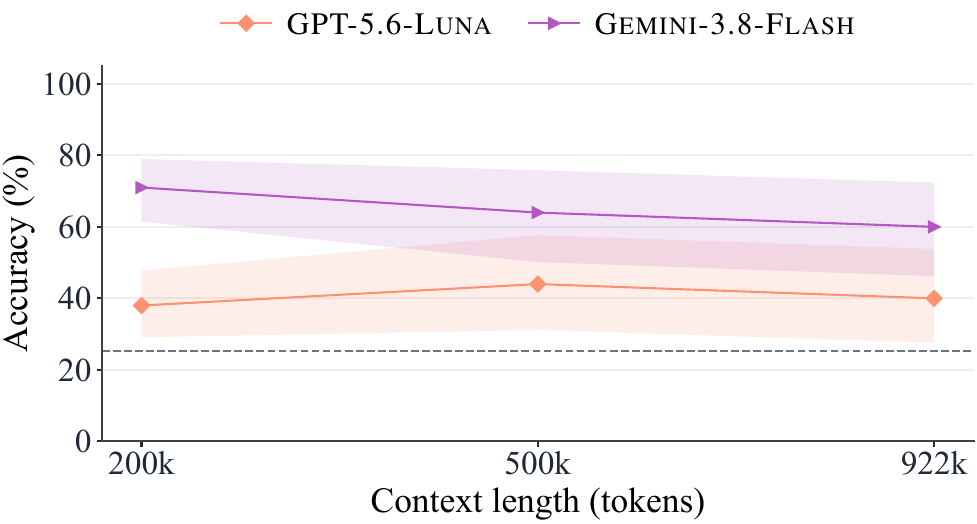}
  \caption{Accuracy by context length.}
  \label{fig:accuracy_by_context_length}
\end{figure}

We vary the target context length over 200,000, 500,000, and 922,000
tokens for \model{GPT-5.6-Luna} and \model{Gemini-3.8-Flash}. \model{GPT-5.6-Luna} stays at 38-44\%, while \model{Gemini-3.8-Flash} falls from 71\% at 200,000 tokens to 64\% and 60\% at 500,000 and 922,000 tokens, respectively (Figure~\ref{fig:accuracy_by_context_length}; Table~\ref{tab:v1_accuracy_by_context_length}). Neither change is significant (Fisher's exact test, all $p > 0.1$; Table~\ref{tab:v1_context_length_tests}).

\subsection{Effects of Distractor Type}
\label{subsec:effects_of_distractor_type}

\begin{figure}[htbp]
  \centering
  \includegraphics[width=\linewidth]{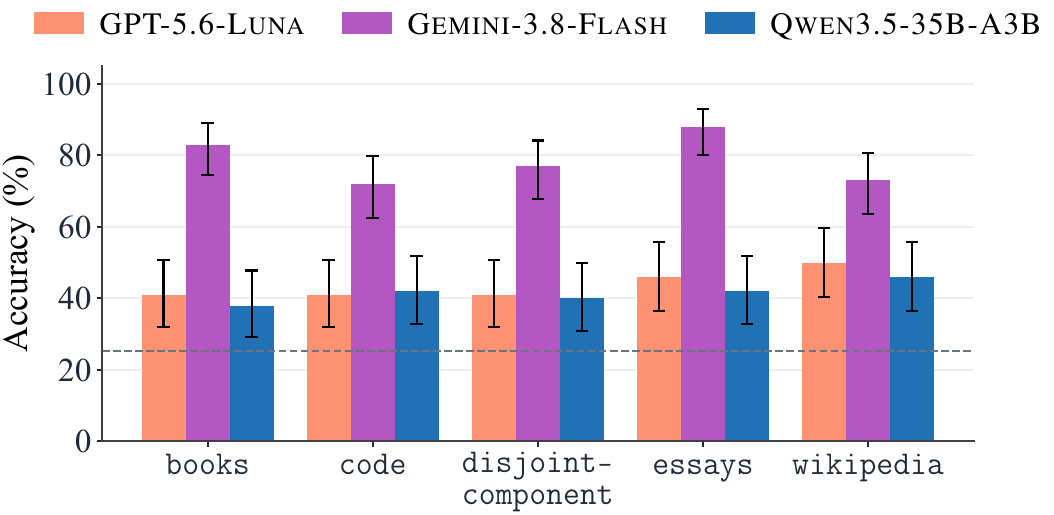}
  \caption{Accuracy by distractor type.}
  \label{fig:accuracy_by_distractor_type}
\end{figure}

We run experiments where we vary the type of the distractors for \model{GPT-5.6-Luna}, \model{Gemini-3.8-Flash}, and \model{Qwen3.5-35B-A3B}. The distractor type has no obvious effect on accuracy (Figure~\ref{fig:accuracy_by_distractor_type}; Table~\ref{tab:v1_accuracy_by_distractor_type}): \model{GPT-5.6-Luna} stays between 41\% and 50\%, \model{Qwen3.5-35B-A3B} between 38\% and 46\%, and \model{Gemini-3.8-Flash} between 72\% and 88\%. Only \model{Gemini-3.8-Flash} demonstrates a minor difference in performance across the different distractor types: an overall $\chi^2$ test across the five distractor types gives $p = 0.03$, mainly because it is more accurate with \texttt{essays} (88\%) than with \texttt{code} (72\%) and \texttt{wikipedia} (73\%). However, no pair of distractor types differs significantly after correcting for the ten pairwise comparisons (Fisher's exact test with Holm correction, all adjusted $p > 0.05$; Table~\ref{tab:v1_distractor_type_tests}).

\subsection{Effects of Reasoning Effort}
\label{subsec:effects_of_reasoning_effort}

\begin{figure}[htbp]
  \centering
  \includegraphics[width=\linewidth]{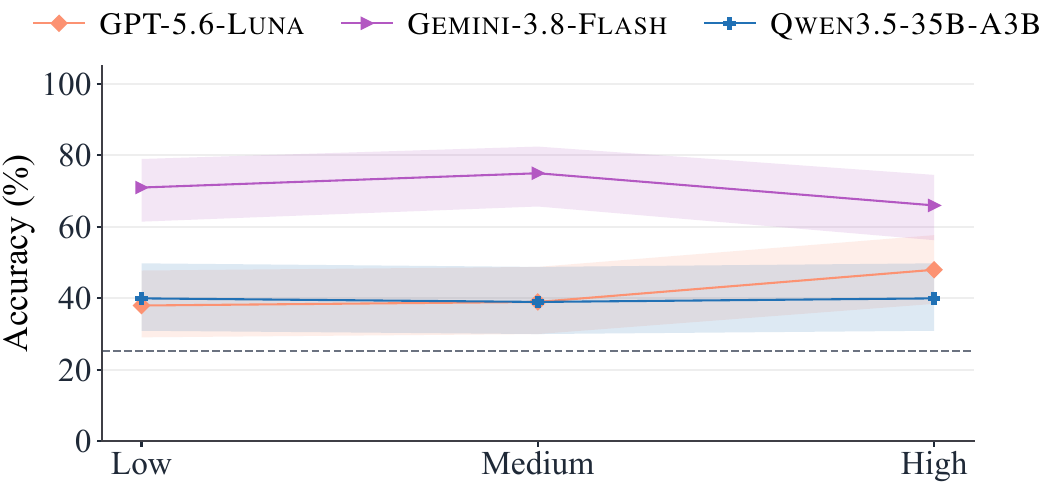}
  \caption{Accuracy by reasoning effort.}
  \label{fig:accuracy_by_reasoning_effort}
\end{figure}

We vary the reasoning effort of \model{GPT-5.6-Luna}, \model{Gemini-3.8-Flash}, and \model{Qwen3.5-35B-A3B}: \texttt{low}, \texttt{medium}, and \texttt{high} (Figure~\ref{fig:accuracy_by_reasoning_effort}). More effort does not reliably help the LLMs (Table~\ref{tab:v1_accuracy_by_reasoning_effort}): \model{Gemini-3.8-Flash} reaches 71\%, 75\%, and 66\% at \texttt{low}, \texttt{medium}, and \texttt{high} effort, respectively, and \model{Qwen3.5-35B-A3B} stays around 39\% and 40\% for all three reasoning effort levels. Only \model{GPT-5.6-Luna} improves from 38\% at \texttt{low} effort to 48\% at \texttt{high} effort, 
where it correctly answers 10 examples which were incorrectly answered at \texttt{low} effort, without producing any incorrect answers on an example which was answered correctly
(Table~\ref{tab:v1_reasoning_effort_paired_comparison}); however, the tradeoff is that, on average, \texttt{high} reasoning effort results in eight times as many output tokens, a longer response time by factor of 4.7, and 30.26\% higher API cost as compared to the \texttt{low} setting (Table~\ref{tab:v1_cost_and_response_time_reasoning_effort}).

\section{Conclusion}
\label{sec:conclusion}
In this paper, we introduce \Benchmark as an updatable, realistic needle-in-a-haystack benchmark. By automatically constructing examples from the Semantic Scholar citation graph using post-cutoff papers, we prevent benchmark contamination. 
\Benchmark supports multi-hop, non-linear reasoning, with diverse families of distracting content, while reducing reliance on synthetic data, human annotation, and static published datasets. Our experiments with eight LLMs show that this search remains difficult. 
Accuracy falls as question depth grows, and at large depths, most models do not outperform random guessing, often falling back to choosing the negative option. Half of the models do not correctly answer any counting example, and the weaker models significantly underestimate the number of reachable papers. Neither the type of distracting content nor a higher level of reasoning effort reliably closes this gap. These results suggest that current LLMs often stop exploring the context before they reach the answer, even though every required fact is present. %

There are multiple promising directions for future work. 
Firstly, we use a limited set of templates for the questions in the reachability tasks. Future work can therefore utilize a more diverse set of templates, or utilize LLMs to paraphrase the questions to produce more naturalistic language. Secondly, since \Benchmark is limited to the domain of citation graphs, we can further expand this work to other domains with different kinds of relational data. Thirdly, the experiments are performed over a limited number of examples per configuration due to the high cost of inference, so future work can evaluate LLMs with a higher number of examples to increase the statistical power of our analysis, strengthening the claims and comparisons. Finally, the findings show that when complexity increases, models perform no better than random guessing and fallback to choosing the negative option. We can therefore explore methods to encourage models to search more deeply when given a highly complex query and to avoid falling back to a default response.

\section*{Limitations}
\label{sec:limitations}
\Benchmark has several limitations. First, it covers one domain and one kind of relationship, citations between scientific papers, and all prompts are in English, so our findings may not transfer to other relational data or other kinds of long-context reasoning. Second, because long-context inference is expensive, we evaluate eight models in total, and some experiments only test two or three models, and each model is evaluated on at most 100 examples per setting, so many confidence intervals are fairly wide. Third, using papers from Semantic Scholar published after the training cutoff dates of the evaluated models reduces the risk of benchmark contamination but does not remove it: an earlier version of a paper, such as a preprint, may have been public before its recorded publication date, and a model may still guess an answer from how closely related two papers appear. The reachability counting task is one possible solution, since the model must find all reachable papers to answer correctly.

\section*{Acknowledgments}
\label{sec:acknowledgements}
We thank Abdulrahman Mahmoud for his valuable insights and constructive discussion. We gratefully acknowledge Google’s support through Gemini API credits.

\bibliography{bib/anthology1, bib/custom}

\appendix

\begin{center}
  {\large \bfseries Appendix for ``\PaperTitle{}''}
\end{center}

\noindent The appendix is structured as follows:
\begin{itemize}
  \item \Cref{appdx:dataset_details,appdx:selection_of_sampling_parameter,appdx:prompt_templates} describe the benchmark: its data sources, construction details,dataset statistics, and templates; how we select the concentration parameter $\alpha$;

  \item \Cref{appdx:models_and_inference_settings,appdx:random_guessing_baseline} describe the evaluation setup: the models and their inference settings, and the random guessing baseline.

  \item \Cref{appdx:extra_experimental_results,appdx:cost_and_response_time} report additional results: accuracy tables, results by task and by ground truth answer, statistical tests, and the inference cost and response time of each experiment.
\end{itemize}

\section{Benchmark Details}
\label{appdx:dataset_details}

This section provides the experimental configurations, dataset statistics, concentration-parameter selection procedure, question templates, and APA-style template for the papers.

\subsection{Data Sources and Preparation}
\label{appdx:dataset_configuration}

\paragraph{Data sources and snapshot.}
We use scientific paper metadata and citation relationships from release \texttt{2026-10-06} of the Semantic Scholar S2AG Datasets~\citep{kinney2023semantic}, and we use them for research purposes only, following their license and terms of use\footnote{\url{https://api.semanticscholar.org/license/}}.

For the \texttt{wikipedia} distractors, we retrieve Wikipedia passages with BM25, using the pre-built index of \texttt{pyserini}~\citep{Lin_etal_SIGIR2021_Pyserini}.

\subsection{Benchmark Construction Details}
\label{appdx:benchmark_construction_details}

\subsubsection{Citation Path Sampling}
\label{appdx:citation_path_sampling}

In Step~1 (\S\ref{subsec:task_specific_example_construction_pipeline}), each connected component starts from an undirected path of $3k-2$ papers. We sample these paths from the citation graph of all eligible papers, without starting from seed papers.

\paragraph{Citation graph.} A paper is eligible if it has a complete publication date on or after the cutoff date (April 1, 2026, in this study) and a non-empty title without MathML markup. We build an undirected graph whose nodes are the eligible papers and whose edges are the citation relationships between them.

\paragraph{Pruning.}  For question depth $k$, each path has $3k-2$ papers. Before sampling, we remove the papers that cannot lie on any simple path of $3k-2$ papers. Because finding the longest simple path is NP-hard, we use two upper bounds on the length of such a path: we remove connected components with fewer than $3k-2$ papers, and we use the biconnected components of the graph to bound the length of a path through each paper. We repeat both steps until no paper is removed.

\paragraph{Sampling.} We sample each path with a randomized depth-first search. We choose a start paper uniformly at random from the papers that can be an endpoint of a $(3k-2)$-paper path. At each step, we move to a neighboring paper chosen uniformly at random among those that are not on the path and can still complete a path of the required length. If no such neighbor remains, we backtrack to the previous paper on the path and try another neighbor. If a search does not find a path within 100,000 steps, we stop it and draw a new start paper. We keep only distinct paths, counting a path and its reverse as the same path, and we allow each paper to appear in at most 100 paths to reduce the overlap between paths. We discard papers without usable metadata and replace the paths that contain them. For each question depth, we sample a pool of 1,000 paths.

\paragraph{Paths for each example.} For each example, we draw the paths of its components uniformly at random from the pool of the required question depth, independently of other examples. The drawn paths must be disjoint: no two of them share a paper or a title, and no citation relationship connects them. Different examples may use the same path, but the components of one example never share a paper. We use each path in a random direction.

\subsubsection{Backbone Structure}
\label{appdx:initial_component_structure}

This part provides detail information about the initial structure of a component, which is described in \S\ref{subsec:task_specific_example_construction_pipeline}.

For question depth $k \geq 2$, the sampled path $(p_1, \ldots, p_{3k-2})$ of component $c$ forms its backbone, with source node $s = p_k$ and target node $t = p_{2k-1}$. The true path is $(p_k, \ldots, p_{2k-1})$, and the two distractor paths are $(p_1, \ldots, p_k)$, which ends at $s$, and $(p_{2k-1}, \ldots, p_{3k-2})$, which starts at $t$. Each of the three paths has $k$ nodes and $k-1$ edges, and the distractor paths share only $s$ and $t$ with the true path.

\subsubsection{Component Sampling}
\label{appdx:component_growth}

To mimic the scale-free, heavy-tailed degree distribution of citation graphs~\citep{zhong2024scale}, Step~2 (\S\ref{subsec:task_specific_example_construction_pipeline}) uses a sampling rule akin to the one-parameter Chinese restaurant process~\citep{aldous2006exchangeability}, which favors nodes that have already received children.

We sample all components of an example in one loop. Let $\sF_\ell$ be the set of nodes that are eligible to receive a new child at the $\ell$-th sampling step, over all components of the example. A node $v$ is eligible if (1) its component has not yet reached the component size of the example, and (2) $v$ still has untried neighboring papers that pass the publication date filter. At the first step, $\sF_0$ contains the eligible nodes of all backbones.

At each sampling step $\ell = 0, 1, \ldots$, let $g_\ell(v)$ be the number of children that node $v$ has received before step $\ell$; edges of the backbone do not count. We select a node $v \in \sF_\ell$ to receive a new child with probability
\begin{multline}
  P_\ell(v)=\frac{w_\ell(v)}{\sum_{u\in\sF_\ell}w_\ell(u)}, \\
  \text{where} \qquad
  w_\ell(v)=
  \begin{cases}
    \alpha, & g_\ell(v)=0,\\
    g_\ell(v), & g_\ell(v)>0,
  \end{cases}
  \label{eqn:preferential_growth}
\end{multline}
where $\alpha>0$ is the concentration parameter. Each node without children has its own weight $\alpha$, and smaller values of $\alpha$ favor nodes that have already received children.

Sampling stops when every component reaches the component size of the example. The additional components that Step~4 adds as distractors are sampled in the same way, and no two components of an example share a paper.

\subsubsection{Prompt Assembly}
\label{appdx:prompt_assembly}

This part details Step~4 of \S\ref{subsec:task_specific_example_construction_pipeline}.

\paragraph{One prompt per example for all models.}
Every evaluated model receives the same prompt for each example, so that all models are compared on the same input. However, the models use different tokenizers, so the same text has a different number of tokens for each model. We therefore treat the target context length $L$ as a limit on the whole prompt: the system prompt, the question, the context, the response-format reminder, and the tokens added by the chat template. A prompt satisfies this limit if it has at most $L$ tokens under each of the following tokenizers:

\begin{itemize}
  \item \model{Gemini}: \model{Gemma-4} uses the same tokenizer as \model{Gemini}~\citep{team2026gemma}, so we use the \model{Gemma-4} tokenizer to estimate the token counts of the \model{Gemini} models.
  \item \model{Qwen3.5}: the tokenizer shared by the evaluated \model{Qwen3.5} models~\citep{qwen3.5}.
  \item \model{GPT-5.6}: the \texttt{o200k\_base} encoding of \texttt{tiktoken}\footnote{\url{https://github.com/openai/tiktoken}}, with a fixed overhead of 3 tokens per message and 3 tokens per prompt.
\end{itemize}
We use the same three tokenizers for every target context length.

\paragraph{Filling the context.}
We build the prompt of each example as follows.
\begin{enumerate}
  \item If the needles alone exceed $L$ under any tokenizer, we construct the example again. We choose the component sizes and target context lengths so that this is unlikely.
  \item We draw distractors one at a time. A distractor is one source file, one essay, one book file, one Wikipedia passage, or one additional component. We split text distractors into blocks at blank lines, and we render an additional component as in Step~3, with one block for each paper that cites another paper in the component.
  \item Each needle and each distractor block receives an independent random number, and the context lists them in increasing order of these numbers, so the needles appear at random positions among the distractors.
  \item After each distractor, we count the tokens of the whole prompt with each tokenizer. Once the prompt reaches $L$, we keep the longest sequence of blocks, in the order they were drawn, for which the largest of the three counts is at most $L$. We stop when the next block does not fit, or when the prompt is within 100 tokens of $L$.
\end{enumerate}
The last distractor may therefore be cut. In that case, we keep its blocks from the beginning, in their original order, up to the last block that fits; we never split a block. For an additional component, this order is the order in which component sampling added its papers, so the papers added last are dropped. Because a block is never split, the prompt can fall short of $L$ by up to the size of the first block that does not fit.

The prompt is close to $L$ only under the tokenizer that gives the largest count, and it is shorter under the other two. For example, with \texttt{wikipedia} distractors and $L=200{,}000$, a prompt has about 189,000 tokens under the GPT tokenizer (\Cref{tab:v1_dataset_statistics}).

\subsubsection{Answer Balance}
\label{appdx:answer_balance}

Within each value of the tested variable, we go through the YES/NO examples in order and assign their labels in pairs, one YES and one NO in random order. Similarly, we assign the correct option letters to consecutive groups of four multiple-choice examples, one per letter in random order. Because the task of each example is sampled, the last pair or group may be incomplete (\Cref{tab:v1_answer_distribution}).

\subsection{Dataset Statistics}
\label{appdx:dataset_statistics}

\Cref{tab:v1_dataset_statistics} summarizes the evaluated examples of each experiment, and \Cref{tab:v1_answer_distribution} gives their ground truth answers.

\begin{table*}[t]
  \centering
  \small
  \setlength{\tabcolsep}{4pt}
  \begin{tabular}{@{}lrrrrrrr@{}}
    \toprule
    \textbf{Value} & \textbf{\#Examples} & \makecell[r]{\textbf{Question}\\\textbf{(words)}} & \makecell[r]{\textbf{Context}\\\textbf{(k words)}} & \makecell[r]{\textbf{Input}\\\textbf{(k tokens)}} & \textbf{Needles} & \makecell[r]{\textbf{Needle}\\\textbf{words (\%)}} & \textbf{\#Nodes} \\
    \midrule
    \multicolumn{8}{@{}l}{\textit{Question depth $k$}} \\
    \quad 4 & 100 & 72.65\textsubscript{$\pm$20.87} & 101.78\textsubscript{$\pm$25.55} & 174.79\textsubscript{$\pm$16.52} & 256.79\textsubscript{$\pm$111.50} & 27.30\textsubscript{$\pm$15.53} & 691.30\textsubscript{$\pm$508.34} \\
    \quad 8 & 100 & 73.55\textsubscript{$\pm$20.19} & 98.41\textsubscript{$\pm$28.68} & 171.80\textsubscript{$\pm$26.64} & 266.35\textsubscript{$\pm$129.73} & 30.85\textsubscript{$\pm$22.46} & 809.74\textsubscript{$\pm$586.53} \\
    \quad 16 & 100 & 68.76\textsubscript{$\pm$19.39} & 101.48\textsubscript{$\pm$26.39} & 171.55\textsubscript{$\pm$22.59} & 249.62\textsubscript{$\pm$128.76} & 28.52\textsubscript{$\pm$22.55} & 737.17\textsubscript{$\pm$566.70} \\
    \quad 32 & 100 & 68.30\textsubscript{$\pm$18.73} & 101.59\textsubscript{$\pm$27.15} & 169.96\textsubscript{$\pm$25.30} & 239.49\textsubscript{$\pm$108.88} & 27.86\textsubscript{$\pm$22.16} & 671.86\textsubscript{$\pm$525.55} \\
    \addlinespace
    \multicolumn{8}{@{}l}{\textit{Component size}} \\
    \quad 100 & 100 & 70.94\textsubscript{$\pm$20.85} & 102.27\textsubscript{$\pm$27.98} & 171.38\textsubscript{$\pm$17.52} & 145.67\textsubscript{$\pm$52.83} & 15.41\textsubscript{$\pm$7.61} & 530.73\textsubscript{$\pm$565.42} \\
    \quad 150 & 100 & 73.72\textsubscript{$\pm$20.01} & 99.19\textsubscript{$\pm$29.36} & 171.15\textsubscript{$\pm$25.48} & 222.87\textsubscript{$\pm$84.02} & 26.69\textsubscript{$\pm$19.53} & 631.39\textsubscript{$\pm$509.17} \\
    \quad 200 & 100 & 69.49\textsubscript{$\pm$19.73} & 92.56\textsubscript{$\pm$28.02} & 168.00\textsubscript{$\pm$25.78} & 282.40\textsubscript{$\pm$103.96} & 35.14\textsubscript{$\pm$21.77} & 777.08\textsubscript{$\pm$519.18} \\
    \quad 250 & 100 & 71.59\textsubscript{$\pm$18.23} & 95.69\textsubscript{$\pm$29.42} & 166.83\textsubscript{$\pm$33.27} & 367.29\textsubscript{$\pm$133.00} & 45.21\textsubscript{$\pm$26.84} & 797.78\textsubscript{$\pm$412.91} \\
    \addlinespace
    \multicolumn{8}{@{}l}{\textit{Task}} \\
    \quad YES/NO & 100 & 54.29\textsubscript{$\pm$6.09} & 104.77\textsubscript{$\pm$26.06} & 175.14\textsubscript{$\pm$17.60} & 183.88\textsubscript{$\pm$54.32} & 19.61\textsubscript{$\pm$9.87} & 634.58\textsubscript{$\pm$565.27} \\
    \quad Multiple choice & 100 & 97.24\textsubscript{$\pm$9.92} & 86.75\textsubscript{$\pm$27.44} & 163.65\textsubscript{$\pm$32.74} & 387.19\textsubscript{$\pm$112.43} & 50.41\textsubscript{$\pm$24.38} & 856.47\textsubscript{$\pm$409.24} \\
    \quad Counting & 100 & 63.93\textsubscript{$\pm$3.62} & 97.04\textsubscript{$\pm$29.08} & 171.72\textsubscript{$\pm$22.95} & 189.88\textsubscript{$\pm$53.94} & 22.70\textsubscript{$\pm$14.51} & 671.65\textsubscript{$\pm$587.54} \\
    \addlinespace
    \multicolumn{8}{@{}l}{\textit{Distractor type}} \\
    \quad \texttt{books} & 100 & 71.40\textsubscript{$\pm$20.28} & 113.84\textsubscript{$\pm$10.71} & 176.48\textsubscript{$\pm$9.13} & 243.38\textsubscript{$\pm$110.58} & 22.23\textsubscript{$\pm$12.76} & 438.84\textsubscript{$\pm$197.92} \\
    \quad \texttt{code} & 100 & 70.32\textsubscript{$\pm$19.97} & 59.09\textsubscript{$\pm$7.23} & 155.13\textsubscript{$\pm$8.25} & 257.72\textsubscript{$\pm$121.82} & 41.95\textsubscript{$\pm$14.94} & 468.08\textsubscript{$\pm$213.22} \\
    \quad \makecell[l]{\texttt{disjoint-}\\\texttt{component}} & 100 & 70.30\textsubscript{$\pm$19.14} & 90.93\textsubscript{$\pm$1.85} & 186.08\textsubscript{$\pm$1.26} & 232.10\textsubscript{$\pm$90.69} & 25.71\textsubscript{$\pm$11.15} & 1,674.75\textsubscript{$\pm$92.83} \\
    \quad \texttt{essays} & 100 & 73.09\textsubscript{$\pm$19.53} & 95.58\textsubscript{$\pm$40.39} & 138.95\textsubscript{$\pm$42.07} & 268.15\textsubscript{$\pm$127.40} & 44.38\textsubscript{$\pm$40.25} & 482.44\textsubscript{$\pm$218.73} \\
    \quad \texttt{wikipedia} & 100 & 75.73\textsubscript{$\pm$19.46} & 121.72\textsubscript{$\pm$5.80} & 188.70\textsubscript{$\pm$0.61} & 280.18\textsubscript{$\pm$113.73} & 23.08\textsubscript{$\pm$10.63} & 502.76\textsubscript{$\pm$200.20} \\
    \addlinespace
    \multicolumn{8}{@{}l}{\textit{Reasoning effort}} \\
    \quad \texttt{low}/\texttt{medium}/\texttt{high} & 100 & 73.76\textsubscript{$\pm$20.55} & 99.23\textsubscript{$\pm$27.98} & 171.39\textsubscript{$\pm$19.43} & 262.73\textsubscript{$\pm$126.45} & 29.72\textsubscript{$\pm$18.57} & 628.93\textsubscript{$\pm$464.16} \\
    \addlinespace
    \multicolumn{8}{@{}l}{\textit{Target context length}} \\
    \quad 200k & 100 & 73.76\textsubscript{$\pm$20.55} & 99.23\textsubscript{$\pm$27.98} & 171.39\textsubscript{$\pm$19.43} & 262.73\textsubscript{$\pm$126.45} & 29.72\textsubscript{$\pm$18.57} & 628.93\textsubscript{$\pm$464.16} \\
    \quad 500k & 50 & 73.22\textsubscript{$\pm$21.71} & 262.21\textsubscript{$\pm$82.65} & 442.57\textsubscript{$\pm$37.30} & 273.32\textsubscript{$\pm$132.35} & 12.15\textsubscript{$\pm$7.96} & 1,150.66\textsubscript{$\pm$1,453.65} \\
    \quad 922k & 50 & 74.48\textsubscript{$\pm$20.52} & 490.65\textsubscript{$\pm$165.54} & 821.01\textsubscript{$\pm$70.67} & 268.92\textsubscript{$\pm$127.14} & 6.68\textsubscript{$\pm$4.81} & 1,798.20\textsubscript{$\pm$2,843.10} \\
    \midrule
    \textbf{Overall} & 1,760 & 71.95\textsubscript{$\pm$19.92} & 113.48\textsubscript{$\pm$81.37} & 196.32\textsubscript{$\pm$119.39} & 256.72\textsubscript{$\pm$121.56} & 29.35\textsubscript{$\pm$22.22} & 752.38\textsubscript{$\pm$762.13} \\
    \bottomrule
  \end{tabular}
  \caption{Statistics of the evaluated examples. Each row is one value of the tested variable of an experiment; the reasoning effort experiment uses the same examples for all three efforts, and these are also the 200k examples of the target context length experiment. \textbf{Overall} counts each distinct example once. Cells show mean\textsubscript{$\pm$std} over examples. Words are whitespace-separated. Context covers the needles and distractors, without the question and instructions. Input tokens cover the whole prompt, including the system prompt, counted with the GPT tokenizer (\texttt{o200k\_base}). Needles is the number of needles, one for each paper of the components generated for the task that cites another paper in its component; Needle words is the percentage of context words that belong to needles. Nodes is the number of papers of the subgraph that appear in the context, including the additional components of \texttt{disjoint-component}. \Benchmark is an evaluation benchmark and has no training or development split.}
  \label{tab:v1_dataset_statistics}
\end{table*}

\begin{table*}[t]
  \centering
  \small
  \begin{tabular}{@{}lrrrrrrrr@{}}
    \toprule
    & & \multicolumn{2}{c}{\textbf{YES/NO}} & \multicolumn{4}{c}{\textbf{Multiple choice}} & \textbf{Counting} \\
    \cmidrule(lr){3-4} \cmidrule(lr){5-8} \cmidrule(l){9-9}
    \textbf{Value} & \textbf{\#Examples} & \textbf{YES} & \textbf{NO} & \textbf{A} & \textbf{B} & \textbf{C} & \textbf{D} & \textbf{\#Examples} \\
    \midrule
    \multicolumn{9}{@{}l}{\textit{Question depth $k$}} \\
    \quad 4 & 100 & 17 & 17 & 8 & 9 & 8 & 9 & 32 \\
    \quad 8 & 100 & 14 & 14 & 9 & 10 & 10 & 9 & 34 \\
    \quad 16 & 100 & 20 & 20 & 7 & 8 & 7 & 7 & 31 \\
    \quad 32 & 100 & 21 & 22 & 6 & 7 & 6 & 7 & 31 \\
    \addlinespace
    \multicolumn{9}{@{}l}{\textit{Component size}} \\
    \quad 100 & 100 & 19 & 19 & 8 & 8 & 8 & 9 & 29 \\
    \quad 150 & 100 & 14 & 14 & 8 & 9 & 8 & 9 & 38 \\
    \quad 200 & 100 & 21 & 21 & 7 & 7 & 7 & 7 & 30 \\
    \quad 250 & 100 & 14 & 14 & 8 & 8 & 8 & 8 & 40 \\
    \addlinespace
    \multicolumn{9}{@{}l}{\textit{Task}} \\
    \quad YES/NO & 100 & 50 & 50 & -- & -- & -- & -- & -- \\
    \quad Multiple choice & 100 & -- & -- & 25 & 25 & 25 & 25 & -- \\
    \quad Counting & 100 & -- & -- & -- & -- & -- & -- & 100 \\
    \addlinespace
    \multicolumn{9}{@{}l}{\textit{Distractor type}} \\
    \quad \texttt{books} & 100 & 20 & 19 & 8 & 7 & 8 & 7 & 31 \\
    \quad \texttt{code} & 100 & 16 & 16 & 7 & 7 & 7 & 7 & 40 \\
    \quad \makecell[l]{\texttt{disjoint-}\\\texttt{component}} & 100 & 16 & 16 & 7 & 7 & 7 & 7 & 40 \\
    \quad \texttt{essays} & 100 & 15 & 16 & 9 & 9 & 9 & 9 & 33 \\
    \quad \texttt{wikipedia} & 100 & 11 & 12 & 11 & 11 & 11 & 11 & 33 \\
    \addlinespace
    \multicolumn{9}{@{}l}{\textit{Reasoning effort}} \\
    \quad \texttt{low}/\texttt{medium}/\texttt{high} & 100 & 14 & 14 & 9 & 8 & 8 & 8 & 39 \\
    \addlinespace
    \multicolumn{9}{@{}l}{\textit{Target context length}} \\
    \quad 200k & 100 & 14 & 14 & 9 & 8 & 8 & 8 & 39 \\
    \quad 500k & 50 & 6 & 6 & 4 & 5 & 5 & 4 & 20 \\
    \quad 922k & 50 & 6 & 6 & 4 & 5 & 5 & 4 & 20 \\
    \midrule
    \textbf{Overall} & 1,760 & 288 & 287 & 144 & 147 & 145 & 146 & 603 \\
    \bottomrule
  \end{tabular}
  \caption{Ground truth answers of the evaluated examples in Table~\ref{tab:v1_dataset_statistics}: the number of YES-label and NO-label examples, of multiple-choice examples whose correct option has each letter, and of counting examples. Within each value of the tested variable, YES/NO labels are assigned in pairs and option letters in groups of four (Appendix~\ref{appdx:answer_balance}). The ground truth count of a counting example is its component size minus one. A dash marks a task that the row does not contain.}
  \label{tab:v1_answer_distribution}
\end{table*}

\subsection{Question Templates}
\label{appdx:question_templates}

\Cref{fig:question_templates} shows the question templates of the three tasks; the system and user prompts that contain them are in Appendix~\ref{appdx:prompt_templates}. The multiple-choice template has four candidate papers. When the none option is included in the multiple-choice question, the last entry in the option list is \texttt{E. None of these targets is reachable from the source.}

\begin{figure*}[htbp]
  \begin{tcblisting}{
      templatebox,
      title={Question Template for YES/NO reachability}
    }
    Using only the relationships described in the context, can you start from the Paper "{Title of the source paper}" and arrive at the Paper "{Title of the target paper}" by following one or more `cites' or `cited_by' relationships?
  \end{tcblisting}

  \begin{tcblisting}{
      templatebox,
      title={Question Template for Multiple-choice Reachability}
    }
    Using only the relationships described in the context, which of the following is reachable from Paper "{Title of the source paper}" by following one or more `cites' or `cited_by' relationships?

    A. Paper "{Title of the paper in option A}"
    B. Paper "{Title of the paper in option B}"
    C. Paper "{Title of the paper in option C}"
    D. Paper "{Title of the paper in option D}"
  \end{tcblisting}

  \begin{tcblisting}{
      templatebox,
      title={Question Template for Reachability Counting},
      listing options={
        basicstyle=\ttfamily\small,
        breaklines=true,
        columns=fullflexible,
        keepspaces=true,
        showstringspaces=false,
        breakindent=0pt,
        breakatwhitespace=true,
        literate={’}{{'}}1
      }
    }
    Using only the relationships described in the context, how many distinct papers are reachable from Paper "{Title of the source paper}" by following one or more `cites' or `cited_by' relationships? Count each reachable paper once, whether it has its own paper block or appears only in another paper's reference list. Do not count the starting paper.
  \end{tcblisting}
  \caption{Question templates for the three reachability tasks. Placeholders in braces denote example-specific values.}
  \label{fig:question_templates}
\end{figure*}

\subsection{APA-Style Template for Papers}
\label{appdx:needle_templates}

\Cref{fig:question_templates} shows the question templates of the three tasks; the system and user prompts that contain them are in Appendix~\ref{appdx:prompt_templates}. The multiple-choice template has four candidate papers. For the negative option test of the multiple-choice reachability task in \S\ref{subsec:performance_on_different_reachability_tasks}, we append a fifth option, \texttt{E. None of these targets is reachable from the source.}, after option D and keep the rest of the question unchanged.

\begin{figure}
  \begin{tcblisting}{
      templatebox,
      title={APA-Style Template}
    }
    Title: {Title of the paper}.
    Authors: {List of author names}.
    Published in {Venue of the paper} on {Publication date}.
    References:
    - {List of author names of the 1st reference paper} ({Publication year}). {Title of the 1st reference paper}. {Source of the 1st reference paper}.
    - {List of author names of the 2nd reference paper} ({Publication year}). {Title of the 2nd reference paper}. {Source of the 2nd reference paper}.
    ...
  \end{tcblisting}
  \caption{APA-style template used to render each paper. Placeholders in braces denote example-specific values.}
  \label{fig:paper_block_template}
\end{figure}

Dates are rendered as a month name, day, and year (e.g., September 1, 2025). For an arXiv paper without another publication venue, the publication information line \texttt{Published in \{Venue of the paper\} on \{Publication date\}.} is replaced with \texttt{Uploaded to arXiv on \{Publication date\}.}

In the \texttt{References} part, a reference source is a publication venue, \texttt{arXiv:\{arXiv ID\}}, or \texttt{arXiv}, depending on available metadata. Missing metadata is omitted. For example, a publication line may contain only the venue or date, or may be omitted entirely. Similar to the main block, Reference entries omit unavailable authors, year, or source.

As in Step~3, a paper with no selected outgoing citation has no block of its own and appears only in the reference lists of the papers that cite it. Such a paper is still a node of the subgraph, so the model must include it when it decides reachability or counts reachable papers.

\section{Selection of the Concentration Parameter}
\label{appdx:selection_of_sampling_parameter}

Step~2 aims to mimic the degree distribution of citation graphs (\S\ref{subsec:task_specific_example_construction_pipeline}). We therefore select the concentration parameter $\alpha$ in Equation~\ref{eqn:preferential_growth} by comparing the in-degree distribution across generated examples with the in-degree distribution of the filtered Semantic Scholar citation graph, which contains the papers published on or after the cutoff date (April 1, 2026, in this study). We measure the difference with the Kullback-Leibler (KL) divergence~\citep{kullback1951information} from the first distribution to the second.

\subsection{Selection Setup}

We test ten values, $\alpha \in \mathcal{A} = \{0.1, 0.2, \ldots, 1.0\}$. For each value, we generate 2,000 examples of each task, which gives 6,000 examples per value and 60,000 examples in total. We sample the question depth and component size of each example uniformly at random, as in the experiments (\S\ref{subsec:experimental_setup}), and use a target context length of 200,000 tokens. We add no distractors to these examples, because they make the search slower and four of the five distractor types add no papers to the subgraph.

\subsection{Degree Distribution Calculation}

An edge $u\to v$ means that paper $u$ cites paper $v$. The in-degree of paper $v$ is the number of papers that cite it, i.e., $d_{\mathrm{in}}(v)=|\{u:u\to v\}|$. For a generated example, we count only the citation edges inside its subgraph; for the filtered Semantic Scholar citation graph, we count only the citation edges between papers published on or after the cutoff date.

In both distributions, we keep only in-degree values of at least 1 and normalize the remaining counts. This filtering affects only the comparison, not the generation of examples. We select this in-degree filtering because all papers in the subgraph must have at least one citation edge (due to the nature of the component sampling procedure). Hence, we only consider degrees greater than or equal to 1 to avoid the mismatch between the two distributions at low degrees.

For a fixed concentration parameter $\alpha$ and reachability task $\tau$, let $h_i(a)$ count papers with in-degree $a$ in example $i$; we pool these counts and normalize over degrees greater than or equal to 1 to obtain the conditional degree distribution across generated examples:
\begin{equation}
  P_{\alpha, \tau}(a)=\frac{H_{\alpha, \tau}(a)}{\sum_{a'\geq 1} H_{\alpha, \tau}(a')}
  \label{eqn:alpha_pooled_distribution}
\end{equation}
where $H_{\alpha, \tau}(a)=\sum_i h_i(a)$ and $a \geq 1$.

Let $H^*(a)$ count papers with in-degree $a$ in the filtered Semantic Scholar citation graph. Its conditional degree distribution is
\begin{equation}
  Q(a)=\frac{H^*(a)}{\sum_{a'\geq 1} H^*(a')},\quad a\geq 1.
  \label{eqn:alpha_reference_distribution}
\end{equation}

\subsection{Selection of the Concentration Parameter with KL Divergence}

We form a common support $\mathcal S$ from all in-degree values of at least 1 observed in the filtered Semantic Scholar citation graph or across generated examples for any candidate value of $\alpha$ and any reachability task. Unobserved degree values within this support receive zero probability. Let $M=|\mathcal S|$ denote the number of support values.  To prevent the KL divergence from becoming infinite when a degree value has zero probability, we smooth both conditional distributions with $\varepsilon=10^{-8}$
\begin{equation}
  \begin{split}
    \widetilde P_{\alpha, \tau}(a)&=(1-\varepsilon)P_{\alpha, \tau}(a)+\frac{\varepsilon}{M}, \\
    \widetilde Q(a)&=(1-\varepsilon)Q(a)+\frac{\varepsilon}{M}.
  \end{split}
  \label{eqn:alpha_smoothing}
\end{equation}

For each value $\alpha$ and task $\tau$, we calculate KL divergence from the conditional degree distribution across generated examples to that of the filtered Semantic Scholar citation graph:
\begin{equation}
  D_{\alpha,\tau}
  =\sum_{a\in\mathcal S} \widetilde P_{\alpha,\tau}(a) \cdot \log\frac{\widetilde P_{\alpha,\tau}(a)}{\widetilde Q(a)}.
  \label{eqn:alpha_kl_divergence}
\end{equation}

We average the three KL divergences, one for each task, to obtain the mean KL divergence of each candidate:
\begin{equation}
  S(\alpha)=\frac{1}{3}\sum_\tau D_{\alpha,\tau}
  \label{eqn:alpha_selection_score}
\end{equation}

Finally, we select the value of $\alpha$ with the minimum mean KL divergence:
\begin{equation}
  \alpha^\star=\argmin_{\alpha\in\mathcal A}S(\alpha).
  \label{eqn:theta_discount_alpha_selection}
\end{equation}

\subsection{Concentration Parameter Selection Results}

\begin{figure*}[tp]
  \centering
  \includegraphics[width=\textwidth]{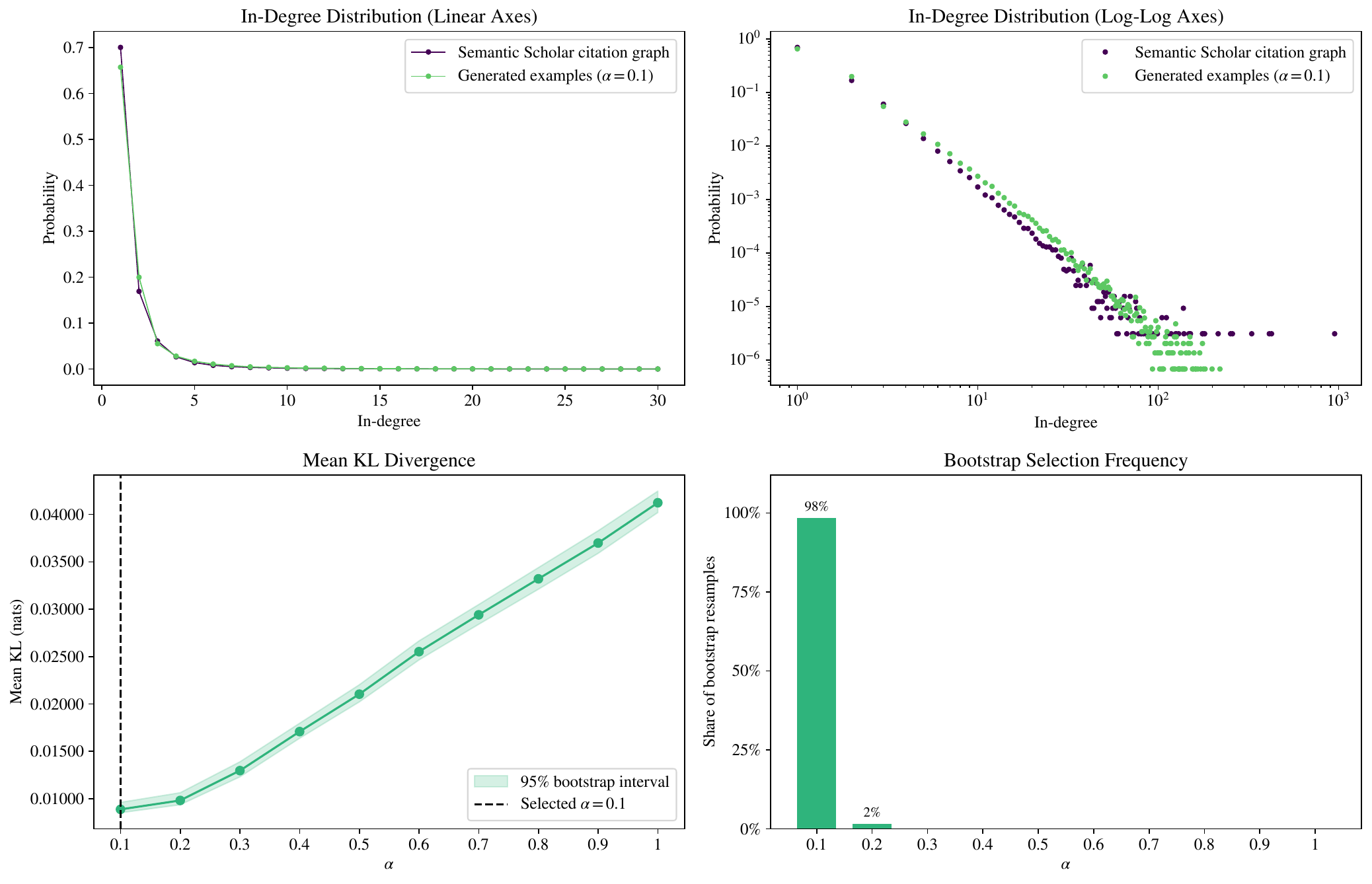}
  \caption{Concentration parameter selection results. Top: In-degree distributions of the filtered Semantic Scholar citation graph and generated examples using the selected value $\alpha^\star=0.1$, shown on linear (left) and logarithmic (right) axes. Bottom: Mean KL divergence $S(\alpha)$ with 95\% bootstrap intervals (left) and bootstrap selection frequencies (right).}
  \label{fig:sampling_parameter_selection_results}
\end{figure*}

Among the ten values, $\alpha^\star=0.1$ gives the smallest mean KL divergence, $S(0.1)=0.0089$ (Figure~\ref{fig:sampling_parameter_selection_results}), so we use it to generate the examples of all experiments (\S\ref{subsec:experimental_setup}). To check whether this choice is stable, we draw 500 bootstrap replicates, resampling the examples with replacement within each task and value of $\alpha$, and report pointwise 95\% percentile bootstrap intervals. The value $\alpha=0.1$ is selected in 98.4\% of the replicates, and $\alpha=0.2$ in the remaining 1.6\%.

\section{Models and Inference Settings}
\label{appdx:models_and_inference_settings}

Table~\ref{tab:model_specs} lists the evaluated models. We evaluate all models with $\texttt{temperature}=1.0$ and $\texttt{reasoning\_effort}=\texttt{low}$; the reasoning effort experiment (\S\ref{subsec:effects_of_reasoning_effort}) also uses \texttt{medium} and \texttt{high}. We set $\texttt{top\_p}=0.95$ for every model except the \model{GPT-5.6} models, which do not accept $\texttt{top\_p}$ while reasoning is enabled. We set no limit on the number of output tokens, except for the \model{Qwen3.5} models. We call all models through LiteLLM\footnote{\url{https://github.com/BerriAI/litellm}}.

We send the \model{GPT-5.6} and \model{Gemini} requests at the flex service tier of OpenAI and resend a request at the standard tier when the flex tier has no capacity.

For the \model{Qwen3.5} models, we add $\texttt{top\_k}=20$, $\texttt{min\_p}=0$, and $\texttt{presence\_penalty}=1.5$, the values that Qwen recommends for thinking mode on general tasks, and allow at most 40,000 output tokens. We access these models through OpenRouter, with Alibaba as the first provider and DeepInfra (FP8 weights) as the second.

\begin{table*}[htbp]
  \centering
  \resizebox{\textwidth}{!}{%
    \begin{tabular}{@{}llrrrll@{}}
      \toprule
      \textbf{Model} & \textbf{\#Params} & \textbf{Release date} & \textbf{Cutoff date} & \multicolumn{1}{r}{\textbf{Context length}} & \textbf{I/O pricing} & \textbf{Provider} \\
      \midrule
      \midrule
      \multicolumn{7}{c}{\textbf{Proprietary Models}} \\
      \midrule
      \model{GPT-5.6-Sol}           & N/A               & Jul. 9, 2026  & Feb. 16, 2026 & 1,050,000 & 4.00 / 20.00 & OpenAI \\
      \model{GPT-5.6-Luna}          & N/A               & Jul. 9, 2026  & Feb. 16, 2026 & 1,050,000 & 0.20 / 1.20  & OpenAI \\
      \model{Gemini-3.8-Flash}      & N/A               & Sep. 2, 2026  & Mar. 2026     & 1,048,576 & 0.75 / 3.75  & Google \\
      \model{Gemini-3.6-Flash}      & N/A               & Jul. 21, 2026 & Mar. 2026     & 1,048,576 & 0.75 / 3.75  & Google \\
      \model{Gemini-3.1-Pro} & N/A              & Feb. 19, 2026 & Jan. 2025    & 1,048,576 & 2.00 / 12.00 & Google \\
      \midrule
      \multicolumn{7}{c}{\textbf{Open-source Models}} \\
      \midrule
      \model{Qwen3.5-397B-A17B}     & 397B (17B active) & Feb. 16, 2026 & Feb. 16, 2026 &   262,144 & 0.39 / 2.34 & OpenRouter \\
      \model{Qwen3.5-122B-A10B}     & 122B (10B active) & Feb. 24, 2026 & Feb. 24, 2026 &   262,144 & 0.26 / 2.08 & OpenRouter \\
      \model{Qwen3.5-35B-A3B}       & 35B (3B active)   & Feb. 24, 2026 & Feb. 24, 2026 &   262,144 & 0.16 / 1.30 & OpenRouter \\
      \bottomrule
    \end{tabular}%
  }
  \caption{Model specifications. Qwen has not published a training cutoff date for the \model{Qwen3.5} models, so we use their release dates as their cutoff dates. I/O pricing is the standard price in US dollars per million input and output tokens; for the \model{Qwen3.5} models, it is the price of Alibaba, the first provider in our OpenRouter routing order. For \model{Gemini-3.1-Pro}, the listed price applies to prompts of at most 200,000 tokens;above this length, the standard price is 4.00 / 18.00. Its flex prices are 1.00 / 6.00 up to 200,000 tokens and 2.00 / 9.00 above. N/A: not published.}
  \label{tab:model_specs}
\end{table*}

\section{Random Guessing Baseline}
\label{appdx:random_guessing_baseline}

We compare each model with random guessing, which ignores the context and chooses an answer uniformly at random from the possible answers of the task. We compute its expected accuracy exactly from the sampling probabilities of the benchmark construction pipeline, rather than measuring it on the evaluated examples, so the baseline does not depend on which examples were drawn.

\paragraph{Accuracy on one task.}
Let $\sA_\tau$ denote the set of possible answers of task $\tau$. For any ground truth answer $y \in \sA_\tau$, a uniform random guess equals $y$ with probability $1/\lvert\sA_\tau\rvert$. This probability is the same for every $y$, so the expected accuracy of random guessing on task $\tau$ is $\mathrm{Acc}_\tau = 1/\lvert\sA_\tau\rvert$ for any distribution of ground truth answers:
\begin{itemize}
  \item \textbf{YES/NO reachability:} $\sA_\tau=\{\texttt{YES}, \texttt{NO}\}$, so $\mathrm{Acc}_\tau = 1/2 = 50.00\%$.
  \item \textbf{Multiple-choice reachability:} four options, so $\mathrm{Acc}_\tau = 1/4 = 25.00\%$.
  \item \textbf{Reachability counting:} the component size is an integer in $[100, 250]$ and $y_{\mathrm{count}} = \lvert\sK_s\rvert - 1$, so $\sA_\tau$ contains the 151 integers from 99 to 249 and $\mathrm{Acc}_\tau = 1/151 \approx 0.66\%$.
\end{itemize}
We use these answer sets in every experiment.

\paragraph{Accuracy on a mix of tasks.}
In every experiment except the task experiment, the task of each example is sampled uniformly from the three tasks. By the law of total probability, the expected accuracy of random guessing is
\begin{equation}
  \begin{split}
    \mathrm{Acc}_{\mathrm{mix}}
    &= \sum_{\tau} \Pr(\tau)\,\mathrm{Acc}_\tau \\
    &= \frac{1}{3}\left(\frac{1}{2} + \frac{1}{4} + \frac{1}{151}\right) \\
    &\approx 25.22\%.
  \end{split}
  \label{eqn:random_guessing_mix}
\end{equation}

\paragraph{Task experiment.}
In the task experiment, each value of the tested variable is one task, so the baseline for each value is $\mathrm{Acc}_\tau$: 50.00\%, 25.00\%, and 0.66\%. Pooled over the three tasks, which have the same number of examples, the baseline is again 25.22\%.

\section{Extra Experimental Results}
\label{appdx:extra_experimental_results}

This section breaks the results of \S\ref{sec:experiments_and_analysis} down by task and by ground truth answer, and it reports the statistical tests used in the analysis.

\subsection{Complexity Scaling}

\Cref{tab:v1_accuracy_by_question_depth_and_yesno,tab:v1_accuracy_by_question_depth_and_multiple,tab:v1_accuracy_by_question_depth_and_counting,tab:v1_accuracy_by_node_count_and_yesno,tab:v1_accuracy_by_node_count_and_multiple,tab:v1_accuracy_by_node_count_and_counting} split the accuracy of the question depth and component size experiments (\S\ref{subsec:complexity_scaling}) by task. \Cref{tab:v1_yesno_accuracy_by_question_depth_and_label,tab:v1_yesno_accuracy_by_node_count_and_label} further split their YES/NO reachability accuracy into YES-label and NO-label examples.

\begin{table*}[t]
  \centering
  \small
  \setlength{\tabcolsep}{5pt}
  \begin{tabular}{@{}lccccc@{}}
    \toprule
    & \multicolumn{4}{c}{\textbf{Question depth $k$}} & \\
    \cmidrule(lr){2-5}
    \textbf{Model} & \textbf{4} & \textbf{8} & \textbf{16} & \textbf{32} & \textbf{Overall} \\
    \midrule
    \model{GPT-5.6-Sol} & -- & \makecell{96.43\\{\scriptsize[82.29, 99.37]}} & \makecell{57.50\\{\scriptsize[42.19, 71.49]}} & \makecell{53.49\\{\scriptsize[38.92, 67.49]}} & \makecell{65.77$^\dagger$\\{\scriptsize[56.54, 73.93]}} \\
    \addlinespace[2pt]
    \model{GPT-5.6-Luna} & \makecell{94.12\\{\scriptsize[80.91, 98.37]}} & \makecell{50.00\\{\scriptsize[32.63, 67.37]}} & \makecell{50.00\\{\scriptsize[35.20, 64.80]}} & \makecell{51.16\\{\scriptsize[36.75, 65.38]}} & \makecell{60.69\\{\scriptsize[52.56, 68.27]}} \\
    \addlinespace[2pt]
    \model{Gemini-3.1-Pro} & \makecell{\textbf{100.00}\\{\scriptsize[89.85, 100.00]}} & \makecell{89.29\\{\scriptsize[72.80, 96.29]}} & \makecell{67.50\\{\scriptsize[52.02, 79.92]}} & \makecell{\textbf{67.44}\\{\scriptsize[52.52, 79.51]}} & \makecell{79.31\\{\scriptsize[72.00, 85.11]}} \\
    \addlinespace[2pt]
    \model{Gemini-3.8-Flash} & \makecell{97.06\\{\scriptsize[85.08, 99.48]}} & \makecell{\textbf{100.00}\\{\scriptsize[87.94, 100.00]}} & \makecell{\textbf{85.00}\\{\scriptsize[70.93, 92.94]}} & \makecell{\textbf{67.44}\\{\scriptsize[52.52, 79.51]}} & \makecell{\textbf{85.52}\\{\scriptsize[78.87, 90.33]}} \\
    \addlinespace[2pt]
    \model{Gemini-3.6-Flash} & \makecell{\textbf{100.00}\\{\scriptsize[89.85, 100.00]}} & \makecell{82.14\\{\scriptsize[64.41, 92.12]}} & \makecell{67.50\\{\scriptsize[52.02, 79.92]}} & \makecell{58.14\\{\scriptsize[43.33, 71.62]}} & \makecell{75.17\\{\scriptsize[67.55, 81.49]}} \\
    \addlinespace[2pt]
    \model{Qwen3.5-397B-A17B} & \makecell{94.12\\{\scriptsize[80.91, 98.37]}} & \makecell{71.43\\{\scriptsize[52.94, 84.75]}} & \makecell{50.00\\{\scriptsize[35.20, 64.80]}} & \makecell{51.16\\{\scriptsize[36.75, 65.38]}} & \makecell{64.83\\{\scriptsize[56.76, 72.13]}} \\
    \addlinespace[2pt]
    \model{Qwen3.5-122B-A10B} & \makecell{94.12\\{\scriptsize[80.91, 98.37]}} & \makecell{60.71\\{\scriptsize[42.41, 76.43]}} & \makecell{50.00\\{\scriptsize[35.20, 64.80]}} & \makecell{51.16\\{\scriptsize[36.75, 65.38]}} & \makecell{62.76\\{\scriptsize[54.66, 70.20]}} \\
    \addlinespace[2pt]
    \model{Qwen3.5-35B-A3B} & \makecell{88.24\\{\scriptsize[73.38, 95.33]}} & \makecell{57.14\\{\scriptsize[39.07, 73.49]}} & \makecell{50.00\\{\scriptsize[35.20, 64.80]}} & \makecell{55.81\\{\scriptsize[41.11, 69.57]}} & \makecell{62.07\\{\scriptsize[53.96, 69.56]}} \\[3pt]
    \midrule
    Random guessing & 50.00 & 50.00 & 50.00 & 50.00 & 50.00 \\
    \midrule
    \#Examples & 34 & 28 & 40 & 43 & 145 \\
    \bottomrule
  \end{tabular}
  \caption{Accuracy (\%) of each model on the YES/NO reachability examples of the question depth experiment (\S\ref{subsec:complexity_scaling}), with reasoning effort \texttt{low}. Overall pools all question depths. The number of examples differs between columns because the task of each example is sampled at random. Below each accuracy value is its 95\% Wilson confidence interval. Bold marks the highest accuracy in each column. $^\dagger$\model{GPT-5.6-Sol} has fewer examples than the other models in this cell (111 Overall), so it is not bold. A dash marks a missing result.}
  \label{tab:v1_accuracy_by_question_depth_and_yesno}
\end{table*}

\begin{table*}[t]
  \centering
  \small
  \setlength{\tabcolsep}{5pt}
  \begin{tabular}{@{}lccccc@{}}
    \toprule
    & \multicolumn{4}{c}{\textbf{Question depth $k$}} & \\
    \cmidrule(lr){2-5}
    \textbf{Model} & \textbf{4} & \textbf{8} & \textbf{16} & \textbf{32} & \textbf{Overall} \\
    \midrule
    \model{GPT-5.6-Sol} & -- & \makecell{\textbf{100.00}\\{\scriptsize[90.82, 100.00]}} & \makecell{86.21\\{\scriptsize[69.44, 94.50]}} & \makecell{69.23\\{\scriptsize[50.01, 83.50]}} & \makecell{87.10$^\dagger$\\{\scriptsize[78.79, 92.46]}} \\
    \addlinespace[2pt]
    \model{GPT-5.6-Luna} & \makecell{\textbf{100.00}\\{\scriptsize[89.85, 100.00]}} & \makecell{65.79\\{\scriptsize[49.89, 78.79]}} & \makecell{58.62\\{\scriptsize[40.74, 74.49]}} & \makecell{46.15\\{\scriptsize[28.76, 64.54]}} & \makecell{69.29\\{\scriptsize[60.80, 76.65]}} \\
    \addlinespace[2pt]
    \model{Gemini-3.1-Pro} & \makecell{\textbf{100.00}\\{\scriptsize[89.85, 100.00]}} & \makecell{92.11\\{\scriptsize[79.20, 97.28]}} & \makecell{89.66\\{\scriptsize[73.61, 96.42]}} & \makecell{65.38\\{\scriptsize[46.22, 80.59]}} & \makecell{88.19\\{\scriptsize[81.43, 92.71]}} \\
    \addlinespace[2pt]
    \model{Gemini-3.8-Flash} & \makecell{\textbf{100.00}\\{\scriptsize[89.85, 100.00]}} & \makecell{\textbf{100.00}\\{\scriptsize[90.82, 100.00]}} & \makecell{\textbf{100.00}\\{\scriptsize[88.30, 100.00]}} & \makecell{\textbf{80.77}\\{\scriptsize[62.12, 91.49]}} & \makecell{\textbf{96.06}\\{\scriptsize[91.11, 98.31]}} \\
    \addlinespace[2pt]
    \model{Gemini-3.6-Flash} & \makecell{\textbf{100.00}\\{\scriptsize[89.85, 100.00]}} & \makecell{73.68\\{\scriptsize[57.99, 85.03]}} & \makecell{62.07\\{\scriptsize[44.00, 77.31]}} & \makecell{46.15\\{\scriptsize[28.76, 64.54]}} & \makecell{72.44\\{\scriptsize[64.10, 79.47]}} \\
    \addlinespace[2pt]
    \model{Qwen3.5-397B-A17B} & \makecell{97.06\\{\scriptsize[85.08, 99.48]}} & \makecell{73.68\\{\scriptsize[57.99, 85.03]}} & \makecell{75.86\\{\scriptsize[57.89, 87.78]}} & \makecell{61.54\\{\scriptsize[42.53, 77.57]}} & \makecell{77.95\\{\scriptsize[69.98, 84.28]}} \\
    \addlinespace[2pt]
    \model{Qwen3.5-122B-A10B} & \makecell{97.06\\{\scriptsize[85.08, 99.48]}} & \makecell{73.68\\{\scriptsize[57.99, 85.03]}} & \makecell{68.97\\{\scriptsize[50.77, 82.72]}} & \makecell{57.69\\{\scriptsize[38.95, 74.46]}} & \makecell{75.59\\{\scriptsize[67.44, 82.24]}} \\
    \addlinespace[2pt]
    \model{Qwen3.5-35B-A3B} & \makecell{97.06\\{\scriptsize[85.08, 99.48]}} & \makecell{68.42\\{\scriptsize[52.54, 80.92]}} & \makecell{58.62\\{\scriptsize[40.74, 74.49]}} & \makecell{57.69\\{\scriptsize[38.95, 74.46]}} & \makecell{71.65\\{\scriptsize[63.27, 78.77]}} \\[3pt]
    \midrule
    Random guessing & 25.00 & 25.00 & 25.00 & 25.00 & 25.00 \\
    \midrule
    \#Examples & 34 & 38 & 29 & 26 & 127 \\
    \bottomrule
  \end{tabular}
  \caption{Accuracy (\%) of each model on the multiple-choice reachability examples of the question depth experiment (\S\ref{subsec:complexity_scaling}), with reasoning effort \texttt{low}. Overall pools all question depths. The number of examples differs between columns because the task of each example is sampled at random. Below each accuracy value is its 95\% Wilson confidence interval. Bold marks the highest accuracy in each column. $^\dagger$\model{GPT-5.6-Sol} has fewer examples than the other models in this cell (93 Overall), so it is not bold. A dash marks a missing result.}
  \label{tab:v1_accuracy_by_question_depth_and_multiple}
\end{table*}

\begin{table*}[t]
  \centering
  \small
  \setlength{\tabcolsep}{5pt}
  \begin{tabular}{@{}lccccc@{}}
    \toprule
    & \multicolumn{4}{c}{\textbf{Question depth $k$}} & \\
    \cmidrule(lr){2-5}
    \textbf{Model} & \textbf{4} & \textbf{8} & \textbf{16} & \textbf{32} & \textbf{Overall} \\
    \midrule
    \model{GPT-5.6-Sol} & -- & \makecell{29.41\\{\scriptsize[16.83, 46.17]}} & \makecell{25.81\\{\scriptsize[13.70, 43.25]}} & \makecell{\textbf{41.94}\\{\scriptsize[26.42, 59.23]}} & \makecell{32.29$^\dagger$\\{\scriptsize[23.78, 42.17]}} \\
    \addlinespace[2pt]
    \model{GPT-5.6-Luna} & \makecell{0.00\\{\scriptsize[0.00, 10.72]}} & \makecell{0.00\\{\scriptsize[0.00, 10.15]}} & \makecell{0.00\\{\scriptsize[0.00, 11.03]}} & \makecell{0.00\\{\scriptsize[0.00, 11.03]}} & \makecell{0.00\\{\scriptsize[0.00, 2.91]}} \\
    \addlinespace[2pt]
    \model{Gemini-3.1-Pro} & \makecell{9.38\\{\scriptsize[3.24, 24.22]}} & \makecell{5.88\\{\scriptsize[1.63, 19.09]}} & \makecell{12.90\\{\scriptsize[5.13, 28.85]}} & \makecell{19.35\\{\scriptsize[9.19, 36.28]}} & \makecell{11.72\\{\scriptsize[7.23, 18.44]}} \\
    \addlinespace[2pt]
    \model{Gemini-3.8-Flash} & \makecell{\textbf{62.50}\\{\scriptsize[45.25, 77.07]}} & \makecell{\textbf{50.00}\\{\scriptsize[34.07, 65.93]}} & \makecell{\textbf{77.42}\\{\scriptsize[60.19, 88.61]}} & \makecell{29.03\\{\scriptsize[16.10, 46.59]}} & \makecell{\textbf{54.69}\\{\scriptsize[46.05, 63.05]}} \\
    \addlinespace[2pt]
    \model{Gemini-3.6-Flash} & \makecell{21.88\\{\scriptsize[11.02, 38.76]}} & \makecell{14.71\\{\scriptsize[6.45, 30.13]}} & \makecell{6.45\\{\scriptsize[1.79, 20.72]}} & \makecell{12.90\\{\scriptsize[5.13, 28.85]}} & \makecell{14.06\\{\scriptsize[9.08, 21.14]}} \\
    \addlinespace[2pt]
    \model{Qwen3.5-397B-A17B} & \makecell{0.00\\{\scriptsize[0.00, 10.72]}} & \makecell{0.00\\{\scriptsize[0.00, 10.15]}} & \makecell{0.00\\{\scriptsize[0.00, 11.03]}} & \makecell{0.00\\{\scriptsize[0.00, 11.03]}} & \makecell{0.00\\{\scriptsize[0.00, 2.91]}} \\
    \addlinespace[2pt]
    \model{Qwen3.5-122B-A10B} & \makecell{0.00\\{\scriptsize[0.00, 10.72]}} & \makecell{0.00\\{\scriptsize[0.00, 10.15]}} & \makecell{0.00\\{\scriptsize[0.00, 11.03]}} & \makecell{0.00\\{\scriptsize[0.00, 11.03]}} & \makecell{0.00\\{\scriptsize[0.00, 2.91]}} \\
    \addlinespace[2pt]
    \model{Qwen3.5-35B-A3B} & \makecell{0.00\\{\scriptsize[0.00, 10.72]}} & \makecell{0.00\\{\scriptsize[0.00, 10.15]}} & \makecell{0.00\\{\scriptsize[0.00, 11.03]}} & \makecell{0.00\\{\scriptsize[0.00, 11.03]}} & \makecell{0.00\\{\scriptsize[0.00, 2.91]}} \\[3pt]
    \midrule
    Random guessing & 0.66 & 0.66 & 0.66 & 0.66 & 0.66 \\
    \midrule
    \#Examples & 32 & 34 & 31 & 31 & 128 \\
    \bottomrule
  \end{tabular}
  \caption{Accuracy (\%) of each model on the reachability counting examples of the question depth experiment (\S\ref{subsec:complexity_scaling}), with reasoning effort \texttt{low}. Overall pools all question depths. The number of examples differs between columns because the task of each example is sampled at random. Below each accuracy value is its 95\% Wilson confidence interval. Bold marks the highest accuracy in each column. $^\dagger$\model{GPT-5.6-Sol} has fewer examples than the other models in this cell (96 Overall), so it is not bold. A dash marks a missing result.}
  \label{tab:v1_accuracy_by_question_depth_and_counting}
\end{table*}

\begin{table*}[t]
  \centering
  \small
  \setlength{\tabcolsep}{5pt}
  \begin{tabular}{@{}lccccc@{}}
    \toprule
    & \multicolumn{4}{c}{\textbf{Component size}} & \\
    \cmidrule(lr){2-5}
    \textbf{Model} & \textbf{100} & \textbf{150} & \textbf{200} & \textbf{250} & \textbf{Overall} \\
    \midrule
    \model{GPT-5.6-Sol} & \makecell{75.00$^\dagger$\\{\scriptsize[46.77, 91.11]}} & \makecell{57.14$^\dagger$\\{\scriptsize[25.05, 84.18]}} & \makecell{76.19\\{\scriptsize[61.47, 86.52]}} & \makecell{71.43\\{\scriptsize[52.94, 84.75]}} & \makecell{73.03$^\dagger$\\{\scriptsize[63.00, 81.16]}} \\
    \addlinespace[2pt]
    \model{GPT-5.6-Luna} & \makecell{65.79\\{\scriptsize[49.89, 78.79]}} & \makecell{57.14\\{\scriptsize[39.07, 73.49]}} & \makecell{59.52\\{\scriptsize[44.49, 72.96]}} & \makecell{53.57\\{\scriptsize[35.81, 70.47]}} & \makecell{59.56\\{\scriptsize[51.16, 67.43]}} \\
    \addlinespace[2pt]
    \model{Gemini-3.1-Pro} & \makecell{\textbf{86.84}\\{\scriptsize[72.67, 94.25]}} & \makecell{82.14\\{\scriptsize[64.41, 92.12]}} & \makecell{\textbf{78.57}\\{\scriptsize[64.06, 88.29]}} & \makecell{64.29\\{\scriptsize[45.83, 79.29]}} & \makecell{78.68\\{\scriptsize[71.05, 84.72]}} \\
    \addlinespace[2pt]
    \model{Gemini-3.8-Flash} & \makecell{\textbf{86.84}\\{\scriptsize[72.67, 94.25]}} & \makecell{\textbf{85.71}\\{\scriptsize[68.51, 94.30]}} & \makecell{\textbf{78.57}\\{\scriptsize[64.06, 88.29]}} & \makecell{\textbf{89.29}\\{\scriptsize[72.80, 96.29]}} & \makecell{\textbf{84.56}\\{\scriptsize[77.55, 89.67]}} \\
    \addlinespace[2pt]
    \model{Gemini-3.6-Flash} & \makecell{71.05\\{\scriptsize[55.24, 83.00]}} & \makecell{71.43\\{\scriptsize[52.94, 84.75]}} & \makecell{76.19\\{\scriptsize[61.47, 86.52]}} & \makecell{64.29\\{\scriptsize[45.83, 79.29]}} & \makecell{71.32\\{\scriptsize[63.22, 78.26]}} \\
    \addlinespace[2pt]
    \model{Qwen3.5-397B-A17B} & \makecell{73.68\\{\scriptsize[57.99, 85.03]}} & \makecell{57.14\\{\scriptsize[39.07, 73.49]}} & \makecell{61.90\\{\scriptsize[46.81, 75.00]}} & \makecell{57.14\\{\scriptsize[39.07, 73.49]}} & \makecell{63.24\\{\scriptsize[54.87, 70.87]}} \\
    \addlinespace[2pt]
    \model{Qwen3.5-122B-A10B} & \makecell{65.79\\{\scriptsize[49.89, 78.79]}} & \makecell{57.14\\{\scriptsize[39.07, 73.49]}} & \makecell{57.14\\{\scriptsize[42.21, 70.88]}} & \makecell{53.57\\{\scriptsize[35.81, 70.47]}} & \makecell{58.82\\{\scriptsize[50.42, 66.74]}} \\
    \addlinespace[2pt]
    \model{Qwen3.5-35B-A3B} & \makecell{60.53\\{\scriptsize[44.72, 74.40]}} & \makecell{57.14\\{\scriptsize[39.07, 73.49]}} & \makecell{57.14\\{\scriptsize[42.21, 70.88]}} & \makecell{53.57\\{\scriptsize[35.81, 70.47]}} & \makecell{57.35\\{\scriptsize[48.95, 65.35]}} \\[3pt]
    \midrule
    Random guessing & 50.00 & 50.00 & 50.00 & 50.00 & 50.00 \\
    \midrule
    \#Examples & 38 & 28 & 42 & 28 & 136 \\
    \bottomrule
  \end{tabular}
  \caption{Accuracy (\%) of each model on the YES/NO reachability examples of the component size experiment (\S\ref{subsec:complexity_scaling}), with reasoning effort \texttt{low}. Overall pools all component sizes. The number of examples differs between columns because the task of each example is sampled at random. Below each accuracy value is its 95\% Wilson confidence interval. Bold marks the highest accuracy in each column. $^\dagger$\model{GPT-5.6-Sol} has fewer examples than the other models in these cells (12 at 100, 7 at 150, and 89 Overall), so they are not bold.}
  \label{tab:v1_accuracy_by_node_count_and_yesno}
\end{table*}

\begin{table*}[t]
  \centering
  \small
  \setlength{\tabcolsep}{5pt}
  \begin{tabular}{@{}lccccc@{}}
    \toprule
    & \multicolumn{4}{c}{\textbf{Component size}} & \\
    \cmidrule(lr){2-5}
    \textbf{Model} & \textbf{100} & \textbf{150} & \textbf{200} & \textbf{250} & \textbf{Overall} \\
    \midrule
    \model{GPT-5.6-Sol} & \makecell{87.50$^\dagger$\\{\scriptsize[52.91, 97.76]}} & \makecell{90.91$^\dagger$\\{\scriptsize[62.26, 98.38]}} & \makecell{78.57\\{\scriptsize[60.46, 89.79]}} & \makecell{84.38\\{\scriptsize[68.25, 93.14]}} & \makecell{83.54$^\dagger$\\{\scriptsize[73.85, 90.12]}} \\
    \addlinespace[2pt]
    \model{GPT-5.6-Luna} & \makecell{57.58\\{\scriptsize[40.81, 72.76]}} & \makecell{64.71\\{\scriptsize[47.91, 78.51]}} & \makecell{60.71\\{\scriptsize[42.41, 76.43]}} & \makecell{68.75\\{\scriptsize[51.43, 82.05]}} & \makecell{62.99\\{\scriptsize[54.33, 70.89]}} \\
    \addlinespace[2pt]
    \model{Gemini-3.1-Pro} & \makecell{\textbf{96.97}\\{\scriptsize[84.68, 99.46]}} & \makecell{91.18\\{\scriptsize[77.04, 96.95]}} & \makecell{78.57\\{\scriptsize[60.46, 89.79]}} & \makecell{84.38\\{\scriptsize[68.25, 93.14]}} & \makecell{88.19\\{\scriptsize[81.43, 92.71]}} \\
    \addlinespace[2pt]
    \model{Gemini-3.8-Flash} & \makecell{\textbf{96.97}\\{\scriptsize[84.68, 99.46]}} & \makecell{\textbf{94.12}\\{\scriptsize[80.91, 98.37]}} & \makecell{\textbf{100.00}\\{\scriptsize[87.94, 100.00]}} & \makecell{\textbf{100.00}\\{\scriptsize[89.28, 100.00]}} & \makecell{\textbf{97.64}\\{\scriptsize[93.28, 99.19]}} \\
    \addlinespace[2pt]
    \model{Gemini-3.6-Flash} & \makecell{81.82\\{\scriptsize[65.61, 91.39]}} & \makecell{76.47\\{\scriptsize[60.00, 87.56]}} & \makecell{60.71\\{\scriptsize[42.41, 76.43]}} & \makecell{78.12\\{\scriptsize[61.24, 88.98]}} & \makecell{74.80\\{\scriptsize[66.60, 81.55]}} \\
    \addlinespace[2pt]
    \model{Qwen3.5-397B-A17B} & \makecell{78.79\\{\scriptsize[62.25, 89.32]}} & \makecell{70.59\\{\scriptsize[53.83, 83.17]}} & \makecell{60.71\\{\scriptsize[42.41, 76.43]}} & \makecell{75.00\\{\scriptsize[57.89, 86.75]}} & \makecell{71.65\\{\scriptsize[63.27, 78.77]}} \\
    \addlinespace[2pt]
    \model{Qwen3.5-122B-A10B} & \makecell{75.76\\{\scriptsize[58.98, 87.17]}} & \makecell{70.59\\{\scriptsize[53.83, 83.17]}} & \makecell{60.71\\{\scriptsize[42.41, 76.43]}} & \makecell{65.62\\{\scriptsize[48.31, 79.59]}} & \makecell{68.50\\{\scriptsize[59.98, 75.94]}} \\
    \addlinespace[2pt]
    \model{Qwen3.5-35B-A3B} & \makecell{60.61\\{\scriptsize[43.68, 75.32]}} & \makecell{64.71\\{\scriptsize[47.91, 78.51]}} & \makecell{64.29\\{\scriptsize[45.83, 79.29]}} & \makecell{68.75\\{\scriptsize[51.43, 82.05]}} & \makecell{64.57\\{\scriptsize[55.93, 72.35]}} \\[3pt]
    \midrule
    Random guessing & 25.00 & 25.00 & 25.00 & 25.00 & 25.00 \\
    \midrule
    \#Examples & 33 & 34 & 28 & 32 & 127 \\
    \bottomrule
  \end{tabular}
  \caption{Accuracy (\%) of each model on the multiple-choice reachability examples of the component size experiment (\S\ref{subsec:complexity_scaling}), with reasoning effort \texttt{low}. Overall pools all component sizes. The number of examples differs between columns because the task of each example is sampled at random. Below each accuracy value is its 95\% Wilson confidence interval. Bold marks the highest accuracy in each column. $^\dagger$\model{GPT-5.6-Sol} has fewer examples than the other models in these cells (8 at 100, 11 at 150, and 79 Overall), so they are not bold.}
  \label{tab:v1_accuracy_by_node_count_and_multiple}
\end{table*}

\begin{table*}[t]
  \centering
  \small
  \setlength{\tabcolsep}{5pt}
  \begin{tabular}{@{}lccccc@{}}
    \toprule
    & \multicolumn{4}{c}{\textbf{Component size}} & \\
    \cmidrule(lr){2-5}
    \textbf{Model} & \textbf{100} & \textbf{150} & \textbf{200} & \textbf{250} & \textbf{Overall} \\
    \midrule
    \model{GPT-5.6-Sol} & \makecell{71.43$^\dagger$\\{\scriptsize[35.89, 91.78]}} & \makecell{25.00$^\dagger$\\{\scriptsize[7.15, 59.07]}} & \makecell{20.00\\{\scriptsize[9.50, 37.31]}} & \makecell{5.00\\{\scriptsize[1.38, 16.50]}} & \makecell{17.65$^\dagger$\\{\scriptsize[11.00, 27.10]}} \\
    \addlinespace[2pt]
    \model{GPT-5.6-Luna} & \makecell{0.00\\{\scriptsize[0.00, 11.70]}} & \makecell{0.00\\{\scriptsize[0.00, 9.18]}} & \makecell{0.00\\{\scriptsize[0.00, 11.35]}} & \makecell{0.00\\{\scriptsize[0.00, 8.76]}} & \makecell{0.00\\{\scriptsize[0.00, 2.73]}} \\
    \addlinespace[2pt]
    \model{Gemini-3.1-Pro} & \makecell{31.03\\{\scriptsize[17.28, 49.23]}} & \makecell{7.89\\{\scriptsize[2.72, 20.80]}} & \makecell{0.00\\{\scriptsize[0.00, 11.35]}} & \makecell{2.50\\{\scriptsize[0.44, 12.88]}} & \makecell{9.49\\{\scriptsize[5.63, 15.56]}} \\
    \addlinespace[2pt]
    \model{Gemini-3.8-Flash} & \makecell{\textbf{79.31}\\{\scriptsize[61.61, 90.15]}} & \makecell{\textbf{55.26}\\{\scriptsize[39.71, 69.85]}} & \makecell{\textbf{50.00}\\{\scriptsize[33.15, 66.85]}} & \makecell{\textbf{25.00}\\{\scriptsize[14.19, 40.19]}} & \makecell{\textbf{50.36}\\{\scriptsize[42.10, 58.61]}} \\
    \addlinespace[2pt]
    \model{Gemini-3.6-Flash} & \makecell{51.72\\{\scriptsize[34.43, 68.61]}} & \makecell{15.79\\{\scriptsize[7.44, 30.42]}} & \makecell{16.67\\{\scriptsize[7.34, 33.56]}} & \makecell{0.00\\{\scriptsize[0.00, 8.76]}} & \makecell{18.98\\{\scriptsize[13.29, 26.36]}} \\
    \addlinespace[2pt]
    \model{Qwen3.5-397B-A17B} & \makecell{0.00\\{\scriptsize[0.00, 11.70]}} & \makecell{0.00\\{\scriptsize[0.00, 9.18]}} & \makecell{0.00\\{\scriptsize[0.00, 11.35]}} & \makecell{0.00\\{\scriptsize[0.00, 8.76]}} & \makecell{0.00\\{\scriptsize[0.00, 2.73]}} \\
    \addlinespace[2pt]
    \model{Qwen3.5-122B-A10B} & \makecell{0.00\\{\scriptsize[0.00, 11.70]}} & \makecell{0.00\\{\scriptsize[0.00, 9.18]}} & \makecell{0.00\\{\scriptsize[0.00, 11.35]}} & \makecell{0.00\\{\scriptsize[0.00, 8.76]}} & \makecell{0.00\\{\scriptsize[0.00, 2.73]}} \\
    \addlinespace[2pt]
    \model{Qwen3.5-35B-A3B} & \makecell{0.00\\{\scriptsize[0.00, 11.70]}} & \makecell{0.00\\{\scriptsize[0.00, 9.18]}} & \makecell{0.00\\{\scriptsize[0.00, 11.35]}} & \makecell{0.00\\{\scriptsize[0.00, 8.76]}} & \makecell{0.00\\{\scriptsize[0.00, 2.73]}} \\[3pt]
    \midrule
    Random guessing & 0.66 & 0.66 & 0.66 & 0.66 & 0.66 \\
    \midrule
    \#Examples & 29 & 38 & 30 & 40 & 137 \\
    \bottomrule
  \end{tabular}
  \caption{Accuracy (\%) of each model on the reachability counting examples of the component size experiment (\S\ref{subsec:complexity_scaling}), with reasoning effort \texttt{low}. Overall pools all component sizes. The number of examples differs between columns because the task of each example is sampled at random. Below each accuracy value is its 95\% Wilson confidence interval. Bold marks the highest accuracy in each column. $^\dagger$\model{GPT-5.6-Sol} has fewer examples than the other models in these cells (7 at 100, 8 at 150, and 85 Overall), so they are not bold.}
  \label{tab:v1_accuracy_by_node_count_and_counting}
\end{table*}

\begin{table*}[t]
  \centering
  \small
  \setlength{\tabcolsep}{5pt}
  \begin{tabular}{@{}llccccc@{}}
    \toprule
    & & \multicolumn{4}{c}{\textbf{Question depth $k$}} & \\
    \cmidrule(lr){3-6}
    \textbf{Model} & \textbf{Label} & \textbf{4} & \textbf{8} & \textbf{16} & \textbf{32} & \textbf{Overall} \\
    \midrule
    \multirow{3}{*}[-1pt]{\model{GPT-5.6-Sol}} & YES-label & -- & \makecell{92.86\\{\scriptsize[68.53, 98.73]}} & \makecell{15.00\\{\scriptsize[5.24, 36.04]}} & \makecell{4.76\\{\scriptsize[0.85, 22.67]}} & \makecell{30.91$^\dagger$\\{\scriptsize[20.28, 44.03]}} \\
    \addlinespace[2pt]
     & NO-label & -- & \makecell{100.00\\{\scriptsize[78.47, 100.00]}} & \makecell{100.00\\{\scriptsize[83.89, 100.00]}} & \makecell{100.00\\{\scriptsize[85.13, 100.00]}} & \makecell{100.00$^\dagger$\\{\scriptsize[93.58, 100.00]}} \\[3pt]
    \cmidrule(l){2-7}
    \multirow{3}{*}[-1pt]{\model{GPT-5.6-Luna}} & YES-label & \makecell{88.24\\{\scriptsize[65.66, 96.71]}} & \makecell{7.14\\{\scriptsize[1.27, 31.47]}} & \makecell{0.00\\{\scriptsize[0.00, 16.11]}} & \makecell{0.00\\{\scriptsize[0.00, 15.46]}} & \makecell{22.22\\{\scriptsize[14.17, 33.09]}} \\
    \addlinespace[2pt]
     & NO-label & \makecell{100.00\\{\scriptsize[81.57, 100.00]}} & \makecell{92.86\\{\scriptsize[68.53, 98.73]}} & \makecell{100.00\\{\scriptsize[83.89, 100.00]}} & \makecell{100.00\\{\scriptsize[85.13, 100.00]}} & \makecell{98.63\\{\scriptsize[92.64, 99.76]}} \\[3pt]
    \cmidrule(l){2-7}
    \multirow{3}{*}[-1pt]{\model{Gemini-3.1-Pro}} & YES-label & \makecell{100.00\\{\scriptsize[81.57, 100.00]}} & \makecell{78.57\\{\scriptsize[52.41, 92.43]}} & \makecell{35.00\\{\scriptsize[18.12, 56.71]}} & \makecell{33.33\\{\scriptsize[17.19, 54.63]}} & \makecell{58.33\\{\scriptsize[46.81, 69.01]}} \\
    \addlinespace[2pt]
     & NO-label & \makecell{100.00\\{\scriptsize[81.57, 100.00]}} & \makecell{100.00\\{\scriptsize[78.47, 100.00]}} & \makecell{100.00\\{\scriptsize[83.89, 100.00]}} & \makecell{100.00\\{\scriptsize[85.13, 100.00]}} & \makecell{100.00\\{\scriptsize[95.00, 100.00]}} \\[3pt]
    \cmidrule(l){2-7}
    \multirow{3}{*}[-1pt]{\model{Gemini-3.8-Flash}} & YES-label & \makecell{100.00\\{\scriptsize[81.57, 100.00]}} & \makecell{100.00\\{\scriptsize[78.47, 100.00]}} & \makecell{90.00\\{\scriptsize[69.90, 97.21]}} & \makecell{100.00\\{\scriptsize[84.54, 100.00]}} & \makecell{97.22\\{\scriptsize[90.43, 99.23]}} \\
    \addlinespace[2pt]
     & NO-label & \makecell{94.12\\{\scriptsize[73.02, 98.95]}} & \makecell{100.00\\{\scriptsize[78.47, 100.00]}} & \makecell{80.00\\{\scriptsize[58.40, 91.93]}} & \makecell{36.36\\{\scriptsize[19.73, 57.05]}} & \makecell{73.97\\{\scriptsize[62.89, 82.66]}} \\[3pt]
    \cmidrule(l){2-7}
    \multirow{3}{*}[-1pt]{\model{Gemini-3.6-Flash}} & YES-label & \makecell{100.00\\{\scriptsize[81.57, 100.00]}} & \makecell{64.29\\{\scriptsize[38.76, 83.66]}} & \makecell{35.00\\{\scriptsize[18.12, 56.71]}} & \makecell{19.05\\{\scriptsize[7.67, 40.00]}} & \makecell{51.39\\{\scriptsize[40.07, 62.57]}} \\
    \addlinespace[2pt]
     & NO-label & \makecell{100.00\\{\scriptsize[81.57, 100.00]}} & \makecell{100.00\\{\scriptsize[78.47, 100.00]}} & \makecell{100.00\\{\scriptsize[83.89, 100.00]}} & \makecell{95.45\\{\scriptsize[78.20, 99.19]}} & \makecell{98.63\\{\scriptsize[92.64, 99.76]}} \\[3pt]
    \cmidrule(l){2-7}
    \multirow{3}{*}[-1pt]{\model{Qwen3.5-397B-A17B}} & YES-label & \makecell{88.24\\{\scriptsize[65.66, 96.71]}} & \makecell{42.86\\{\scriptsize[21.38, 67.41]}} & \makecell{0.00\\{\scriptsize[0.00, 16.11]}} & \makecell{0.00\\{\scriptsize[0.00, 15.46]}} & \makecell{29.17\\{\scriptsize[19.94, 40.51]}} \\
    \addlinespace[2pt]
     & NO-label & \makecell{100.00\\{\scriptsize[81.57, 100.00]}} & \makecell{100.00\\{\scriptsize[78.47, 100.00]}} & \makecell{100.00\\{\scriptsize[83.89, 100.00]}} & \makecell{100.00\\{\scriptsize[85.13, 100.00]}} & \makecell{100.00\\{\scriptsize[95.00, 100.00]}} \\[3pt]
    \cmidrule(l){2-7}
    \multirow{3}{*}[-1pt]{\model{Qwen3.5-122B-A10B}} & YES-label & \makecell{88.24\\{\scriptsize[65.66, 96.71]}} & \makecell{21.43\\{\scriptsize[7.57, 47.59]}} & \makecell{0.00\\{\scriptsize[0.00, 16.11]}} & \makecell{0.00\\{\scriptsize[0.00, 15.46]}} & \makecell{25.00\\{\scriptsize[16.44, 36.09]}} \\
    \addlinespace[2pt]
     & NO-label & \makecell{100.00\\{\scriptsize[81.57, 100.00]}} & \makecell{100.00\\{\scriptsize[78.47, 100.00]}} & \makecell{100.00\\{\scriptsize[83.89, 100.00]}} & \makecell{100.00\\{\scriptsize[85.13, 100.00]}} & \makecell{100.00\\{\scriptsize[95.00, 100.00]}} \\[3pt]
    \cmidrule(l){2-7}
    \multirow{3}{*}[-1pt]{\model{Qwen3.5-35B-A3B}} & YES-label & \makecell{76.47\\{\scriptsize[52.74, 90.45]}} & \makecell{14.29\\{\scriptsize[4.01, 39.94]}} & \makecell{0.00\\{\scriptsize[0.00, 16.11]}} & \makecell{9.52\\{\scriptsize[2.65, 28.91]}} & \makecell{23.61\\{\scriptsize[15.30, 34.60]}} \\
    \addlinespace[2pt]
     & NO-label & \makecell{100.00\\{\scriptsize[81.57, 100.00]}} & \makecell{100.00\\{\scriptsize[78.47, 100.00]}} & \makecell{100.00\\{\scriptsize[83.89, 100.00]}} & \makecell{100.00\\{\scriptsize[85.13, 100.00]}} & \makecell{100.00\\{\scriptsize[95.00, 100.00]}} \\[3pt]
    \cmidrule(l){2-7}
    Random guessing &  & 50.00 & 50.00 & 50.00 & 50.00 & 50.00 \\
    \midrule
    \multirow{2}{*}{\#Examples} & YES-label & 17 & 14 & 20 & 21 & 72 \\
     & NO-label & 17 & 14 & 20 & 22 & 73 \\
    \bottomrule
  \end{tabular}
  \caption{YES/NO reachability accuracy (\%) of each model on YES-label and NO-label examples of the question depth experiment (\S\ref{subsec:complexity_scaling}), with reasoning effort \texttt{low}. Overall pools all question depths. Below each accuracy value is its 95\% Wilson confidence interval. $^\dagger$\model{GPT-5.6-Sol} has fewer examples than the other models in these cells: YES-label, 55 Overall; NO-label, 56 Overall. A dash marks a missing result.}
  \label{tab:v1_yesno_accuracy_by_question_depth_and_label}
\end{table*}

\begin{table*}[t]
  \centering
  \small
  \setlength{\tabcolsep}{5pt}
  \begin{tabular}{@{}llccccc@{}}
    \toprule
    & & \multicolumn{4}{c}{\textbf{Component size}} & \\
    \cmidrule(lr){3-6}
    \textbf{Model} & \textbf{Label} & \textbf{100} & \textbf{150} & \textbf{200} & \textbf{250} & \textbf{Overall} \\
    \midrule
    \multirow{3}{*}[-1pt]{\model{GPT-5.6-Sol}} & YES-label & \makecell{50.00$^\dagger$\\{\scriptsize[18.76, 81.24]}} & \makecell{25.00$^\dagger$\\{\scriptsize[4.56, 69.94]}} & \makecell{52.38\\{\scriptsize[32.37, 71.66]}} & \makecell{42.86\\{\scriptsize[21.38, 67.41]}} & \makecell{46.67$^\dagger$\\{\scriptsize[32.93, 60.92]}} \\
    \addlinespace[2pt]
     & NO-label & \makecell{100.00$^\dagger$\\{\scriptsize[60.97, 100.00]}} & \makecell{100.00$^\dagger$\\{\scriptsize[43.85, 100.00]}} & \makecell{100.00\\{\scriptsize[84.54, 100.00]}} & \makecell{100.00\\{\scriptsize[78.47, 100.00]}} & \makecell{100.00$^\dagger$\\{\scriptsize[91.97, 100.00]}} \\[3pt]
    \cmidrule(l){2-7}
    \multirow{3}{*}[-1pt]{\model{GPT-5.6-Luna}} & YES-label & \makecell{31.58\\{\scriptsize[15.36, 53.99]}} & \makecell{14.29\\{\scriptsize[4.01, 39.94]}} & \makecell{19.05\\{\scriptsize[7.67, 40.00]}} & \makecell{7.14\\{\scriptsize[1.27, 31.47]}} & \makecell{19.12\\{\scriptsize[11.53, 30.01]}} \\
    \addlinespace[2pt]
     & NO-label & \makecell{100.00\\{\scriptsize[83.18, 100.00]}} & \makecell{100.00\\{\scriptsize[78.47, 100.00]}} & \makecell{100.00\\{\scriptsize[84.54, 100.00]}} & \makecell{100.00\\{\scriptsize[78.47, 100.00]}} & \makecell{100.00\\{\scriptsize[94.65, 100.00]}} \\[3pt]
    \cmidrule(l){2-7}
    \multirow{3}{*}[-1pt]{\model{Gemini-3.1-Pro}} & YES-label & \makecell{73.68\\{\scriptsize[51.21, 88.19]}} & \makecell{64.29\\{\scriptsize[38.76, 83.66]}} & \makecell{57.14\\{\scriptsize[36.55, 75.53]}} & \makecell{28.57\\{\scriptsize[11.72, 54.65]}} & \makecell{57.35\\{\scriptsize[45.52, 68.40]}} \\
    \addlinespace[2pt]
     & NO-label & \makecell{100.00\\{\scriptsize[83.18, 100.00]}} & \makecell{100.00\\{\scriptsize[78.47, 100.00]}} & \makecell{100.00\\{\scriptsize[84.54, 100.00]}} & \makecell{100.00\\{\scriptsize[78.47, 100.00]}} & \makecell{100.00\\{\scriptsize[94.65, 100.00]}} \\[3pt]
    \cmidrule(l){2-7}
    \multirow{3}{*}[-1pt]{\model{Gemini-3.8-Flash}} & YES-label & \makecell{100.00\\{\scriptsize[83.18, 100.00]}} & \makecell{92.86\\{\scriptsize[68.53, 98.73]}} & \makecell{100.00\\{\scriptsize[84.54, 100.00]}} & \makecell{100.00\\{\scriptsize[78.47, 100.00]}} & \makecell{98.53\\{\scriptsize[92.13, 99.74]}} \\
    \addlinespace[2pt]
     & NO-label & \makecell{73.68\\{\scriptsize[51.21, 88.19]}} & \makecell{78.57\\{\scriptsize[52.41, 92.43]}} & \makecell{57.14\\{\scriptsize[36.55, 75.53]}} & \makecell{78.57\\{\scriptsize[52.41, 92.43]}} & \makecell{70.59\\{\scriptsize[58.89, 80.08]}} \\[3pt]
    \cmidrule(l){2-7}
    \multirow{3}{*}[-1pt]{\model{Gemini-3.6-Flash}} & YES-label & \makecell{57.89\\{\scriptsize[36.28, 76.86]}} & \makecell{42.86\\{\scriptsize[21.38, 67.41]}} & \makecell{52.38\\{\scriptsize[32.37, 71.66]}} & \makecell{28.57\\{\scriptsize[11.72, 54.65]}} & \makecell{47.06\\{\scriptsize[35.67, 58.76]}} \\
    \addlinespace[2pt]
     & NO-label & \makecell{84.21\\{\scriptsize[62.43, 94.48]}} & \makecell{100.00\\{\scriptsize[78.47, 100.00]}} & \makecell{100.00\\{\scriptsize[84.54, 100.00]}} & \makecell{100.00\\{\scriptsize[78.47, 100.00]}} & \makecell{95.59\\{\scriptsize[87.81, 98.49]}} \\[3pt]
    \cmidrule(l){2-7}
    \multirow{3}{*}[-1pt]{\model{Qwen3.5-397B-A17B}} & YES-label & \makecell{47.37\\{\scriptsize[27.33, 68.29]}} & \makecell{14.29\\{\scriptsize[4.01, 39.94]}} & \makecell{23.81\\{\scriptsize[10.63, 45.09]}} & \makecell{14.29\\{\scriptsize[4.01, 39.94]}} & \makecell{26.47\\{\scriptsize[17.45, 38.01]}} \\
    \addlinespace[2pt]
     & NO-label & \makecell{100.00\\{\scriptsize[83.18, 100.00]}} & \makecell{100.00\\{\scriptsize[78.47, 100.00]}} & \makecell{100.00\\{\scriptsize[84.54, 100.00]}} & \makecell{100.00\\{\scriptsize[78.47, 100.00]}} & \makecell{100.00\\{\scriptsize[94.65, 100.00]}} \\[3pt]
    \cmidrule(l){2-7}
    \multirow{3}{*}[-1pt]{\model{Qwen3.5-122B-A10B}} & YES-label & \makecell{31.58\\{\scriptsize[15.36, 53.99]}} & \makecell{14.29\\{\scriptsize[4.01, 39.94]}} & \makecell{14.29\\{\scriptsize[4.98, 34.64]}} & \makecell{7.14\\{\scriptsize[1.27, 31.47]}} & \makecell{17.65\\{\scriptsize[10.39, 28.36]}} \\
    \addlinespace[2pt]
     & NO-label & \makecell{100.00\\{\scriptsize[83.18, 100.00]}} & \makecell{100.00\\{\scriptsize[78.47, 100.00]}} & \makecell{100.00\\{\scriptsize[84.54, 100.00]}} & \makecell{100.00\\{\scriptsize[78.47, 100.00]}} & \makecell{100.00\\{\scriptsize[94.65, 100.00]}} \\[3pt]
    \cmidrule(l){2-7}
    \multirow{3}{*}[-1pt]{\model{Qwen3.5-35B-A3B}} & YES-label & \makecell{21.05\\{\scriptsize[8.51, 43.33]}} & \makecell{21.43\\{\scriptsize[7.57, 47.59]}} & \makecell{14.29\\{\scriptsize[4.98, 34.64]}} & \makecell{7.14\\{\scriptsize[1.27, 31.47]}} & \makecell{16.18\\{\scriptsize[9.28, 26.69]}} \\
    \addlinespace[2pt]
     & NO-label & \makecell{100.00\\{\scriptsize[83.18, 100.00]}} & \makecell{92.86\\{\scriptsize[68.53, 98.73]}} & \makecell{100.00\\{\scriptsize[84.54, 100.00]}} & \makecell{100.00\\{\scriptsize[78.47, 100.00]}} & \makecell{98.53\\{\scriptsize[92.13, 99.74]}} \\[3pt]
    \cmidrule(l){2-7}
    Random guessing &  & 50.00 & 50.00 & 50.00 & 50.00 & 50.00 \\
    \midrule
    \multirow{2}{*}{\#Examples} & YES-label & 19 & 14 & 21 & 14 & 68 \\
     & NO-label & 19 & 14 & 21 & 14 & 68 \\
    \bottomrule
  \end{tabular}
  \caption{YES/NO reachability accuracy (\%) of each model on YES-label and NO-label examples of the component size experiment (\S\ref{subsec:complexity_scaling}), with reasoning effort \texttt{low}. Overall pools all component sizes. Below each accuracy value is its 95\% Wilson confidence interval. $^\dagger$\model{GPT-5.6-Sol} has fewer examples than the other models in these cells: YES-label, 6 at 100, 4 at 150, and 45 Overall; NO-label, 6 at 100, 3 at 150, and 44 Overall.}
  \label{tab:v1_yesno_accuracy_by_node_count_and_label}
\end{table*}

\subsection{Reachability Tasks}

For the task experiment (\S\ref{subsec:performance_on_different_reachability_tasks}), \Cref{tab:v1_yesno_accuracy_by_label,tab:v1_yesno_no_answers_by_question_depth} report YES/NO reachability accuracy by ground truth answer and the percentage of NO answers by question depth. \Cref{tab:v1_counting_answer_outcomes,tab:v1_counting_accuracy_by_component_size} compare the counting answers with the ground truth counts and report counting accuracy by component size.
Tables~\ref{tab:v1_multiple_choice_none_of_the_above} and~\ref{tab:v1_multiple_choice_none_of_the_above_tokens} report the accuracy and the number of tokens in the model's answer for the multiple-choice task with a "None of the above" option.

\begin{table*}[t]
  \centering
  \small
  \setlength{\tabcolsep}{5pt}
  \begin{tabular}{@{}lccc@{}}
    \toprule
    & \multicolumn{2}{c}{\textbf{Label}} & \\
    \cmidrule(lr){2-3}
    \textbf{Model} & \textbf{YES-label} & \textbf{NO-label} & \textbf{Overall} \\
    \midrule
    \model{GPT-5.6-Luna} & \makecell{32.00\\{\scriptsize[20.76, 45.81]}} & \makecell{\textbf{100.00}\\{\scriptsize[92.86, 100.00]}} & \makecell{66.00\\{\scriptsize[56.28, 74.54]}} \\
    \addlinespace[2pt]
    \model{Gemini-3.8-Flash} & \makecell{\textbf{96.00}\\{\scriptsize[86.54, 98.90]}} & \makecell{78.00\\{\scriptsize[64.76, 87.25]}} & \makecell{\textbf{87.00}\\{\scriptsize[79.02, 92.24]}} \\
    \addlinespace[2pt]
    \model{Qwen3.5-35B-A3B} & \makecell{28.00\\{\scriptsize[17.47, 41.67]}} & \makecell{\textbf{100.00}\\{\scriptsize[92.86, 100.00]}} & \makecell{64.00\\{\scriptsize[54.24, 72.73]}} \\[3pt]
    \midrule
    Random guessing & 50.00 & 50.00 & 50.00 \\
    \midrule
    \#Examples & 50 & 50 & 100 \\
    \bottomrule
  \end{tabular}
  \caption{YES/NO reachability accuracy (\%) of each model on YES-label and NO-label examples of the task experiment (\S\ref{subsec:performance_on_different_reachability_tasks}), with reasoning effort \texttt{low}. Below each accuracy value is its 95\% Wilson confidence interval. Bold marks the highest accuracy in each column.}
  \label{tab:v1_yesno_accuracy_by_label}
\end{table*}

\begin{table*}[t]
  \centering
  \small
  \setlength{\tabcolsep}{5pt}
  \begin{tabular}{@{}lccccc@{}}
    \toprule
    & \multicolumn{4}{c}{\textbf{Question depth $k$}} & \\
    \cmidrule(lr){2-5}
    \textbf{Model} & \textbf{4} & \textbf{8} & \textbf{16} & \textbf{32} & \textbf{Overall} \\
    \midrule
    Ground truth & 43.48 & 64.29 & 43.48 & 46.15 & 50.00 \\[3pt]
    \addlinespace[2pt]
    \model{GPT-5.6-Luna} & \makecell{43.48\\{\scriptsize[25.63, 63.19]}} & \makecell{89.29\\{\scriptsize[72.80, 96.29]}} & \makecell{100.00\\{\scriptsize[85.69, 100.00]}} & \makecell{100.00\\{\scriptsize[87.13, 100.00]}} & \makecell{84.00\\{\scriptsize[75.58, 89.90]}} \\
    \addlinespace[2pt]
    \model{Gemini-3.8-Flash} & \makecell{43.48\\{\scriptsize[25.63, 63.19]}} & \makecell{64.29\\{\scriptsize[45.83, 79.29]}} & \makecell{39.13\\{\scriptsize[22.16, 59.21]}} & \makecell{15.38\\{\scriptsize[6.15, 33.53]}} & \makecell{41.00\\{\scriptsize[31.87, 50.80]}} \\
    \addlinespace[2pt]
    \model{Qwen3.5-35B-A3B} & \makecell{47.83\\{\scriptsize[29.24, 67.04]}} & \makecell{92.86\\{\scriptsize[77.35, 98.02]}} & \makecell{100.00\\{\scriptsize[85.69, 100.00]}} & \makecell{100.00\\{\scriptsize[87.13, 100.00]}} & \makecell{86.00\\{\scriptsize[77.86, 91.47]}} \\[3pt]
    \midrule
    \#Examples & 23 & 28 & 23 & 26 & 100 \\
    \bottomrule
  \end{tabular}
  \caption{Percentage of the YES/NO reachability examples of the task experiment (\S\ref{subsec:performance_on_different_reachability_tasks}) that each model answers \texttt{NO}, overall and by question depth $k$, with reasoning effort \texttt{low}. In this experiment, the question depth of each example is sampled uniformly at random. Below each percentage is its 95\% Wilson confidence interval. The first row is the percentage of NO-label examples, that is, the percentage of \texttt{NO} answers of a model that is always correct. Every answer is \texttt{YES} or \texttt{NO}, so the percentage of \texttt{YES} answers is 100 minus the value shown.}
  \label{tab:v1_yesno_no_answers_by_question_depth}
\end{table*}

\begin{table*}[t]
  \centering
  \small
  \setlength{\tabcolsep}{5pt}
  \begin{tabular}{@{}lcccccc@{}}
    \toprule
    & \multicolumn{6}{c}{\textbf{Counting answers}} \\
    \cmidrule(lr){2-7}
    \textbf{Model} & \textbf{Exact} & \textbf{Too low} & \textbf{Too high} & \textbf{No number} & \textbf{Within 10\%} & \textbf{Median ratio} \\
    \midrule
    \model{GPT-5.6-Luna} & \makecell{0.00\\{\scriptsize[0.00, 3.70]}} & \makecell{98.00\\{\scriptsize[93.00, 99.45]}} & \makecell{2.00\\{\scriptsize[0.55, 7.00]}} & \makecell{0.00\\{\scriptsize[0.00, 3.70]}} & \makecell{1.00\\{\scriptsize[0.18, 5.45]}} & 0.30 \\
    \addlinespace[2pt]
    \model{Gemini-3.8-Flash} & \makecell{55.00\\{\scriptsize[45.24, 64.39]}} & \makecell{12.00\\{\scriptsize[7.00, 19.81]}} & \makecell{15.00\\{\scriptsize[9.31, 23.28]}} & \makecell{18.00\\{\scriptsize[11.70, 26.67]}} & \makecell{64.00\\{\scriptsize[54.24, 72.73]}} & 1.00 \\
    \addlinespace[2pt]
    \model{Qwen3.5-35B-A3B} & \makecell{0.00\\{\scriptsize[0.00, 3.70]}} & \makecell{99.00\\{\scriptsize[94.55, 99.82]}} & \makecell{0.00\\{\scriptsize[0.00, 3.70]}} & \makecell{1.00\\{\scriptsize[0.18, 5.45]}} & \makecell{0.00\\{\scriptsize[0.00, 3.70]}} & 0.16 \\[3pt]
    \bottomrule
  \end{tabular}
  \caption{Outcomes of the reachability counting answers of each model in the task experiment (\S\ref{subsec:performance_on_different_reachability_tasks}), with reasoning effort \texttt{low}, as a percentage of its 100 counting examples; below each percentage is its 95\% Wilson confidence interval. Exact, Too low, Too high, and No number add up to 100\%: Too low and Too high compare a wrong numeric answer with the ground truth count, and No number covers responses without a number in the final answer block, including responses without an answer block. Within 10\% includes exact answers. Median ratio is the median of the predicted count divided by the ground truth count over the numeric answers.}
  \label{tab:v1_counting_answer_outcomes}
\end{table*}

\begin{table*}[t]
  \centering
  \small
  \setlength{\tabcolsep}{5pt}
  \begin{tabular}{@{}lcccc@{}}
    \toprule
    & \multicolumn{3}{c}{\textbf{Component size}} & \\
    \cmidrule(lr){2-4}
    \textbf{Model} & \textbf{100--149} & \textbf{150--199} & \textbf{200--250} & \textbf{Overall} \\
    \midrule
    \model{GPT-5.6-Luna} & \makecell{0.00\\{\scriptsize[0.00, 10.43]}} & \makecell{0.00\\{\scriptsize[0.00, 10.72]}} & \makecell{0.00\\{\scriptsize[0.00, 9.89]}} & \makecell{0.00\\{\scriptsize[0.00, 3.70]}} \\
    \addlinespace[2pt]
    \model{Gemini-3.8-Flash} & \makecell{\textbf{78.79}\\{\scriptsize[62.25, 89.32]}} & \makecell{\textbf{59.38}\\{\scriptsize[42.26, 74.48]}} & \makecell{\textbf{28.57}\\{\scriptsize[16.33, 45.06]}} & \makecell{\textbf{55.00}\\{\scriptsize[45.24, 64.39]}} \\
    \addlinespace[2pt]
    \model{Qwen3.5-35B-A3B} & \makecell{0.00\\{\scriptsize[0.00, 10.43]}} & \makecell{0.00\\{\scriptsize[0.00, 10.72]}} & \makecell{0.00\\{\scriptsize[0.00, 9.89]}} & \makecell{0.00\\{\scriptsize[0.00, 3.70]}} \\[3pt]
    \midrule
    Random guessing & 0.66 & 0.66 & 0.66 & 0.66 \\
    \midrule
    \#Examples & 33 & 32 & 35 & 100 \\
    \bottomrule
  \end{tabular}
  \caption{Reachability counting accuracy (\%) of each model in the task experiment (\S\ref{subsec:performance_on_different_reachability_tasks}) by component size, with reasoning effort \texttt{low}. In this experiment, the component size of each example is sampled uniformly at random from the integers in $[100, 250]$. Below each accuracy value is its 95\% Wilson confidence interval. Bold marks the highest accuracy in each column.}
  \label{tab:v1_counting_accuracy_by_component_size}
\end{table*}

\begin{table*}[t]
  \centering
  \small
  \setlength{\tabcolsep}{5pt}
  \begin{tabular}{@{}lccccc@{}}
    \toprule
    & \multicolumn{5}{c}{\textbf{Adding ``None of the above''}} \\
    \cmidrule(lr){2-6}
    \textbf{Model} & \textbf{\#Examples} & \textbf{Before} & \textbf{After} & \textbf{Change (pp)} & \textbf{Gained / lost} \\
    \midrule
    \model{GPT-5.6-Luna} & 100 & \makecell{68.00\\{\scriptsize[58.34, 76.33]}} & \makecell{14.00\\{\scriptsize[8.53, 22.14]}} & $-54.00$ & \makecell{$+$0 / $-$54\\{\scriptsize $p$ = $<$0.01}} \\
    \addlinespace[2pt]
    \model{Gemini-3.8-Flash} & 100 & \makecell{97.00\\{\scriptsize[91.55, 98.97]}} & \makecell{97.00\\{\scriptsize[91.55, 98.97]}} & $+0.00$ & \makecell{$+$3 / $-$3\\{\scriptsize $p$ = 1.00}} \\
    \addlinespace[2pt]
    \model{Qwen3.5-35B-A3B} & 100 & \makecell{72.00\\{\scriptsize[62.51, 79.86]}} & \makecell{15.00\\{\scriptsize[9.31, 23.28]}} & $-57.00$ & \makecell{$+$0 / $-$57\\{\scriptsize $p$ = $<$0.01}} \\
    \addlinespace[2pt]
    Random guessing &  & 25.00 & 20.00 & $-5.00$ &  \\
    \bottomrule
  \end{tabular}
  \caption{Multiple-choice reachability accuracy (\%) of each model on the multiple-choice examples of the task experiment (\S\ref{subsec:performance_on_different_reachability_tasks}), with reasoning effort \texttt{low}, before and after we append a fifth option, ``None of these targets is reachable from the source.'' The context, the four original options, and the ground truth answer are unchanged, so the new option is never correct. \#Examples counts the examples that the model answered both before and after, and both accuracies use only these examples. Below each accuracy value is its 95\% Wilson confidence interval. Change is After minus Before, in percentage points. Gained / lost gives the number of examples that become correct ($+$) and wrong ($-$) after the option is added, and below it the $p$-value of the exact McNemar test. Random guessing is the expected accuracy of uniformly random answers (Appendix~\ref{appdx:random_guessing_baseline}), with four options before and five after.}
  \label{tab:v1_multiple_choice_none_of_the_above}
\end{table*}

\begin{table*}[t]
  \centering
  \small
  \setlength{\tabcolsep}{5pt}
  \begin{tabular}{@{}llcccc@{}}
    \toprule
    \textbf{Model} & \textbf{Tokens} & \textbf{Before} & \textbf{After} & \textbf{Median change} & \textbf{$p$} \\
    \midrule
    \multirow{3}{*}[-1pt]{\model{GPT-5.6-Luna}} & Output & \makecell{638.95\\{\scriptsize$\pm$300.07}} & \makecell{306.26\\{\scriptsize$\pm$167.83}} & $-$318.00 & $<$0.01 \\
    \addlinespace[2pt]
     & Reasoning & \makecell{621.85\\{\scriptsize$\pm$296.96}} & \makecell{288.22\\{\scriptsize$\pm$163.40}} & $-$318.00 & $<$0.01 \\[3pt]
    \cmidrule(l){2-6}
    \multirow{3}{*}[-1pt]{\model{Gemini-3.8-Flash}} & Output & \makecell{5,271.84\\{\scriptsize$\pm$5,039.82}} & \makecell{5,702.91\\{\scriptsize$\pm$7,893.90}} & $+$45.50 & 0.42 \\
    \addlinespace[2pt]
     & Reasoning & \makecell{4,595.61\\{\scriptsize$\pm$4,774.19}} & \makecell{4,981.18\\{\scriptsize$\pm$7,575.28}} & $+$34.50 & 0.68 \\[3pt]
    \cmidrule(l){2-6}
    \multirow{3}{*}[-1pt]{\model{Qwen3.5-35B-A3B}} & Output & \makecell{19,619.60\\{\scriptsize$\pm$9,206.76}} & \makecell{10,954.93\\{\scriptsize$\pm$3,664.83}} & $-$8,295.00 & $<$0.01 \\
    \addlinespace[2pt]
     & Reasoning & \makecell{18,064.96\\{\scriptsize$\pm$6,741.73}} & \makecell{10,505.45\\{\scriptsize$\pm$3,679.49}} & $-$8,477.00 & $<$0.01 \\[3pt]
    \bottomrule
  \end{tabular}
  \caption{Mean output tokens and reasoning tokens per example of each model before and after adding the ``None of these targets is reachable from the source.'' option, on the same examples as Table~\ref{tab:v1_multiple_choice_none_of_the_above}, with reasoning effort \texttt{low}. Below each mean is its standard deviation. Output tokens are the completion tokens that the provider reports, and reasoning tokens are the tokens that it reports as reasoning. Median change is the median over the examples of After minus Before, and $p$ is the two-sided Wilcoxon signed-rank test of this difference; the $p$-values are not adjusted for multiple tests.}
  \label{tab:v1_multiple_choice_none_of_the_above_tokens}
\end{table*}

\subsection{Distractor Type}

\Cref{tab:v1_accuracy_by_distractor_type_and_task} splits the accuracy of the distractor type experiment (\S\ref{subsec:effects_of_distractor_type}) by task, and \Cref{tab:v1_distractor_type_tests} compares all five distractor types with a $\chi^2$ test, as well as compare each pair of distractor types with Fisher's exact test, corrected for the ten pairs with the Holm method.

\begin{table*}[t]
  \centering
  \small
  \setlength{\tabcolsep}{5pt}
  \begin{tabular}{@{}llcccccc@{}}
    \toprule
    & & \multicolumn{5}{c}{\textbf{Distractor type}} & \\
    \cmidrule(lr){3-7}
    \textbf{Model} & \textbf{Task} & \textbf{\texttt{books}} & \textbf{\texttt{code}} & \textbf{\makecell{\texttt{disjoint-}\\\texttt{component}}} & \textbf{\texttt{essays}} & \textbf{\texttt{wikipedia}} & \textbf{Overall} \\
    \midrule
    \multirow{5}{*}[-2pt]{\model{GPT-5.6-Luna}} & YES/NO & \makecell{56.41\\{\scriptsize[40.98, 70.70]}} & \makecell{65.62\\{\scriptsize[48.31, 79.59]}} & \makecell{59.38\\{\scriptsize[42.26, 74.48]}} & \makecell{67.74\\{\scriptsize[50.14, 81.43]}} & \makecell{56.52\\{\scriptsize[36.81, 74.37]}} & \makecell{61.15\\{\scriptsize[53.34, 68.42]}} \\
    \addlinespace[2pt]
     & Multiple choice & \makecell{63.33\\{\scriptsize[45.51, 78.13]}} & \makecell{71.43\\{\scriptsize[52.94, 84.75]}} & \makecell{78.57\\{\scriptsize[60.46, 89.79]}} & \makecell{69.44\\{\scriptsize[53.14, 82.00]}} & \makecell{84.09\\{\scriptsize[70.63, 92.07]}} & \makecell{74.10\\{\scriptsize[66.94, 80.16]}} \\
    \addlinespace[2pt]
     & Counting & \makecell{0.00\\{\scriptsize[0.00, 11.03]}} & \makecell{0.00\\{\scriptsize[0.00, 8.76]}} & \makecell{0.00\\{\scriptsize[0.00, 8.76]}} & \makecell{0.00\\{\scriptsize[0.00, 10.43]}} & \makecell{0.00\\{\scriptsize[0.00, 10.43]}} & \makecell{0.00\\{\scriptsize[0.00, 2.12]}} \\[3pt]
    \cmidrule(l){2-8}
    \multirow{5}{*}[-2pt]{\model{Gemini-3.8-Flash}} & YES/NO & \makecell{89.74\\{\scriptsize[76.42, 95.94]}} & \makecell{90.62\\{\scriptsize[75.78, 96.76]}} & \makecell{90.62\\{\scriptsize[75.78, 96.76]}} & \makecell{96.77\\{\scriptsize[83.81, 99.43]}} & \makecell{69.57\\{\scriptsize[49.13, 84.40]}} & \makecell{88.54\\{\scriptsize[82.61, 92.62]}} \\
    \addlinespace[2pt]
     & Multiple choice & \makecell{96.67\\{\scriptsize[83.33, 99.41]}} & \makecell{96.43\\{\scriptsize[82.29, 99.37]}} & \makecell{96.43\\{\scriptsize[82.29, 99.37]}} & \makecell{100.00\\{\scriptsize[90.36, 100.00]}} & \makecell{97.73\\{\scriptsize[88.19, 99.60]}} & \makecell{97.59\\{\scriptsize[93.97, 99.06]}} \\
    \addlinespace[2pt]
     & Counting & \makecell{61.29\\{\scriptsize[43.82, 76.27]}} & \makecell{40.00\\{\scriptsize[26.35, 55.40]}} & \makecell{52.50\\{\scriptsize[37.50, 67.06]}} & \makecell{66.67\\{\scriptsize[49.61, 80.25]}} & \makecell{42.42\\{\scriptsize[27.24, 59.19]}} & \makecell{51.98\\{\scriptsize[44.65, 59.22]}} \\[3pt]
    \cmidrule(l){2-8}
    \multirow{5}{*}[-2pt]{\model{Qwen3.5-35B-A3B}} & YES/NO & \makecell{56.41\\{\scriptsize[40.98, 70.70]}} & \makecell{71.88\\{\scriptsize[54.63, 84.44]}} & \makecell{59.38\\{\scriptsize[42.26, 74.48]}} & \makecell{67.74\\{\scriptsize[50.14, 81.43]}} & \makecell{56.52\\{\scriptsize[36.81, 74.37]}} & \makecell{62.42\\{\scriptsize[54.63, 69.61]}} \\
    \addlinespace[2pt]
     & Multiple choice & \makecell{53.33\\{\scriptsize[36.14, 69.77]}} & \makecell{67.86\\{\scriptsize[49.34, 82.07]}} & \makecell{75.00\\{\scriptsize[56.64, 87.32]}} & \makecell{58.33\\{\scriptsize[42.20, 72.86]}} & \makecell{75.00\\{\scriptsize[60.56, 85.43]}} & \makecell{66.27\\{\scriptsize[58.78, 73.02]}} \\
    \addlinespace[2pt]
     & Counting & \makecell{0.00\\{\scriptsize[0.00, 11.03]}} & \makecell{0.00\\{\scriptsize[0.00, 8.76]}} & \makecell{0.00\\{\scriptsize[0.00, 8.76]}} & \makecell{0.00\\{\scriptsize[0.00, 10.43]}} & \makecell{0.00\\{\scriptsize[0.00, 10.43]}} & \makecell{0.00\\{\scriptsize[0.00, 2.12]}} \\[3pt]
    \cmidrule(l){2-8}
    \multirow{3}{*}{Random guessing} & YES/NO & 50.00 & 50.00 & 50.00 & 50.00 & 50.00 & 50.00 \\
     & Multiple choice & 25.00 & 25.00 & 25.00 & 25.00 & 25.00 & 25.00 \\
     & Counting & 0.66 & 0.66 & 0.66 & 0.66 & 0.66 & 0.66 \\
    \midrule
    \multirow{3}{*}{\#Examples} & YES/NO & 39 & 32 & 32 & 31 & 23 & 157 \\
     & Multiple choice & 30 & 28 & 28 & 36 & 44 & 166 \\
     & Counting & 31 & 40 & 40 & 33 & 33 & 177 \\
    \bottomrule
  \end{tabular}
  \caption{Accuracy (\%) of each model on each task in the distractor type experiment (\S\ref{subsec:effects_of_distractor_type}), with reasoning effort \texttt{low}. Overall pools all distractor types. The number of examples per task differs between columns because the task of each example is sampled at random. Below each accuracy value is its 95\% Wilson confidence interval.}
  \label{tab:v1_accuracy_by_distractor_type_and_task}
\end{table*}

\begin{table*}[t]
  \centering
  \small
  \setlength{\tabcolsep}{5pt}
  \begin{tabular}{@{}lccc@{}}
    \toprule
    & \multicolumn{3}{c}{\textbf{Model}} \\
    \cmidrule(lr){2-4}
    \textbf{Distractor types} & \textbf{\model{GPT-5.6-Luna}} & \textbf{\model{Gemini-3.8-Flash}} & \textbf{\model{Qwen3.5-35B-A3B}} \\
    \midrule
    \texttt{books} vs \texttt{code} & \makecell{$+0.00$\\{\scriptsize $p$ = 1.00}} & \makecell{$-11.00$\\{\scriptsize $p$ = 0.63}} & \makecell{$+4.00$\\{\scriptsize $p$ = 1.00}} \\
    \addlinespace[2pt]
    \texttt{books} vs \texttt{disjoint-component} & \makecell{$+0.00$\\{\scriptsize $p$ = 1.00}} & \makecell{$-6.00$\\{\scriptsize $p$ = 1.00}} & \makecell{$+2.00$\\{\scriptsize $p$ = 1.00}} \\
    \addlinespace[2pt]
    \texttt{books} vs \texttt{essays} & \makecell{$+5.00$\\{\scriptsize $p$ = 1.00}} & \makecell{$+5.00$\\{\scriptsize $p$ = 1.00}} & \makecell{$+4.00$\\{\scriptsize $p$ = 1.00}} \\
    \addlinespace[2pt]
    \texttt{books} vs \texttt{wikipedia} & \makecell{$+9.00$\\{\scriptsize $p$ = 1.00}} & \makecell{$-10.00$\\{\scriptsize $p$ = 0.74}} & \makecell{$+8.00$\\{\scriptsize $p$ = 1.00}} \\
    \addlinespace[2pt]
    \texttt{code} vs \texttt{disjoint-component} & \makecell{$+0.00$\\{\scriptsize $p$ = 1.00}} & \makecell{$+5.00$\\{\scriptsize $p$ = 1.00}} & \makecell{$-2.00$\\{\scriptsize $p$ = 1.00}} \\
    \addlinespace[2pt]
    \texttt{code} vs \texttt{essays} & \makecell{$+5.00$\\{\scriptsize $p$ = 1.00}} & \makecell{$+16.00$\\{\scriptsize $p$ = 0.07}} & \makecell{$+0.00$\\{\scriptsize $p$ = 1.00}} \\
    \addlinespace[2pt]
    \texttt{code} vs \texttt{wikipedia} & \makecell{$+9.00$\\{\scriptsize $p$ = 1.00}} & \makecell{$+1.00$\\{\scriptsize $p$ = 1.00}} & \makecell{$+4.00$\\{\scriptsize $p$ = 1.00}} \\
    \addlinespace[2pt]
    \texttt{disjoint-component} vs \texttt{essays} & \makecell{$+5.00$\\{\scriptsize $p$ = 1.00}} & \makecell{$+11.00$\\{\scriptsize $p$ = 0.49}} & \makecell{$+2.00$\\{\scriptsize $p$ = 1.00}} \\
    \addlinespace[2pt]
    \texttt{disjoint-component} vs \texttt{wikipedia} & \makecell{$+9.00$\\{\scriptsize $p$ = 1.00}} & \makecell{$-4.00$\\{\scriptsize $p$ = 1.00}} & \makecell{$+6.00$\\{\scriptsize $p$ = 1.00}} \\
    \addlinespace[2pt]
    \texttt{essays} vs \texttt{wikipedia} & \makecell{$+4.00$\\{\scriptsize $p$ = 1.00}} & \makecell{$-15.00$\\{\scriptsize $p$ = 0.11}} & \makecell{$+4.00$\\{\scriptsize $p$ = 1.00}} \\
    \addlinespace[2pt]
    All five, $\chi^2$ test & $p$ = 0.61 & $p$ = 0.03 & $p$ = 0.84 \\
    \bottomrule
  \end{tabular}
  \caption{Pairwise comparison of the distractor types in the distractor type experiment (\S\ref{subsec:effects_of_distractor_type}), with reasoning effort \texttt{low}. All tasks are pooled, because the task of each example is sampled at random, independently of its distractor type. Each cell gives the accuracy with the second distractor type minus the accuracy with the first, in percentage points, and below it the $p$-value of Fisher's exact test after the Holm correction for the ten pairs of each model. The last row gives the $p$-value of the $\chi^2$ test of whether accuracy is the same for all five distractor types.}
  \label{tab:v1_distractor_type_tests}
\end{table*}

\subsection{Reasoning Effort}

\Cref{tab:v1_reasoning_effort_paired_comparison} compares each pair of reasoning efforts on the same 100 examples (\S\ref{subsec:effects_of_reasoning_effort}): for each pair, it gives the number of examples that become correct or wrong and the $p$-value of the exact McNemar test.

\begin{table*}[t]
  \centering
  \small
  \setlength{\tabcolsep}{5pt}
  \begin{tabular}{@{}lccc@{}}
    \toprule
    & \multicolumn{3}{c}{\textbf{Change of reasoning effort}} \\
    \cmidrule(lr){2-4}
    \textbf{Model} & \textbf{\texttt{low} $\rightarrow$ \texttt{medium}} & \textbf{\texttt{low} $\rightarrow$ \texttt{high}} & \textbf{\texttt{medium} $\rightarrow$ \texttt{high}} \\
    \midrule
    \model{GPT-5.6-Luna} & \makecell{$+$3 / $-$2\\{\scriptsize $p$ = 1.00}} & \makecell{$+$10 / $-$0\\{\scriptsize $p$ = $<$0.01}} & \makecell{$+$10 / $-$1\\{\scriptsize $p$ = 0.01}} \\
    \addlinespace[2pt]
    \model{Gemini-3.8-Flash} & \makecell{$+$12 / $-$8\\{\scriptsize $p$ = 0.50}} & \makecell{$+$7 / $-$12\\{\scriptsize $p$ = 0.36}} & \makecell{$+$2 / $-$11\\{\scriptsize $p$ = 0.02}} \\
    \addlinespace[2pt]
    \model{Qwen3.5-35B-A3B} & \makecell{$+$3 / $-$4\\{\scriptsize $p$ = 1.00}} & \makecell{$+$4 / $-$4\\{\scriptsize $p$ = 1.00}} & \makecell{$+$5 / $-$4\\{\scriptsize $p$ = 1.00}} \\[3pt]
    \bottomrule
  \end{tabular}
  \caption{Paired comparison of the reasoning efforts in the reasoning effort experiment (\S\ref{subsec:effects_of_reasoning_effort}). Each model answers the same 100 examples at every effort. Each cell gives the number of examples that become correct ($+$) and that become wrong ($-$) when the effort changes as in the column header, and below it the $p$-value of the exact McNemar test.}
  \label{tab:v1_reasoning_effort_paired_comparison}
\end{table*}

\subsection{Context Length Tests}

\Cref{tab:v1_context_length_tests} compares the accuracy at each pair of target context lengths (\S\ref{subsec:effects_of_context_length}) with Fisher's exact test. Because the examples differ between context lengths and share at most their settings, the test treats the two sets of answers as independent samples instead of pairs.

\begin{table*}[t]
  \centering
  \small
  \setlength{\tabcolsep}{5pt}
  \begin{tabular}{@{}lccc@{}}
    \toprule
    & \multicolumn{3}{c}{\textbf{Change of target context length}} \\
    \cmidrule(lr){2-4}
    \textbf{Model} & \textbf{200k $\rightarrow$ 500k} & \textbf{200k $\rightarrow$ 922k} & \textbf{500k $\rightarrow$ 922k} \\
    \midrule
    \model{GPT-5.6-Luna} & \makecell{38.00 $\rightarrow$ 44.00\\{\scriptsize $p$ = 0.49}} & \makecell{38.00 $\rightarrow$ 40.00\\{\scriptsize $p$ = 0.86}} & \makecell{44.00 $\rightarrow$ 40.00\\{\scriptsize $p$ = 0.84}} \\
    \addlinespace[2pt]
    \model{Gemini-3.8-Flash} & \makecell{71.00 $\rightarrow$ 64.00\\{\scriptsize $p$ = 0.46}} & \makecell{71.00 $\rightarrow$ 60.00\\{\scriptsize $p$ = 0.20}} & \makecell{64.00 $\rightarrow$ 60.00\\{\scriptsize $p$ = 0.84}} \\[3pt]
    \bottomrule
  \end{tabular}
  \caption{Fisher's exact test of whether accuracy in the target context length experiment (\S\ref{subsec:effects_of_context_length}) differs between two target context lengths, with reasoning effort \texttt{low}. Each cell gives the accuracy (\%) at the two context lengths and below it the two-sided $p$-value. Each target context length has its own examples, so the test treats the answers at two lengths as independent samples. Examples per target context length: 100 (200k), 50 (500k), 50 (922k).}
  \label{tab:v1_context_length_tests}
\end{table*}

\subsection{Full Experimental Results by Experiments}
\label{appdx:full_experimental_results_by_experiments}

Details of the experimental results are provided in \Cref{tab:v1_accuracy_by_question_depth,tab:v1_accuracy_by_node_count,tab:v1_accuracy_by_context_length,tab:v1_accuracy_by_question_format,tab:v1_accuracy_by_distractor_type,tab:v1_accuracy_by_reasoning_effort}. Each table shows the accuracy of each model for each value of the studied variable, along with 95\% Wilson confidence intervals. The tables also include the expected accuracy of random guessing, which serves as a baseline for comparison.

\begin{table*}[t]
  \centering
  \small
  \setlength{\tabcolsep}{5pt}
  \begin{tabular}{@{}lccccc@{}}
    \toprule
    & \multicolumn{4}{c}{\textbf{Question depth $k$}} & \\
    \cmidrule(lr){2-5}
    \textbf{Model} & \textbf{4} & \textbf{8} & \textbf{16} & \textbf{32} & \textbf{Overall} \\
    \midrule
    \model{GPT-5.6-Sol} & -- & \makecell{75.00\\{\scriptsize[65.70, 82.45]}} & \makecell{56.00\\{\scriptsize[46.23, 65.33]}} & \makecell{54.00\\{\scriptsize[44.26, 63.44]}} & \makecell{61.67$^\dagger$\\{\scriptsize[56.05, 66.99]}} \\
    \addlinespace[2pt]
    \model{GPT-5.6-Luna} & \makecell{66.00\\{\scriptsize[56.28, 74.54]}} & \makecell{39.00\\{\scriptsize[30.02, 48.80]}} & \makecell{37.00\\{\scriptsize[28.18, 46.78]}} & \makecell{34.00\\{\scriptsize[25.46, 43.72]}} & \makecell{44.00\\{\scriptsize[39.22, 48.90]}} \\
    \addlinespace[2pt]
    \model{Gemini-3.1-Pro} & \makecell{71.00\\{\scriptsize[61.46, 78.99]}} & \makecell{62.00\\{\scriptsize[52.21, 70.90]}} & \makecell{57.00\\{\scriptsize[47.22, 66.27]}} & \makecell{52.00\\{\scriptsize[42.32, 61.54]}} & \makecell{60.50\\{\scriptsize[55.63, 65.17]}} \\
    \addlinespace[2pt]
    \model{Gemini-3.8-Flash} & \makecell{\textbf{87.00}\\{\scriptsize[79.02, 92.24]}} & \makecell{\textbf{83.00}\\{\scriptsize[74.45, 89.11]}} & \makecell{\textbf{87.00}\\{\scriptsize[79.02, 92.24]}} & \makecell{\textbf{59.00}\\{\scriptsize[49.20, 68.13]}} & \makecell{\textbf{79.00}\\{\scriptsize[74.74, 82.71]}} \\
    \addlinespace[2pt]
    \model{Gemini-3.6-Flash} & \makecell{75.00\\{\scriptsize[65.70, 82.45]}} & \makecell{56.00\\{\scriptsize[46.23, 65.33]}} & \makecell{47.00\\{\scriptsize[37.51, 56.71]}} & \makecell{41.00\\{\scriptsize[31.87, 50.80]}} & \makecell{54.75\\{\scriptsize[49.85, 59.56]}} \\
    \addlinespace[2pt]
    \model{Qwen3.5-397B-A17B} & \makecell{65.00\\{\scriptsize[55.25, 73.64]}} & \makecell{48.00\\{\scriptsize[38.46, 57.68]}} & \makecell{42.00\\{\scriptsize[32.80, 51.79]}} & \makecell{38.00\\{\scriptsize[29.10, 47.79]}} & \makecell{48.25\\{\scriptsize[43.39, 53.14]}} \\
    \addlinespace[2pt]
    \model{Qwen3.5-122B-A10B} & \makecell{65.00\\{\scriptsize[55.25, 73.64]}} & \makecell{45.00\\{\scriptsize[35.61, 54.76]}} & \makecell{40.00\\{\scriptsize[30.94, 49.80]}} & \makecell{37.00\\{\scriptsize[28.18, 46.78]}} & \makecell{46.75\\{\scriptsize[41.91, 51.65]}} \\
    \addlinespace[2pt]
    \model{Qwen3.5-35B-A3B} & \makecell{63.00\\{\scriptsize[53.22, 71.82]}} & \makecell{42.00\\{\scriptsize[32.80, 51.79]}} & \makecell{37.00\\{\scriptsize[28.18, 46.78]}} & \makecell{39.00\\{\scriptsize[30.02, 48.80]}} & \makecell{45.25\\{\scriptsize[40.44, 50.15]}} \\[3pt]
    \midrule
    Random guessing & 25.22 & 25.22 & 25.22 & 25.22 & 25.22 \\
    \bottomrule
  \end{tabular}
  \caption{Accuracy (\%) of each model in the question depth experiment (\S\ref{subsec:complexity_scaling}), with reasoning effort \texttt{low}. Each question depth has 100 examples, and Overall pools all question depths. Below each accuracy value is its 95\% Wilson confidence interval. Bold marks the highest accuracy in each column. $^\dagger$\model{GPT-5.6-Sol} has fewer examples than the other models in this cell (300 Overall), so it is not bold. A dash marks a missing result.}
  \label{tab:v1_accuracy_by_question_depth}
\end{table*}

\begin{table*}[t]
  \centering
  \small
  \setlength{\tabcolsep}{5pt}
  \begin{tabular}{@{}lccccc@{}}
    \toprule
    & \multicolumn{4}{c}{\textbf{Component size}} & \\
    \cmidrule(lr){2-5}
    \textbf{Model} & \textbf{100} & \textbf{150} & \textbf{200} & \textbf{250} & \textbf{Overall} \\
    \midrule
    \model{GPT-5.6-Sol} & \makecell{77.78$^\dagger$\\{\scriptsize[59.24, 89.39]}} & \makecell{61.54$^\dagger$\\{\scriptsize[42.53, 77.57]}} & \makecell{60.00\\{\scriptsize[50.20, 69.06]}} & \makecell{49.00\\{\scriptsize[39.42, 58.65]}} & \makecell{57.71$^\dagger$\\{\scriptsize[51.55, 63.64]}} \\
    \addlinespace[2pt]
    \model{GPT-5.6-Luna} & \makecell{44.00\\{\scriptsize[34.67, 53.77]}} & \makecell{38.00\\{\scriptsize[29.10, 47.79]}} & \makecell{42.00\\{\scriptsize[32.80, 51.79]}} & \makecell{37.00\\{\scriptsize[28.18, 46.78]}} & \makecell{40.25\\{\scriptsize[35.56, 45.13]}} \\
    \addlinespace[2pt]
    \model{Gemini-3.1-Pro} & \makecell{74.00\\{\scriptsize[64.63, 81.60]}} & \makecell{57.00\\{\scriptsize[47.22, 66.27]}} & \makecell{55.00\\{\scriptsize[45.24, 64.39]}} & \makecell{46.00\\{\scriptsize[36.56, 55.74]}} & \makecell{58.00\\{\scriptsize[53.11, 62.74]}} \\
    \addlinespace[2pt]
    \model{Gemini-3.8-Flash} & \makecell{\textbf{88.00}\\{\scriptsize[80.19, 93.00]}} & \makecell{\textbf{77.00}\\{\scriptsize[67.85, 84.16]}} & \makecell{\textbf{76.00}\\{\scriptsize[66.77, 83.31]}} & \makecell{\textbf{67.00}\\{\scriptsize[57.31, 75.44]}} & \makecell{\textbf{77.00}\\{\scriptsize[72.63, 80.86]}} \\
    \addlinespace[2pt]
    \model{Gemini-3.6-Flash} & \makecell{69.00\\{\scriptsize[59.37, 77.22]}} & \makecell{52.00\\{\scriptsize[42.32, 61.54]}} & \makecell{54.00\\{\scriptsize[44.26, 63.44]}} & \makecell{43.00\\{\scriptsize[33.73, 52.78]}} & \makecell{54.50\\{\scriptsize[49.60, 59.31]}} \\
    \addlinespace[2pt]
    \model{Qwen3.5-397B-A17B} & \makecell{54.00\\{\scriptsize[44.26, 63.44]}} & \makecell{40.00\\{\scriptsize[30.94, 49.80]}} & \makecell{43.00\\{\scriptsize[33.73, 52.78]}} & \makecell{40.00\\{\scriptsize[30.94, 49.80]}} & \makecell{44.25\\{\scriptsize[39.46, 49.15]}} \\
    \addlinespace[2pt]
    \model{Qwen3.5-122B-A10B} & \makecell{50.00\\{\scriptsize[40.38, 59.62]}} & \makecell{40.00\\{\scriptsize[30.94, 49.80]}} & \makecell{41.00\\{\scriptsize[31.87, 50.80]}} & \makecell{36.00\\{\scriptsize[27.27, 45.76]}} & \makecell{41.75\\{\scriptsize[37.02, 46.64]}} \\
    \addlinespace[2pt]
    \model{Qwen3.5-35B-A3B} & \makecell{43.00\\{\scriptsize[33.73, 52.78]}} & \makecell{38.00\\{\scriptsize[29.10, 47.79]}} & \makecell{42.00\\{\scriptsize[32.80, 51.79]}} & \makecell{37.00\\{\scriptsize[28.18, 46.78]}} & \makecell{40.00\\{\scriptsize[35.32, 44.87]}} \\[3pt]
    \midrule
    Random guessing & 25.22 & 25.22 & 25.22 & 25.22 & 25.22 \\
    \bottomrule
  \end{tabular}
  \caption{Accuracy (\%) of each model in the component size experiment (\S\ref{subsec:complexity_scaling}), with reasoning effort \texttt{low}. Each component size has 100 examples, and Overall pools all component sizes. Below each accuracy value is its 95\% Wilson confidence interval. Bold marks the highest accuracy in each column. $^\dagger$\model{GPT-5.6-Sol} has fewer examples than the other models in these cells (27 at 100, 26 at 150, and 253 Overall), so they are not bold.}
  \label{tab:v1_accuracy_by_node_count}
\end{table*}

\begin{table*}[t]
  \centering
  \small
  \setlength{\tabcolsep}{5pt}
  \begin{tabular}{@{}lccc@{}}
    \toprule
    & \multicolumn{3}{c}{\textbf{Target context length (tokens)}} \\
    \cmidrule(lr){2-4}
    \textbf{Model} & \textbf{200k} & \textbf{500k} & \textbf{922k} \\
    \midrule
    \model{GPT-5.6-Luna} & \makecell{38.00\\{\scriptsize[29.10, 47.79]}} & \makecell{44.00\\{\scriptsize[31.16, 57.69]}} & \makecell{40.00\\{\scriptsize[27.61, 53.82]}} \\
    \addlinespace[2pt]
    \model{Gemini-3.8-Flash} & \makecell{\textbf{71.00}\\{\scriptsize[61.46, 78.99]}} & \makecell{\textbf{64.00}\\{\scriptsize[50.14, 75.86]}} & \makecell{\textbf{60.00}\\{\scriptsize[46.18, 72.39]}} \\[3pt]
    \midrule
    Random guessing & 25.22 & 25.22 & 25.22 \\
    \midrule
    \#Examples & 100 & 50 & 50 \\
    \bottomrule
  \end{tabular}
  \caption{Accuracy (\%) of each model in the target context length experiment (\S\ref{subsec:effects_of_context_length}), with reasoning effort \texttt{low}. The 200k results reuse the examples and answers of the reasoning effort experiment. Below each accuracy value is its 95\% Wilson confidence interval. Bold marks the highest accuracy in each column.}
  \label{tab:v1_accuracy_by_context_length}
\end{table*}

\begin{table*}[t]
  \centering
  \small
  \setlength{\tabcolsep}{5pt}
  \begin{tabular}{@{}lcccc@{}}
    \toprule
    & \multicolumn{3}{c}{\textbf{Task}} & \\
    \cmidrule(lr){2-4}
    \textbf{Model} & \textbf{YES/NO} & \textbf{Multiple choice} & \textbf{Counting} & \textbf{Overall} \\
    \midrule
    \model{GPT-5.6-Luna} & \makecell{66.00\\{\scriptsize[56.28, 74.54]}} & \makecell{68.00\\{\scriptsize[58.34, 76.33]}} & \makecell{0.00\\{\scriptsize[0.00, 3.70]}} & \makecell{44.67\\{\scriptsize[39.14, 50.32]}} \\
    \addlinespace[2pt]
    \model{Gemini-3.8-Flash} & \makecell{\textbf{87.00}\\{\scriptsize[79.02, 92.24]}} & \makecell{\textbf{97.00}\\{\scriptsize[91.55, 98.97]}} & \makecell{\textbf{55.00}\\{\scriptsize[45.24, 64.39]}} & \makecell{\textbf{79.67}\\{\scriptsize[74.75, 83.83]}} \\
    \addlinespace[2pt]
    \model{Qwen3.5-35B-A3B} & \makecell{64.00\\{\scriptsize[54.24, 72.73]}} & \makecell{72.00\\{\scriptsize[62.51, 79.86]}} & \makecell{0.00\\{\scriptsize[0.00, 3.70]}} & \makecell{45.33\\{\scriptsize[39.79, 50.99]}} \\[3pt]
    \midrule
    Random guessing & 50.00 & 25.00 & 0.66 & 25.22 \\
    \bottomrule
  \end{tabular}
  \caption{Accuracy (\%) of each model in the task experiment (\S\ref{subsec:performance_on_different_reachability_tasks}), with reasoning effort \texttt{low}. Each task has 100 examples, and Overall pools all three tasks. Below each accuracy value is its 95\% Wilson confidence interval. Bold marks the highest accuracy in each column.}
  \label{tab:v1_accuracy_by_question_format}
\end{table*}

\begin{table*}[t]
  \centering
  \small
  \setlength{\tabcolsep}{5pt}
  \begin{tabular}{@{}lcccccc@{}}
    \toprule
    & \multicolumn{5}{c}{\textbf{Distractor type}} & \\
    \cmidrule(lr){2-6}
    \textbf{Model} & \textbf{\texttt{books}} & \textbf{\texttt{code}} & \textbf{\texttt{disjoint-component}} & \textbf{\texttt{essays}} & \textbf{\texttt{wikipedia}} & \textbf{Overall} \\
    \midrule
    \model{GPT-5.6-Luna} & \makecell{41.00\\{\scriptsize[31.87, 50.80]}} & \makecell{41.00\\{\scriptsize[31.87, 50.80]}} & \makecell{41.00\\{\scriptsize[31.87, 50.80]}} & \makecell{46.00\\{\scriptsize[36.56, 55.74]}} & \makecell{50.00\\{\scriptsize[40.38, 59.62]}} & \makecell{43.80\\{\scriptsize[39.51, 48.18]}} \\
    \addlinespace[2pt]
    \model{Gemini-3.8-Flash} & \makecell{\textbf{83.00}\\{\scriptsize[74.45, 89.11]}} & \makecell{\textbf{72.00}\\{\scriptsize[62.51, 79.86]}} & \makecell{\textbf{77.00}\\{\scriptsize[67.85, 84.16]}} & \makecell{\textbf{88.00}\\{\scriptsize[80.19, 93.00]}} & \makecell{\textbf{73.00}\\{\scriptsize[63.57, 80.73]}} & \makecell{\textbf{78.60}\\{\scriptsize[74.79, 81.97]}} \\
    \addlinespace[2pt]
    \model{Qwen3.5-35B-A3B} & \makecell{38.00\\{\scriptsize[29.10, 47.79]}} & \makecell{42.00\\{\scriptsize[32.80, 51.79]}} & \makecell{40.00\\{\scriptsize[30.94, 49.80]}} & \makecell{42.00\\{\scriptsize[32.80, 51.79]}} & \makecell{46.00\\{\scriptsize[36.56, 55.74]}} & \makecell{41.60\\{\scriptsize[37.36, 45.97]}} \\[3pt]
    \midrule
    Random guessing & 25.22 & 25.22 & 25.22 & 25.22 & 25.22 & 25.22 \\
    \bottomrule
  \end{tabular}
  \caption{Accuracy (\%) of each model in the distractor type experiment (\S\ref{subsec:effects_of_distractor_type}), with reasoning effort \texttt{low}. Each distractor type has 100 examples, and Overall pools all distractor types. Below each accuracy value is its 95\% Wilson confidence interval. Bold marks the highest accuracy in each column.}
  \label{tab:v1_accuracy_by_distractor_type}
\end{table*}

\begin{table*}[t]
  \centering
  \small
  \setlength{\tabcolsep}{5pt}
  \begin{tabular}{@{}lccc@{}}
    \toprule
    & \multicolumn{3}{c}{\textbf{Reasoning effort}} \\
    \cmidrule(lr){2-4}
    \textbf{Model} & \textbf{\texttt{low}} & \textbf{\texttt{medium}} & \textbf{\texttt{high}} \\
    \midrule
    \model{GPT-5.6-Luna} & \makecell{38.00\\{\scriptsize[29.10, 47.79]}} & \makecell{39.00\\{\scriptsize[30.02, 48.80]}} & \makecell{48.00\\{\scriptsize[38.46, 57.68]}} \\
    \addlinespace[2pt]
    \model{Gemini-3.8-Flash} & \makecell{\textbf{71.00}\\{\scriptsize[61.46, 78.99]}} & \makecell{\textbf{75.00}\\{\scriptsize[65.70, 82.45]}} & \makecell{\textbf{66.00}\\{\scriptsize[56.28, 74.54]}} \\
    \addlinespace[2pt]
    \model{Qwen3.5-35B-A3B} & \makecell{40.00\\{\scriptsize[30.94, 49.80]}} & \makecell{39.00\\{\scriptsize[30.02, 48.80]}} & \makecell{40.00\\{\scriptsize[30.94, 49.80]}} \\[3pt]
    \midrule
    Random guessing & 25.22 & 25.22 & 25.22 \\
    \bottomrule
  \end{tabular}
  \caption{Accuracy (\%) of each model in the reasoning effort experiment (\S\ref{subsec:effects_of_reasoning_effort}). All reasoning efforts use the same 100 examples. Below each accuracy value is its 95\% Wilson confidence interval. Bold marks the highest accuracy in each column.}
  \label{tab:v1_accuracy_by_reasoning_effort}
\end{table*}

\section{Inference Cost and Response Time}
\label{appdx:cost_and_response_time}

\Cref{tab:v1_cost_and_response_time_question_depth,tab:v1_cost_and_response_time_node_count,tab:v1_cost_and_response_time_question_format,tab:v1_cost_and_response_time_distractor_type,tab:v1_cost_and_response_time_reasoning_effort,tab:v1_cost_and_response_time_context_length} report, for each experiment, the number of answered examples, the output tokens, the response time, and the API cost of each model.

\begin{table*}[t]
  \centering
  \small
  \begin{tabular}{@{}lrrrrr@{}}
    \toprule
    \textbf{Model} & \textbf{\#Examples} & \makecell[r]{\textbf{Output}\\\textbf{tokens}} & \makecell[r]{\textbf{Response}\\\textbf{time (s)}} & \makecell[r]{\textbf{Cost per}\\\textbf{example (\textcent)}} & \makecell[r]{\textbf{Total}\\\textbf{cost (\$)}} \\
    \midrule
    \model{GPT-5.6-Sol}$^*$ & 300 & 1,340.02\textsubscript{$\pm$1,096.37} & 18.97\textsubscript{$\pm$12.86} & 44.12\textsubscript{$\pm$6.39} & 132.35 \\
    \model{GPT-5.6-Luna}$^*$ & 400 & 715.69\textsubscript{$\pm$529.55} & 12.66\textsubscript{$\pm$10.02} & 2.19\textsubscript{$\pm$0.29} & 8.77 \\
    \model{Gemini-3.1-Pro}$^*$ & 400 & 4,642.54\textsubscript{$\pm$4,621.15} & 92.16\textsubscript{$\pm$99.79} & 20.88\textsubscript{$\pm$3.81} & 83.54 \\
    \model{Gemini-3.8-Flash} & 400 & 19,412.21\textsubscript{$\pm$20,608.90} & 104.96\textsubscript{$\pm$104.96} & 20.91\textsubscript{$\pm$8.19} & 83.62 \\
    \model{Gemini-3.6-Flash}$^*$ & 400 & 10,410.75\textsubscript{$\pm$12,035.62} & 61.35\textsubscript{$\pm$59.88} & 7.79\textsubscript{$\pm$3.68} & 31.16 \\
    \model{Qwen3.5-397B-A17B} & 400 & 15,729.28\textsubscript{$\pm$7,681.13} & 439.19\textsubscript{$\pm$371.56} & 12.21\textsubscript{$\pm$3.08} & 48.85 \\
    \model{Qwen3.5-122B-A10B} & 400 & 14,843.83\textsubscript{$\pm$6,818.70} & 210.63\textsubscript{$\pm$79.53} & 7.87\textsubscript{$\pm$1.56} & 31.47 \\
    \model{Qwen3.5-35B-A3B} & 400 & 16,387.97\textsubscript{$\pm$11,616.19} & 162.29\textsubscript{$\pm$99.04} & 5.10\textsubscript{$\pm$1.54} & 20.41 \\
    \midrule
    \textbf{All models} &  &  &  &  & 440.17 \\
    \bottomrule
  \end{tabular}
  \caption{Cost and response time of each model in the question depth experiment (\S\ref{subsec:complexity_scaling}), with reasoning effort \texttt{low}. \#Examples counts the examples that the model answered, and each cell shows mean\textsubscript{$\pm$std} over these examples. Output tokens are the completion tokens reported by the provider, including reasoning tokens. Response time runs from sending the successful request to receiving its response, so it excludes waits before sending and failed attempts. Cost is the LiteLLM estimate from the token counts of each response and the provider's prices, in US cents per example and US dollars in total. $^*$Sent at the flex service tier, which has lower prices and can respond more slowly.}
  \label{tab:v1_cost_and_response_time_question_depth}
\end{table*}

\begin{table*}[t]
  \centering
  \small
  \begin{tabular}{@{}lrrrrr@{}}
    \toprule
    \textbf{Model} & \textbf{\#Examples} & \makecell[r]{\textbf{Output}\\\textbf{tokens}} & \makecell[r]{\textbf{Response}\\\textbf{time (s)}} & \makecell[r]{\textbf{Cost per}\\\textbf{example (\textcent)}} & \makecell[r]{\textbf{Total}\\\textbf{cost (\$)}} \\
    \midrule
    \model{GPT-5.6-Sol}$^*$ & 253 & 1,424.49\textsubscript{$\pm$1,282.98} & 21.68\textsubscript{$\pm$16.10} & 41.83\textsubscript{$\pm$10.64} & 105.83 \\
    \model{GPT-5.6-Luna}$^*$ & 400 & 785.28\textsubscript{$\pm$541.69} & 11.99\textsubscript{$\pm$7.95} & 2.16\textsubscript{$\pm$0.33} & 8.66 \\
    \model{Gemini-3.1-Pro}$^*$ & 400 & 4,744.15\textsubscript{$\pm$5,510.22} & 42.75\textsubscript{$\pm$45.80} & 20.78\textsubscript{$\pm$4.50} & 83.13 \\
    \model{Gemini-3.8-Flash} & 400 & 20,513.35\textsubscript{$\pm$21,872.99} & 120.58\textsubscript{$\pm$122.39} & 21.20\textsubscript{$\pm$8.44} & 84.79 \\
    \model{Gemini-3.6-Flash}$^*$ & 400 & 11,143.68\textsubscript{$\pm$12,347.34} & 58.21\textsubscript{$\pm$60.78} & 6.89\textsubscript{$\pm$3.56} & 27.56 \\
    \model{Qwen3.5-397B-A17B} & 400 & 15,841.35\textsubscript{$\pm$8,552.94} & 350.65\textsubscript{$\pm$214.14} & 11.85\textsubscript{$\pm$3.05} & 47.39 \\
    \model{Qwen3.5-122B-A10B} & 400 & 15,186.61\textsubscript{$\pm$6,564.43} & 213.11\textsubscript{$\pm$85.54} & 7.90\textsubscript{$\pm$1.52} & 31.61 \\
    \model{Qwen3.5-35B-A3B} & 400 & 16,510.21\textsubscript{$\pm$10,638.34} & 152.97\textsubscript{$\pm$81.26} & 5.08\textsubscript{$\pm$1.45} & 20.33 \\
    \midrule
    \textbf{All models} &  &  &  &  & 409.29 \\
    \bottomrule
  \end{tabular}
  \caption{Cost and response time of each model in the component size experiment (\S\ref{subsec:complexity_scaling}), with reasoning effort \texttt{low}. \#Examples counts the examples that the model answered, and each cell shows mean\textsubscript{$\pm$std} over these examples. Output tokens are the completion tokens reported by the provider, including reasoning tokens. Response time runs from sending the successful request to receiving its response, so it excludes waits before sending and failed attempts. Cost is the LiteLLM estimate from the token counts of each response and the provider's prices, in US cents per example and US dollars in total. $^*$Sent at the flex service tier, which has lower prices and can respond more slowly.}
  \label{tab:v1_cost_and_response_time_node_count}
\end{table*}

\begin{table*}[t]
  \centering
  \small
  \begin{tabular}{@{}lrrrrr@{}}
    \toprule
    \textbf{Model} & \textbf{\#Examples} & \makecell[r]{\textbf{Output}\\\textbf{tokens}} & \makecell[r]{\textbf{Response}\\\textbf{time (s)}} & \makecell[r]{\textbf{Cost per}\\\textbf{example (\textcent)}} & \makecell[r]{\textbf{Total}\\\textbf{cost (\$)}} \\
    \midrule
    \model{GPT-5.6-Luna}$^*$ & 300 & 738.38\textsubscript{$\pm$529.67} & 13.23\textsubscript{$\pm$9.33} & 2.17\textsubscript{$\pm$0.32} & 6.51 \\
    \model{Gemini-3.8-Flash} & 300 & 20,248.57\textsubscript{$\pm$21,838.14} & 114.54\textsubscript{$\pm$114.74} & 21.13\textsubscript{$\pm$8.54} & 63.39 \\
    \model{Qwen3.5-35B-A3B} & 300 & 16,471.95\textsubscript{$\pm$9,999.11} & 155.31\textsubscript{$\pm$76.50} & 5.08\textsubscript{$\pm$1.36} & 15.25 \\
    \midrule
    \textbf{All models} &  &  &  &  & 85.15 \\
    \bottomrule
  \end{tabular}
  \caption{Cost and response time of each model in the task experiment (\S\ref{subsec:performance_on_different_reachability_tasks}), with reasoning effort \texttt{low}. \#Examples counts the examples that the model answered, and each cell shows mean\textsubscript{$\pm$std} over these examples. Output tokens are the completion tokens reported by the provider, including reasoning tokens. Response time runs from sending the successful request to receiving its response, so it excludes waits before sending and failed attempts. Cost is the LiteLLM estimate from the token counts of each response and the provider's prices, in US cents per example and US dollars in total. $^*$Sent at the flex service tier, which has lower prices and can respond more slowly.}
  \label{tab:v1_cost_and_response_time_question_format}
\end{table*}

\begin{table*}[t]
  \centering
  \small
  \begin{tabular}{@{}lrrrrr@{}}
    \toprule
    \textbf{Model} & \textbf{\#Examples} & \makecell[r]{\textbf{Output}\\\textbf{tokens}} & \makecell[r]{\textbf{Response}\\\textbf{time (s)}} & \makecell[r]{\textbf{Cost per}\\\textbf{example (\textcent)}} & \makecell[r]{\textbf{Total}\\\textbf{cost (\$)}} \\
    \midrule
    \model{GPT-5.6-Luna}$^*$ & 500 & 759.95\textsubscript{$\pm$519.65} & 13.94\textsubscript{$\pm$8.26} & 2.16\textsubscript{$\pm$0.34} & 10.79 \\
    \model{Gemini-3.8-Flash} & 500 & 21,187.22\textsubscript{$\pm$21,680.07} & 124.62\textsubscript{$\pm$120.08} & 21.43\textsubscript{$\pm$8.57} & 107.17 \\
    \model{Qwen3.5-35B-A3B} & 500 & 17,419.60\textsubscript{$\pm$11,881.69} & 162.62\textsubscript{$\pm$96.15} & 5.17\textsubscript{$\pm$1.61} & 25.84 \\
    \midrule
    \textbf{All models} &  &  &  &  & 143.81 \\
    \bottomrule
  \end{tabular}
  \caption{Cost and response time of each model in the distractor type experiment (\S\ref{subsec:effects_of_distractor_type}), with reasoning effort \texttt{low}. \#Examples counts the examples that the model answered, and each cell shows mean\textsubscript{$\pm$std} over these examples. Output tokens are the completion tokens reported by the provider, including reasoning tokens. Response time runs from sending the successful request to receiving its response, so it excludes waits before sending and failed attempts. Cost is the LiteLLM estimate from the token counts of each response and the provider's prices, in US cents per example and US dollars in total. $^*$Sent at the flex service tier, which has lower prices and can respond more slowly.}
  \label{tab:v1_cost_and_response_time_distractor_type}
\end{table*}

\begin{table*}[t]
  \centering
  \small
  \begin{tabular}{@{}llrrrrr@{}}
    \toprule
    \textbf{Model} & \textbf{Reasoning effort} & \textbf{\#Examples} & \makecell[r]{\textbf{Output}\\\textbf{tokens}} & \makecell[r]{\textbf{Response}\\\textbf{time (s)}} & \makecell[r]{\textbf{Cost per}\\\textbf{example (\textcent)}} & \makecell[r]{\textbf{Total}\\\textbf{cost (\$)}} \\
    \midrule
    \multirow{3}{*}{\model{GPT-5.6-Luna}$^*$} & \texttt{low} & 100 & 823.21\textsubscript{$\pm$539.75} & 13.18\textsubscript{$\pm$8.27} & 1.95\textsubscript{$\pm$0.68} & 1.95 \\
     & \texttt{medium} & 100 & 1,937.24\textsubscript{$\pm$1,427.19} & 25.75\textsubscript{$\pm$18.32} & 2.26\textsubscript{$\pm$0.26} & 2.26 \\
     & \texttt{high} & 100 & 6,672.60\textsubscript{$\pm$6,128.72} & 61.73\textsubscript{$\pm$50.50} & 2.54\textsubscript{$\pm$0.45} & 2.54 \\[3pt]
    \cmidrule(l){2-7}
    \multirow{3}{*}{\model{Gemini-3.8-Flash}} & \texttt{low} & 100 & 23,086.40\textsubscript{$\pm$23,049.86} & 137.27\textsubscript{$\pm$127.62} & 22.46\textsubscript{$\pm$8.82} & 22.46 \\
     & \texttt{medium} & 100 & 29,207.72\textsubscript{$\pm$25,517.02} & 171.27\textsubscript{$\pm$141.94} & 24.75\textsubscript{$\pm$9.68} & 24.75 \\
     & \texttt{high} & 100 & 37,624.94\textsubscript{$\pm$29,513.48} & 204.32\textsubscript{$\pm$154.93} & 27.11\textsubscript{$\pm$10.42} & 27.11 \\[3pt]
    \cmidrule(l){2-7}
    \multirow{3}{*}{\model{Qwen3.5-35B-A3B}} & \texttt{low} & 100 & 15,219.30\textsubscript{$\pm$5,487.97} & 223.67\textsubscript{$\pm$89.98} & 4.92\textsubscript{$\pm$0.84} & 4.92 \\
     & \texttt{medium} & 100 & 16,866.72\textsubscript{$\pm$9,704.88} & 195.15\textsubscript{$\pm$95.51} & 5.14\textsubscript{$\pm$1.35} & 5.14 \\
     & \texttt{high} & 100 & 16,423.57\textsubscript{$\pm$6,641.77} & 172.53\textsubscript{$\pm$65.77} & 5.07\textsubscript{$\pm$0.96} & 5.07 \\
    \midrule
    \textbf{All models} &  &  &  &  &  & 96.20 \\
    \bottomrule
  \end{tabular}
  \caption{Cost and response time of each model in the reasoning effort experiment (\S\ref{subsec:effects_of_reasoning_effort}). \#Examples counts the examples that the model answered, and each cell shows mean\textsubscript{$\pm$std} over these examples. Output tokens are the completion tokens reported by the provider, including reasoning tokens. Response time runs from sending the successful request to receiving its response, so it excludes waits before sending and failed attempts. Cost is the LiteLLM estimate from the token counts of each response and the provider's prices, in US cents per example and US dollars in total. $^*$Sent at the flex service tier, which has lower prices and can respond more slowly.}
  \label{tab:v1_cost_and_response_time_reasoning_effort}
\end{table*}

\begin{table*}[t]
  \centering
  \small
  \begin{tabular}{@{}llrrrrr@{}}
    \toprule
    \textbf{Model} & \textbf{Context length} & \textbf{\#Examples} & \makecell[r]{\textbf{Output}\\\textbf{tokens}} & \makecell[r]{\textbf{Response}\\\textbf{time (s)}} & \makecell[r]{\textbf{Cost per}\\\textbf{example (\textcent)}} & \makecell[r]{\textbf{Total}\\\textbf{cost (\$)}} \\
    \midrule
    \multirow{3}{*}{\model{GPT-5.6-Luna}$^*$} & 200k & 100 & 823.21\textsubscript{$\pm$539.75} & 13.18\textsubscript{$\pm$8.27} & 1.95\textsubscript{$\pm$0.68} & 1.95 \\
     & 500k & 50 & 748.56\textsubscript{$\pm$459.06} & 14.51\textsubscript{$\pm$6.13} & 11.13\textsubscript{$\pm$0.93} & 5.57 \\
     & 922k & 50 & 844.76\textsubscript{$\pm$544.15} & 24.29\textsubscript{$\pm$8.91} & 20.60\textsubscript{$\pm$1.76} & 10.30 \\[3pt]
    \cmidrule(l){2-7}
    \multirow{3}{*}{\model{Gemini-3.8-Flash}} & 200k & 100 & 23,086.40\textsubscript{$\pm$23,049.86} & 137.27\textsubscript{$\pm$127.62} & 22.46\textsubscript{$\pm$8.82} & 22.46 \\
     & 500k & 50 & 31,639.22\textsubscript{$\pm$25,951.47} & 192.96\textsubscript{$\pm$146.21} & 47.85\textsubscript{$\pm$9.81} & 23.93 \\
     & 922k & 50 & 32,091.20\textsubscript{$\pm$26,945.34} & 205.07\textsubscript{$\pm$149.79} & 78.99\textsubscript{$\pm$10.16} & 39.49 \\[3pt]
    \cmidrule(l){2-7}
    \multirow{1}{*}{\model{Qwen3.5-35B-A3B}} & 200k & 100 & 15,219.30\textsubscript{$\pm$5,487.97} & 223.67\textsubscript{$\pm$89.98} & 4.92\textsubscript{$\pm$0.84} & 4.92 \\
    \midrule
    \textbf{All models} &  &  &  &  &  & 108.62 \\
    \bottomrule
  \end{tabular}
  \caption{Cost and response time of each model in the target context length experiment (\S\ref{subsec:effects_of_context_length}), with reasoning effort \texttt{low}. The 200k rows reuse the answers of the reasoning effort experiment. \#Examples counts the examples that the model answered, and each cell shows mean\textsubscript{$\pm$std} over these examples. Output tokens are the completion tokens reported by the provider, including reasoning tokens. Response time runs from sending the successful request to receiving its response, so it excludes waits before sending and failed attempts. Cost is the LiteLLM estimate from the token counts of each response and the provider's prices, in US cents per example and US dollars in total. $^*$Sent at the flex service tier, which has lower prices and can respond more slowly.}
  \label{tab:v1_cost_and_response_time_context_length}
\end{table*}

\FloatBarrier

\raggedbottom

\section{Prompt Templates}
\label{appdx:prompt_templates}

We provide the system and user prompt templates for the three reachability tasks. The question templates and the APA-style template appear in Appendices~\ref{appdx:question_templates} and~\ref{appdx:needle_templates}, respectively. The context describes scientific papers and their citation relationships; author names serve only as publication metadata. Placeholders in braces denote example-specific values.

\subsection{System Prompts}
\label{appdx:system_prompts}

\Cref{fig:yesno_reachability_system_prompt,fig:multiple_choice_reachability_system_prompt,fig:counting_reachability_system_prompt} show the system prompts for the YES/NO reachability, multiple-choice reachability, and reachability counting tasks, respectively.

\subsection{User Prompt Template}
\label{appdx:user_prompt}

All tasks share the user prompt template in \Cref{fig:user_prompt_template}. The question is instantiated from the corresponding template in Appendix~\ref{appdx:question_templates}. The context contains shuffled natural language blocks and distractors, separated by a blank line, i.e., \texttt{\textbackslash{}n\textbackslash{}n}. The response-format reminder, which depends on the task, is shown in \Cref{fig:response_format_reminder_templates}. The ellipsis in the template stands for the remaining blocks and distractors and is not part of the generated message.

\begin{minipage}{\linewidth}
  \section{AI use statement}

  In this work, we used generative AI tools for supporting coding and writing. We have not used generative AI tools for literature review, forming research direction, finding research gap, ideation, or designing method/benchmark/experiments. We have reviewed all AI-assisted work. LLM-generated code was verified and tested for correctness by the first author. We take responsibility for the final content of this work, including text, claims or artifacts produced with the aid of generative AI.
\end{minipage}

\begin{figure*}[tp]
  \begin{tcblisting}{
      promptbox,
      title={System Prompt for YES/NO Reachability}
    }
    You are a careful research assistant answering questions about academic papers using only the provided context.

    You will receive a QUESTION and a CONTEXT containing academic paper documents.

    The CONTEXT is organized into structured blocks about academic papers and their references. A block beginning with Title describes one paper. Each item in References identifies a paper cited by the paper described in that block. The Authors field is publication metadata only; do not follow it as a relationship.

    Use only citation relationships stated in the CONTEXT. Whenever Paper A cites Paper B, there is a `cites' relationship from A to B, and a `cited_by' relationship from B to A. Starting from the SOURCE paper, determine whether there exists a sequence of relationships that leads to the TARGET paper, where each relationship is either a `cites' or `cited_by' relationship.

    Read the QUESTION carefully to identify the SOURCE and TARGET. Use only the information in the provided CONTEXT. The CONTEXT contains all information available for this task. Treat entities and relationships not stated in the CONTEXT as absent. You must not guess the answer based on how closely related the SOURCE and TARGET papers appear to be, such as their topics, research fields, or title wording. Two papers on similar topics may have no connecting sequence of relationships, while two papers on very different topics may still be connected through several intermediate papers.

    Your response may contain reasoning, but its final line must be exactly one of:
    \answer{YES}
    \answer{NO}

    Do not write more than one answer block or put anything after the final answer line.
  \end{tcblisting}
  \caption{System prompt for the YES/NO reachability task.}
  \label{fig:yesno_reachability_system_prompt}
\end{figure*}

\begin{figure*}[tp]
  \begin{tcblisting}{
      promptbox,
      title={System Prompt for Multiple-choice Reachability}
    }
    You are a careful research assistant answering questions about academic papers using only the provided context.

    You will receive a multiple-choice QUESTION and a CONTEXT containing academic paper documents.

    The CONTEXT is organized into structured blocks about academic papers and their references. A block beginning with Title describes one paper. Each item in References identifies a paper cited by the paper described in that block. The Authors field is publication metadata only; do not follow it as a relationship.

    Use only citation relationships stated in the CONTEXT. Whenever Paper A cites Paper B, there is a `cites' relationship from A to B, and a `cited_by' relationship from B to A. For each option, starting from the SOURCE paper, determine whether there exists a sequence of relationships that leads to the paper named by that option, where each relationship is either a `cites' or `cited_by' relationship.

    Read the QUESTION carefully to identify the SOURCE and every option. Use only the information in the provided CONTEXT. The CONTEXT contains all information available for this task. Treat entities and relationships not stated in the CONTEXT as absent. You must not guess the answer based on how closely related the SOURCE paper appears to be to the papers listed in the options, such as similarities in their topics, research fields, or titles. Two papers on similar topics may have no connecting sequence of relationships, while two papers on very different topics may still be connected through several intermediate papers.

    Your response may contain reasoning, but its final line must be exactly one answer block containing the selected uppercase option letter, for example:
    \answer{C}

    Do not write more than one answer block or put anything after the final answer line.
  \end{tcblisting}
  \caption{System prompt for the multiple-choice reachability task.}
  \label{fig:multiple_choice_reachability_system_prompt}
\end{figure*}

\begin{figure*}[tp]
  \begin{tcblisting}{
      promptbox,
      title={System Prompt for Reachability Counting}
    }
    You are a careful research assistant answering questions about academic papers using only the provided context.

    You will receive a QUESTION and a CONTEXT containing academic paper documents.

    The CONTEXT is organized into structured blocks about academic papers and their references. A block beginning with Title describes one paper. Each item in References identifies a paper cited by the paper described in that block. The Authors field is publication metadata only; do not follow it as a relationship.

    Use only citation relationships stated in the CONTEXT. Whenever Paper A cites Paper B, there is a `cites' relationship from A to B, and a `cited_by' relationship from B to A. A paper is reachable from the SOURCE paper if there exists a sequence of relationships that leads from the SOURCE paper to that paper, where each relationship is either a `cites' or `cited_by' relationship.

    Read the QUESTION carefully to identify the SOURCE. Use only the information in the provided CONTEXT. The CONTEXT contains all information available for this task. Treat entities and relationships not stated in the CONTEXT as absent.

    Starting from the SOURCE paper, count every distinct paper reachable through one or more `cites' or `cited_by' relationships. A paper appearing only in a References list is still a paper node and must be counted when reachable. Do not count the SOURCE paper. Paper mentions with the same title and the same complete set of authors denote one paper node. Count that paper only once even if it appears in multiple blocks or reference lists. Paper mentions with the same title but different author sets denote different paper nodes. Ignore superficial differences in capitalization, punctuation, author order, and repeated author names when applying this identity rule. You must not include or exclude a paper based on how closely related it appears to be to the SOURCE paper, such as similarities in their topics, research fields, or titles. A paper on an unrelated topic must be counted if it is reachable, while a paper on a closely related topic must not be counted if it is unreachable.

    Your response may contain reasoning, but its final line must be exactly one answer block containing a non-negative base-10 integer, for example:
    \answer{17}

    Do not use commas or other text inside the answer block. Do not write more than one answer block or put anything after the final answer line.
  \end{tcblisting}
  \caption{System prompt for the reachability counting task.}
  \label{fig:counting_reachability_system_prompt}
\end{figure*}

\begin{figure*}[tp]
  \begin{tcblisting}{
      promptbox,
      title={User Prompt Template Shared by All Tasks}
    }
    # Question

    {QUESTION}

    # Context

    {Natural language block 1}

    {Natural language block 2}

    ...

    # Response Format

    {TASK_SPECIFIC_FORMAT_REMINDER}
  \end{tcblisting}
  \caption{User prompt template shared by all tasks.}
  \label{fig:user_prompt_template}
\end{figure*}

\begin{figure*}
  \begin{tcblisting}{
      promptbox,
      title={Response-Format Reminder for YES/NO Reachability}
    }
    You may give brief reasoning first. End with exactly one final line: \answer{YES} or \answer{NO}. Use only the provided context.
  \end{tcblisting}

  \begin{tcblisting}{
      promptbox,
      title={Response-Format Reminder for Multiple-choice Reachability}
    }
    You may give brief reasoning first. End with exactly one final line in the form \answer{X}, where X is the selected uppercase option letter. Use only the provided context.
  \end{tcblisting}

  \begin{tcblisting}{
      promptbox,
      title={Response-Format Reminder for Reachability Counting}
    }
    You may give brief reasoning first. End with exactly one final line in the form \answer{N}, where N is a non-negative base-10 integer. Do not count the SOURCE. Use only the provided context.
  \end{tcblisting}
  \caption{Response-format reminders for the three reachability tasks. Placeholders in braces denote example-specific values.}
  \label{fig:response_format_reminder_templates}
\end{figure*}

\end{document}